%% file: main.tex
\documentclass[10pt,twocolumn,letterpaper]{article}

\usepackage[dvipsnames]{xcolor}  %
\definecolor{cvprblue}{rgb}{0.21,0.49,0.74}

\usepackage{cvpr}              %
\usepackage{stfloats}          %
\usepackage{graphicx}          %
\usepackage{array}

\usepackage{amsmath}
\usepackage{amssymb}
\usepackage{bm}
\usepackage{bbm}
\usepackage{xspace}

\newcommand\blfootnote[1]{%
  \begingroup
  \renewcommand\thefootnote{}\footnote{#1}%
  \addtocounter{footnote}{-1}%
  \endgroup
}

\usepackage[pagebackref,breaklinks,colorlinks,citecolor=cvprblue]{hyperref}

\author{
    Salvatore Esposito\textsuperscript{1}\textdagger \and
    Qingshan Xu\textsuperscript{3} \and
    Kacper Kania\textsuperscript{4} \and
    Charlie Hewitt\textsuperscript{2} \and
    Octave Mariotti\textsuperscript{1} \and
    Lohit Petikam\textsuperscript{2} \and
    Julien Valentin\textsuperscript{2} \and
    Arno Onken\textsuperscript{1} \and
    Oisin Mac Aodha\textsuperscript{1}
    \\\\
    \textsuperscript{1}University of Edinburgh \quad
    \textsuperscript{2}Microsoft\\
    \textsuperscript{3}Nanyang Technological University
    \textsuperscript{4}Warsaw University of Technology
}

\title{GeoGen: Geometry-Aware Generative Modeling via Signed Distance Functions}

\begin{document}

\maketitle

\begin{abstract}
We introduce a new generative approach for synthesizing 3D geometry and images from single-view collections.  Most existing approaches predict volumetric density to render multi-view consistent images. By employing volumetric rendering using neural radiance fields, they inherit a key limitation: the generated geometry is noisy and unconstrained, limiting the quality and utility of the output meshes. To address this issue, we propose GeoGen, a new SDF-based 3D generative model trained in an end-to-end manner. Initially, we reinterpret the volumetric density as a Signed Distance Function (SDF). This allows us to introduce useful priors to generate valid meshes. However, those priors prevent the generative model from learning details, limiting the applicability of the method to real-world scenarios. To alleviate that problem, we make the transformation learnable and constrain the rendered depth map to be consistent with the zero-level set of the SDF. Through the lens of adversarial training, we encourage the network to produce higher fidelity details on the output meshes. For evaluation, we introduce a synthetic dataset of human avatars captured from 360-degree camera angles, to overcome the challenges presented by real-world datasets, which often lack 3D consistency and do not cover all camera angles. Our experiments on multiple datasets show that GeoGen produces visually and quantitatively better geometry than the previous generative models based on neural radiance fields.
\end{abstract}

\section{Introduction}
\blfootnote{\textdagger Work conducted during an internship at Microsoft.}
The combination of generative models~\cite{mildenhall2020nerf,karras2021alias,karras2019style,karras2020analyzing} and implicit neural representations~\cite{chen2019learning,mescheder2019occupancy,park2019deepsdf} has sparked considerable advancements in 3D representation learning~\cite{chan2021pi,goodfellow2014generative}. It has powered the synthesis of high-quality, multi-view consistent, images. 
However, a common pitfall in the pursuit of higher image quality is the sidelining of the quality of the underlying \emph{geometry}~\cite{yariv2021volume}.

Recent non-generative efforts, such as NeuS~\cite{wang2021neus}, VolSDF~\cite{yariv2021volume}, and Geo-Neus~\cite{fu2022geo}, have made use of the zero-level set of a Signed Distance Function (SDF) to represent the surface of the geometry in a scene via a surface rendering equation, ultimately achieving high-fidelity scene reconstruction. 
While these models have shown impressive potential, given their non-generative nature, they are only able to reconstruct a scene of interest when multi-view image data is available. 
This limitation highlights the need for generative models capable of producing high-quality 2D images that are suitable for content creation while ensuring precise geometric synthesis without multi-view data. 

Other recent methods such as Ball-GAN~\cite{shin2023ballgan}, and EG3D~\cite{chan2022efficient}, have combined generative models with Neural Radiance Fields (NeRFs)~\cite{mildenhall2020nerf} to yield high quality  rendered images. 
Yet, these approaches often result in noisy meshes that contain geometric artifacts, which emerge due to the properties of NeRFs and their lack of constraints on the geometric reconstructions. Attempts have also been made to harmonize SDFs with generative models as in~\cite{or2022stylesdf}. 
However, the generated meshes are often overly smooth, a result of the smoothing prior that encourages the SDF to produce valid values everywhere in 3D space. Additionally, applying this loss can be prohibitive at higher resolutions.

In this work, we address these issues by adding SDF constraints to improve the synthesized geometry of a 3D-aware generative model. 
Our approach, named GeoGen, employs an SDF depth map consistency loss for enhanced geometric generation. 
Specifically, we build on EG3D~\cite{chan2022efficient} by introducing an SDF representation, instead of a density representation, to encode the geometry. 
This allows GeoGen to extract mesh surfaces directly from the zero-level set of the SDF~\cite{oechsle2021unisurf,wang2021neus,yariv2021volume}. 
In order to make the SDF representation learning feasible, and to endow it with the ability to model complex and detailed geometry, we also propose an SDF depth map consistency loss. 
We use a fixed density-to-SDF transformation function to convert the density representation to an SDF.
This facilitates generative feature learning by making the learning objective easier to optimize. 
The SDF also enables the extraction of smooth depth maps that serve as a `pseudo' ground-truth. 
Our approach uses its own depth prediction in a self-supervised manner to improve the reconstruction.
In contrast to commonly used priors, our approach is cheap to compute with only a minor increase in training time. 

GeoGen is able to generate detailed meshes from a single input 2D image via inversion~\cite{roich2022pivotal}. This capability is valuable in applications where the requirement for detailed and realistic meshes is needed. In stark contrast to recent methods like Rodin~\cite{wang2023rodin}, which required 30 million images during training to create 3D meshes, GeoGen uses a fraction of this number -- approximately 50,000 images. 
Other methods such as PanoHead~\cite{an2023panohead} propose an augmented triplane and separate foreground and background in 2D images with the help of a custom in-house dataset. However, with our proposed architecture, we show that by enforcing our geometric constraints,  we are able to reconstruct a detailed 360$^{\circ}$ geometry, with a reduction in visual artifacts (\eg the backs of heads) compared to methods such as  EG3D~\cite{an2023panohead}. 

We make the following contributions: (i) We address the problem of 3D synthesis from 2D images by combining a Signed Distance Function (SDF) network with a StyleGAN generative architecture. Our GeoGen model produces more refined geometry predictions compared to conventional neural volume rendering. 
(ii) We propose an SDF depth map consistency loss that is designed to address geometric inaccuracies from volumetric integration by aligning 3D points with the SDF network's zero-level set for more precise reconstructions.
(iii) We introduce a new dataset of realistic synthetic human heads that contains 360$^{\circ}$ camera views from multiple synthetic humans. 
This dataset will be a valuable resource for training and quantitatively evaluating 3D generative models. 
It can be found on our webpage https://microsoft.github.io/GeoGen.

\section{Related work}

The landscape of generative modeling has seen a shift in recent years, with techniques drawing on neural implicit representations, such as Generative Adversarial Networks (GANs)~\cite{goodfellow2014generative} and Diffusion models~\cite{dhariwal2021diffusion,ho2020denoising,kingma2013auto,song2020score} emerging as powerful tools. 
These techniques blend generative models with neural volume rendering, thereby synthesizing 3D images that capture novel viewpoints from 2D data alone~\cite{mildenhall2020nerf}. 
However, a recurring challenge in this domain has been the reliance on generic density functions to learn the geometry of the images, a factor that often introduces artifacts and results in noisy, low-quality geometric predictions~\cite{or2022stylesdf}. 
To mitigate this, prior work has taken advantage of large amounts of multi-view data to constrain the models, thereby yielding more robust geometry~\cite{wang2021neus,yariv2021volume}, but at the expense of not being fully generative.

The emergence of volumetric implicit representations, bolstered by the strengths of Multi-Layer Perceptrons (MLPs)~\cite{he2015delving} and neural rendering techniques~\cite{mildenhall2020nerf}, has shown substantial promise in extracting detailed geometry from a 3D scene. 
This is most apparent in methods such as NeuS~\cite{wang2021neus} and VolSDF~\cite{yariv2021volume}, which extract high-fidelity surfaces by representing the scene using the Signed Distance Function (SDF) and extracting the surface at the zero level set.

Meanwhile, the broader field of deep learning has seen a surge in novel methods for creating 3D representations from 2D data. One such family of  methods is Neural Radiance Fields (NeRFs)~\cite{mildenhall2020nerf}, which employs a neural network to model the radiance of a 3D scene at any spatial point. The ability of NeRFs to generate high-fidelity 3D models from 2D multi-view supervision, complete with accurate lighting and shading effects, makes them an attractive option for applications requiring realistic 3D representations, such as virtual reality~\cite{yariv2021volume}. 

One set of methods that deserves particular discussion within this landscape is the set of 3D-aware generative models \cite{chan2021pi,deng2022gram,gadelha20173d,gao2022get3d,henderson2020leveraging,nguyen2019hologan,nguyen2020blockgan,niemeyer2021giraffe,schwarz2022voxgraf}. 
These methods are specifically designed to generate 3D representations of objects or scenes, utilizing a variety of techniques, including volumetric representations, SDFs, and implicit neural representations. 
For instance, the Generative Radiance Fields (GRAF) model~\cite{schwarz2020graf} generates high-resolution 3D shapes with intricate detail, leveraging a neural network to model the radiance and shape of a 3D object. 
Other notable models include DeepSDF~\cite{park2019deepsdf}, which learns continuous signed distance functions for arbitrary shapes using 3D supervision, and HoloGAN~\cite{nguyen2019hologan}, which generates 3D objects by imposing structural constraints in the generative process. 
Recently, EG3D~\cite{chan2022efficient} proposed a triplane representation for volume rendering in generative models, which enables efficient 3D-aware generation. 
However, extracting high quality 3D meshes is not guaranteed because of its use of a volume density representation. 
StyleSDF~\cite{or2022stylesdf}, makes use of an SDF representation to directly model geometry, but the extracted surfaces are overly smooth making it challenging to use them in practical applications. 

In our investigation of 3D-aware generative models and SDF representations, we identify certain limitations inherent in existing methodologies. One such limitation appears to be a result of the use of the Eikonal loss~\cite{yariv2021volume,wang2021neus,fu2022geo}, leading to overly smooth geometry synthesis. %
Our methodology, building on the foundation laid by EG3D, aims to overcome this by introducing an SDF depth-consistency constraint. This novel constraint is designed to refine geometric surface predictions by leveraging a self-supervised depth prediction mechanism. Unlike previous efforts, such as StyleSDF~\cite{or2022stylesdf}, which merely translates SDF values into density fields, our approach harnesses the full potential of SDF for geometry representation as exemplified by VolSDF~\cite{yariv2021volume}. We emphasize that incorporating our SDF representation and its associated constraints does not substantially complicate the training of generative models yet provides enhanced control over geometric surface detail.

\begin{figure*}[t]
\begin{center}
   \includegraphics[width=\textwidth]{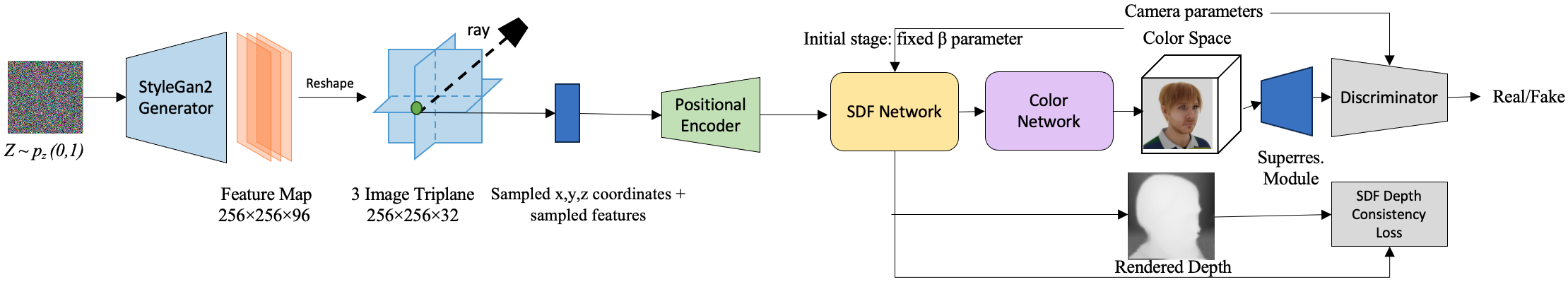}   
\end{center}
\vspace{-15pt}
   \caption{GeoGen, our 3D-aware generator, is trained solely from 2D images. Noise sampling is followed by a StyleGan2 generator that produces triplane features similar to EG3D~\cite{chan2022efficient}. However, we enhance them with positional info and an SDF network for refined geometry. GeoGen is end-to-end trained with a GAN objective along with our SDF depth consistency loss.
   }
   \vspace{-10pt}
\label{fig:long}
\end{figure*}

\section{Method}

Here we present our GeoGen generative approach for enhanced geometric synthesis. 
We begin by revisiting EG3D~\cite{chan2022efficient}, an efficient geometry-aware 3D GAN that introduces notation and provides context for our contributions. 
Then we describe our SDF-based generative model which builds on the EG3D framework. 

\subsection{Efficient geometry-aware 3D GAN}
EG3D~\cite{chan2022efficient} is an efficient geometry-aware 3D generative adversarial network. 
It consists of a StyleGAN2~\cite{karras2020analyzing} based feature generator, triplane representation, implicit volume render, and super-resolution module. In order to generate an image, it first samples a random latent noise code and processes the code via a mapping network. The processed code is used to drive the StyleGAN2 generator to produce feature maps which are reshaped to form three feature planes. During the volume rendering, a queried 3D point $\bm{p}$ is projected onto each of the three feature planes, leading to corresponding feature vector $[F_{xy}(\bm{p}), F_{xz}(\bm{p}), F_{yz}(\bm{p})]$. These feature vectors are further processed by a shallow MLP to yield the color and density at the position $\bm{p}$. 
By the process of volumetric integration, a low-resolution image is generated based on the sampled points along all image rays. Finally, a super-resolution module is used to generate high-resolution output images.

Like EG3D,  we also use a triplane representation to efficiently generate images. Different from EG3D, which targets geometry-aware \emph{image} synthesis, we focus on high-quality \emph{geometry} synthesis. 
To this aim, we introduce an SDF-based generative model and present a novel SDF learning strategy.

\subsection{SDF-based generative model}
Our goal is to develop a model that can learn to generate 3D consistent object-centric images with associated geometry by making use of a collection of posed single-view 2D images at training time.
This transformation is achieved by conceptualizing the surface as the zero-level set of a neural implicit signed distance function. 
To achieve our high-fidelity geometric synthesis, we first introduce our augmented triplane representation. Then, we introduce our SDF-based volume rendering. Finally, we describe an SDF depth-consistency constraint, which is used to enhance SDF learning. Figure~\ref{fig:long} displays our overall pipeline. 

{\bf \noindent Augmented triplane representation.} Our method augments the original EG3D triplane representation with sampling position $\bm{p}$. According to the sampling position $\bm{p}$, we retrieve the corresponding feature vector $[F_{xy}(\bm{p}), F_{xz}(\bm{p}), F_{yz}(\bm{p})]$ via bilinear interpolation. In addition, the position $\bm{p}$ is processed by a position embedder $PE(\cdot)$ that employs multi-level sine and cosine functions similar to NeRFs~\cite{mildenhall2020nerf}:
\begin{equation}
    PE(a) = [a, \gamma_{0}(a), \gamma_{1}(a),\dots,\gamma_{L-1}(a)],
\end{equation} 
where $\gamma_{k}(a) = [\sin(2^{k} \pi a), \cos(2^{k} \pi a)]$,  $L$ is a hyper-parameter that controls the maximum encoded frequency, and $a$ represents each of the three different spatial dimensions of $\bm{p}$. $\bm{p}$ is defined as a vector since it represents the position in 3D space. Each component of $\bm{p}$ (i.e., $p_x$, $p_y$, $p_z$) corresponds to a different spatial dimension.

The function $\gamma_k(a)$ is a positional encoding function that takes a scalar value $a$ and returns a 2D vector representation of the sine and cosine of $2^k \pi a$. This function is used for positional encoding to capture frequency information up to a maximum frequency defined by the hyper-parameter $L$.

The augmented triplane representation is formed by concatenating the triplane features $F_{xy}(\bm{p})$, $F_{xz}(\bm{p})$, and $F_{yz}(\bm{p})$ with the positional encoding $PE(p_x)$, $PE(p_y)$, and $PE(p_z)$. This augmented representation enables the model to capture high-frequency details by combining the local geometric features with positional encoding information.
The absence of the positional encoder destabilizes the training process, often resulting in model collapse (see supplementary material for results). 

{\bf \noindent SDF-based volume rendering.} The augmented tri-plane representation is directed to a shallow MLP to learn the SDF value $s(\bm{p})$ and RGB color ${\bf c}(\bm{p})$ for point $\bm{p}$. 
The SDF value represents the distance to the surface, providing an accurate depiction of its geometry. To convert the SDF value $s(\bm{p})$ into a density field $\sigma$, we follow VolSDF~\cite{yariv2021volume} and use the following Laplace transformation: 
\begin{equation}
  \sigma(s(\bm{p}))=\begin{cases} \frac{1}{2\beta} \exp\left(\frac{s(\bm{p})}{\beta}\right) & \text{if } s(\bm{p}) \leq 0 \\
  \frac{1}{\beta}\left(1-\frac{1}{2} \exp\left(-\frac{s(\bm{p})}{\beta}\right)\right) & \text{if } s(\bm{p})>0 
  \end{cases},
  \label{eq:laplace}
\end{equation}
where $\beta$ is a parameter, which can be fixed or learned. Based on the volumetric integration, the rendered RGB color for a ray ${\bf r}(t) = {\bf o} + t{\bf d}$ is calculated as follows: 
\begin{equation}
    C({\bf r}) = \sum_{i=1}^{M} T_{i} (1 - \exp(- \sigma_{i} \delta_{i})){\bf c}_{i},
\end{equation}
where ${\bf o}$ is the camera position, ${\bf d}$ is the ray direction, $T_{i} = \exp(-\sum_{j=1}^{i-1}\sigma_{j} \delta_{j})$ and $\delta_{i}=t_{i+1} - t_{i}$ is the distance between adjacent sampled points. For simplicity, we use $\sigma_i$ and ${\bf c}_i$ to denote $\sigma(s(\bm{p}_i))$ and ${\bf c}(\bm{p}_i)$ respectively, which mean the color and density value at the \textit{i-th} sampling point $\bm{p}_i$ along ray ${\bf r}$. 
In a similar way, we compute the rendered distance as follows:
\begin{equation}
    d({\bf r}) = \sum_{i=1}^{M} T_{i} (1 - \exp(- \sigma_{i} \delta_{i}))t_{i}.
    \label{eq:dist}
\end{equation}

{\bf \noindent SDF depth consistency.} 
It has been shown in Geo-Neus~\cite{fu2022geo} that there can exist a gap between the rendered image and the true surface and it is important to introduce  explicit constraints to optimize the SDF network. 
Therefore, Geo-Neus introduces sparse points and multi-view photometric consistency to achieve this in the multi-view setting when multiple images are available for each object during training. 
However, these two constraints are obviously not available in our \emph{single}-view GAN setting. To reduce the geometry bias caused by volumetric integration, the 3D point computed from the rendered distance $d({\bf r})$ in Equation~\ref{eq:dist} should be located on the zero-level set of the SDF network. 
Thus, according to the rendered distance $d({\bf r})$, its corresponding 3D point ${\bf p}_{d({\bf r})}$ is computed as: 
\begin{equation}
    {\bf p}_{d({\bf r})} = {\bf o} + d({\bf r}) {\bf d}.
\end{equation}
Since the above 3D point should be approximately on the geometry surface, the SDF value of this point should be approximately zero. Thus, we define an SDF constraint as:
\begin{equation}
    \mathcal{L}_{s} = \frac{1}{|\mathcal{R}|}\sum_{\bf r \in \mathcal{R}} |s({\bf p}_{d(\bf r)})|,
\end{equation} 
where $\mathcal{R}$ denotes all rays for the current camera pose. During training we aim to minimize the above loss. %

\subsection{Training GeoGen}
The SDF-based GeoGen model uses dual discrimination during training, evaluating both the neurally rendered low-resolution 2D image and the super-resolved 2D image. The generative model takes only 2D images as input, and the discriminator encourages both the low-resolution and super-resolved synthesized 2D images to match the distribution of real images. This ensures the consistency between the super-resolved images and the neural rendering, facilitating our method to achieve high-quality high-resolution rendering results. In addition, the SDF depth consistency loss is imposed during training to promote geometric consistency. 
The model can then effectively learn to capture accurate geometry information from the 2D images, leading to more precise and reliable 3D reconstructions.  
Our overall loss is:
\begin{equation}
    \mathcal{L} = \mathcal{L}_{dis} + \lambda \mathcal{L}_{s},
    \label{eq:loss_overal}
\end{equation}
where $\mathcal{L}_{dis}$ is a GAN loss computed using dual discrimination and $\lambda$ is a weighting applied to the SDF constraint. 
Empirically we find that directly training our model from scratch is challenging. 
We suspect that the introduced learnable parameter $\beta$ in Equation~\ref{eq:laplace} prevents the StyleGAN2-based feature generator from learning effective features. 
In addition, the SDF constraint requires good geometry initialization, which is not possible to obtain in the early phase of training. 
Therefore, we design a learning strategy to train our model in which the $\beta$  parameter of the Laplace density distribution is fixed to stabilize the early learning of our generative model. 

Specifically, the significant part of this training process involves managing the $\beta$ parameter of the Laplace transformation in Equation~\ref{eq:laplace}, which directly influences the learning of the SDF network. The $\beta$ parameter remains fixed for the first $N$ iterations to allow the SDF network to focus on learning coarse geometry. 
This enables the learning of the StyleGAN2-based generator to produce stable view synthesis. 
After $N$ iterations, we make the $\beta$ a learnable parameter to increase the ability of the model to capture finer-scale surface details.  
As previously mentioned, the SDF constraint should also be carefully managed. We achieve this by controlling the weight $\lambda$ in Equation~\ref{eq:loss_overal}, where it is initially set to $0$, and then increased to $0.1$ after $N$ iterations. 
As a result, our geometry optimization is conducted in a quasi coarse-to-fine fashion, \ie $N$ iterations, our Geo-Gen learns coarse geometry and then after this, the SDF constraint can concentrate on surface detail refinement.

\section{Synthetic human head dataset} 
Existing methods typically train their models on high resolution human face datasets such as Flickr-Faces-HQ (FFHQ)~\cite{karras2019style}. 
However, FFHQ only contains a limited range of captured viewpoints (\ie no backs of heads) and has no 3D ground-truth, hence the need for our synthetic dataset.  
There are other synthetic datasets, such as ShapeNet Cars~\cite{chang2015shapenet}, which have ground-truth 3D meshes but are not realistic looking. 

To address this, we created a new dataset of semi-realistic synthetic human heads which is generated based on the work of~\citet{wood2021fake}. 
Our dataset features images of different synthetic individuals with diverse facial features, body morphologies, clothing, and hair styles. 
Crucially, unlike FFHQ which primarily captures frontal views, our dataset includes images across the full azimuth range, ensuring comprehensive representation of heads from all sides. This approach not only fills a critical gap in available resources but also shifts the focus towards the quality of the mesh, a vital aspect for advancing the field of 3D generative modeling.

\begin{figure}[t]
\resizebox{\columnwidth}{!}{
 \includegraphics[width=0.9\textwidth]{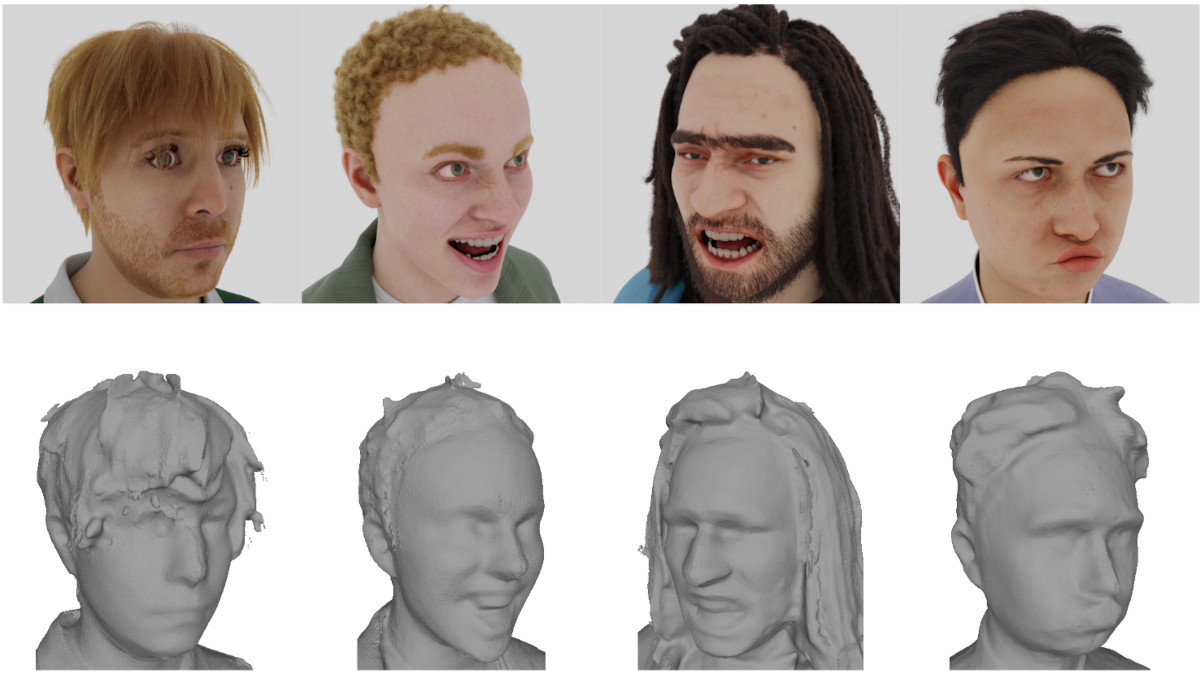}   
}
\vspace{-15pt}
   \caption{Examples from our synthetic human dataset. We display rendered images on top and pseudo 3D  ground-truth below.
   }
   \vspace{-7pt}
\label{fig:synth_data_ex}

\end{figure}

For our dataset, we randomly generate 10 images of 512$\times$512 for each of 19,800 identities, ensuring a comprehensive set of different views, encompassing full azimuthal coverage and utilize multi-view stereo and surface reconstruction techniques to establish pseudo ground-truth meshes. To generate a pseudo ground-truth mesh for quantitative evaluation of 3D reconstruction metrics  we use the ACMP multi-view stereo approach from~\cite{xu2020planar} and Poisson surface reconstruction \cite{kazhdan2006poisson} to reconstruct the full head geometry.  
Example images can be found in Figure~\ref{fig:synth_data_ex}. 
A subset of images from our synthetic dataset will be made available upon acceptance.

\begin{table}[t]
    \centering
    \resizebox{1.0\columnwidth}{!}{
    \begin{tabular}{lllll}
        \toprule
        Dataset & Method & FID$\downarrow$ & KID$\downarrow$ & ID$\uparrow$\\
        \midrule
        FFHQ & GRAF & 79.20 & 55.00 & - \\
        & PiGAN & 83.00 & 85.80 & 0.67 \\
        & GIRAFFE & 31.20 & 20.10 & 0.64 \\
        & HoloGAN & 90.90 & 75.50 & - \\
        & StyleSDF & 11.50 & 2.65 & - \\
        & EG3D & 4.86 & 0.0053 &  0.77 \\
        & EG3D (rebalanced) & \textbf{4.70} & \textbf{0.0044} &  \textbf{0.79} \\
        & EG3D$^{**}$ & 5.70 & 0.0054 & 0.76 \\
        & \textbf{GeoGen} & 5.40 & 0.0049 & 0.75 \\
        \midrule
        Synthetic Heads 
        & EG3D$^{**}$ & 5.90 & 0.65 & \textbf{0.69} \\
        & \textbf{GeoGen} & \textbf{5.10} & \textbf{0.0038} & \textbf{0.69} \\
        \midrule
        ShapeNet Cars 
        & GIRAFFE & 27.30 & 1.70 & -\\
        & Pi-GAN & 17.30 & 0.93 & -\\
        & EG3D & 2.75 & 0.0054 & -\\
        & EG3D$^{**}$ & 2.90 & 0.0043 & - \\
        & \textbf{GeoGen} & \textbf{2.50} & \textbf{0.0028} & -\\
        \bottomrule
    \end{tabular}
    }
    \vspace{-5pt}
    \caption{Comparative analysis of different generative models on FFHQ, our Synthetic Heads, and ShapeNet Cars datasets using standard 2D metrics. Our model surpasses EG3D~\cite{chan2022efficient} and other leading models in both FID and ID metrics for the Synthetic Heads and ShapeNet V1 datasets. However, it does not outperform EG3D on the FFHQ dataset, attributed to a lower number of training iterations due to limited computational resources. Additionally, the original number of training epochs for achieving the reported FID results in EG3D is not specified by its authors. GeoGen was not included in training on the FFHQ rebalanced dataset due to its unavailability during the training period.
    $^{**}$ indicates our retraining with far fewer iterations and computation power.%
    }
    \vspace{-7pt}
    \label{tab:datasets}
\end{table}

\section{Experiments} 
Here we present qualitative and quantitative results comparing GeoGen to existing methods. 
For the baseline EG3D model, we retrained it on each of the evaluation datasets so that the training settings were consistent with our approach (\eg the same number of training epochs). 
Implementation details are provided in the supplementary material.

\begin{figure*}[t]
    \begin{minipage}[t]{0.5\textwidth}
        \centering
        \scriptsize
        \setlength{\tabcolsep}{0pt}
        \begin{tabular}{ccccccc}
            \rotatebox{90}{\parbox[t]{2cm}{\centering EG3D}}\hspace*{5pt}
            & \includegraphics[width=1.7cm,height=1.7cm,keepaspectratio]{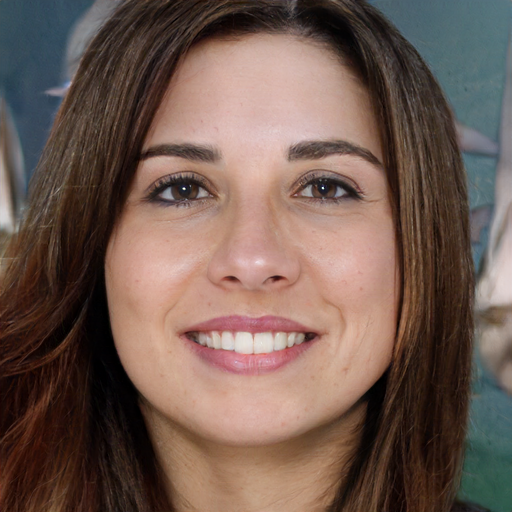}
            & \includegraphics[width=1.7cm,height=1.7cm,keepaspectratio]{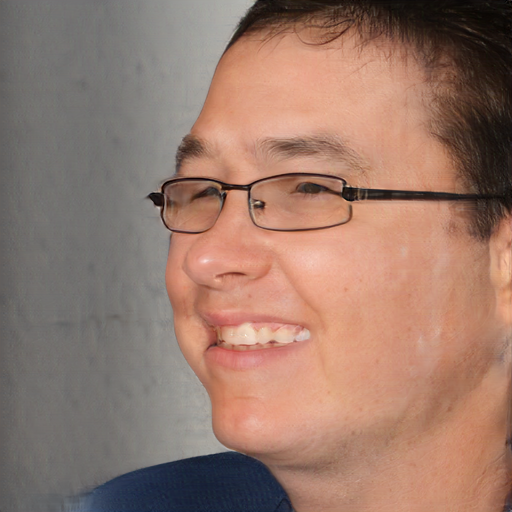}
            & \includegraphics[width=1.7cm,height=1.7cm,keepaspectratio]{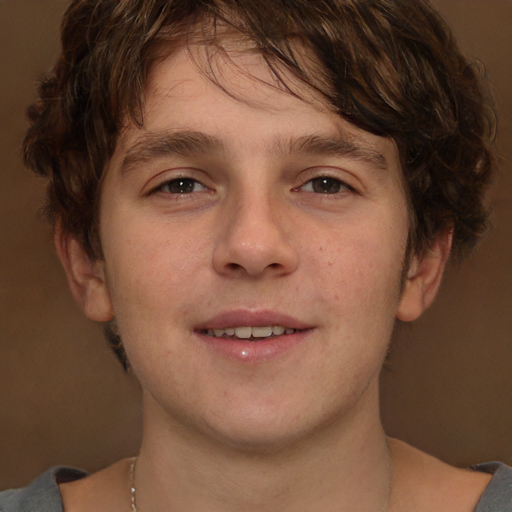}
            & \includegraphics[width=1.7cm,height=1.7cm,keepaspectratio]{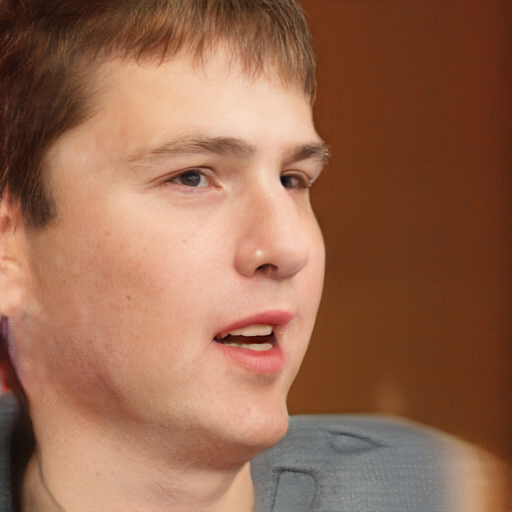}
            & \includegraphics[width=1.7cm,height=1.7cm,keepaspectratio]{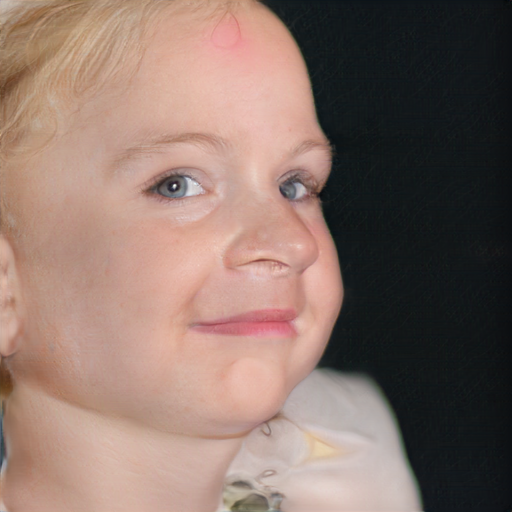} \\
            \rotatebox{90}{\parbox[t]{2cm}{\centering StyleSDF}}\hspace*{5pt}
            & \includegraphics[width=1.7cm,height=1.7cm,keepaspectratio]{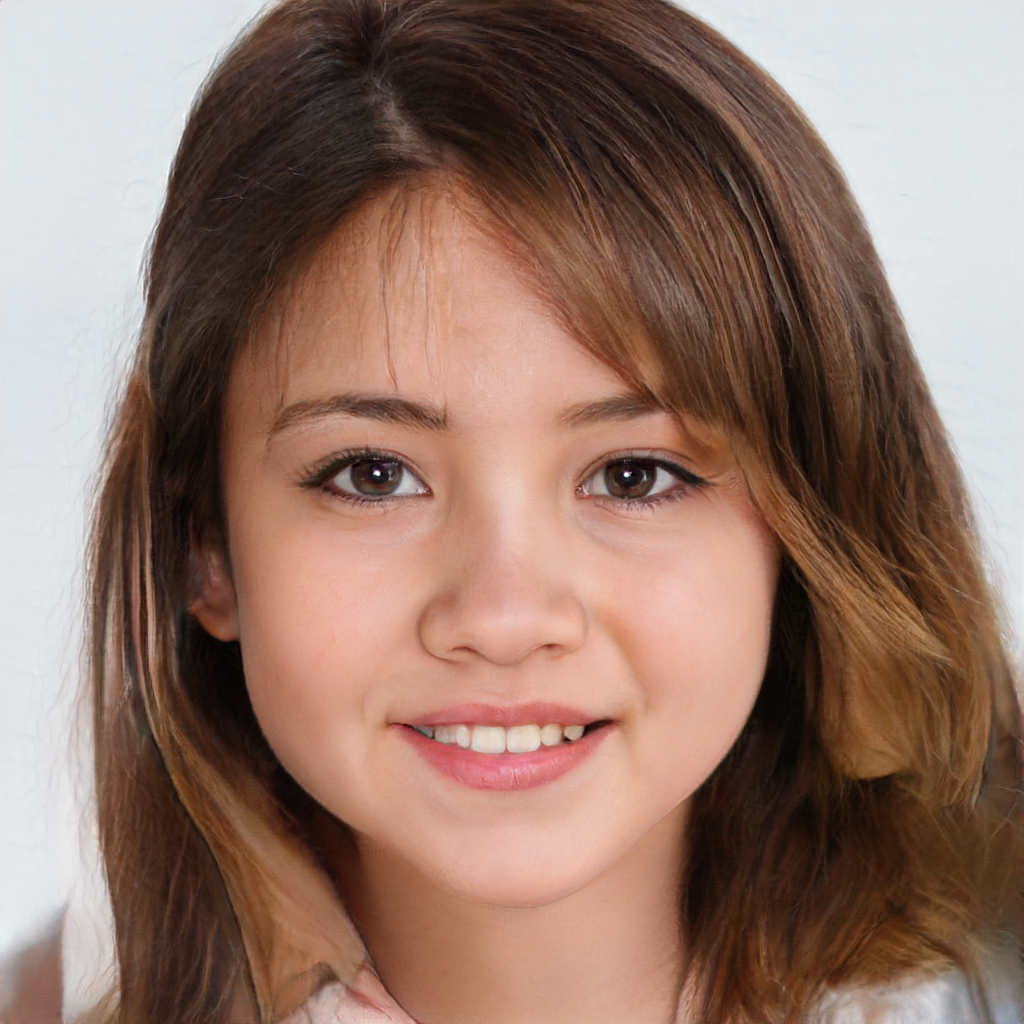}
            & \includegraphics[width=1.7cm,height=1.7cm,keepaspectratio]{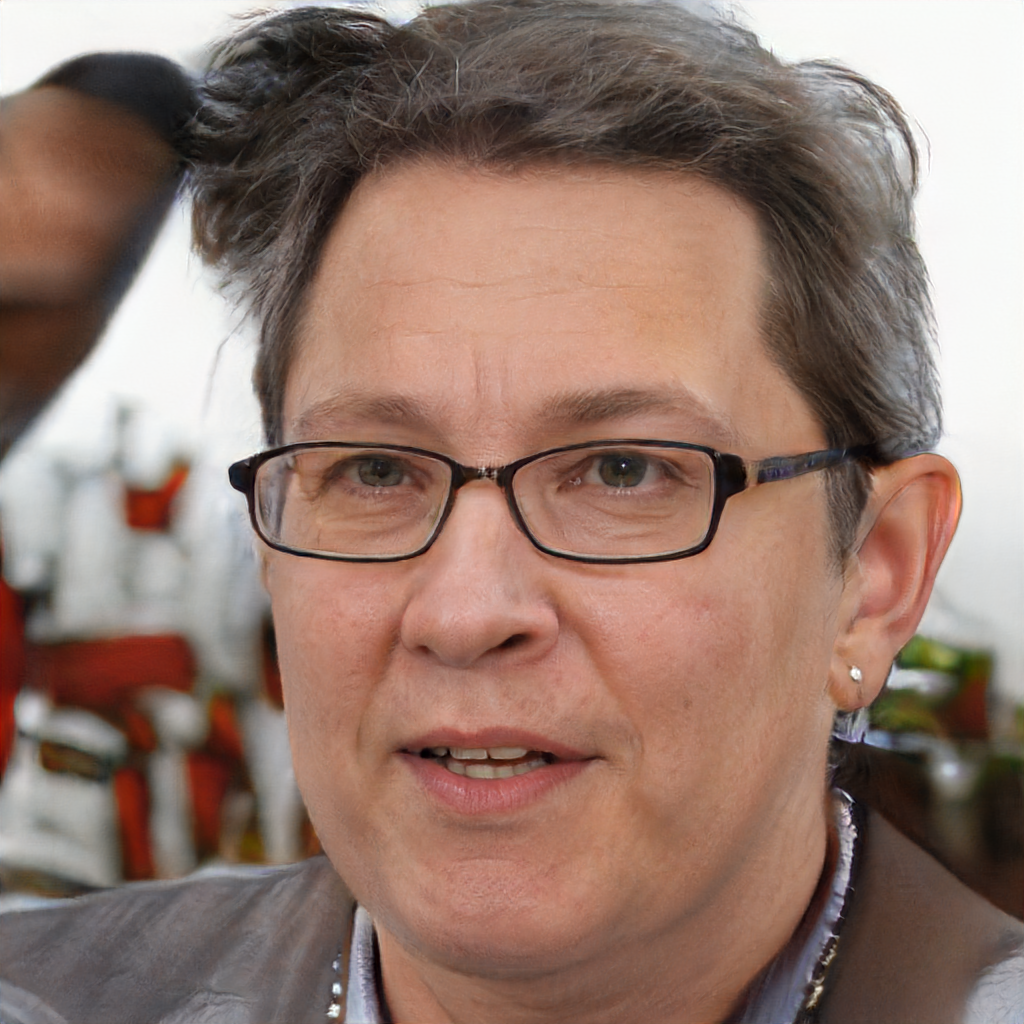}
            & \includegraphics[width=1.7cm,height=1.7cm,keepaspectratio]{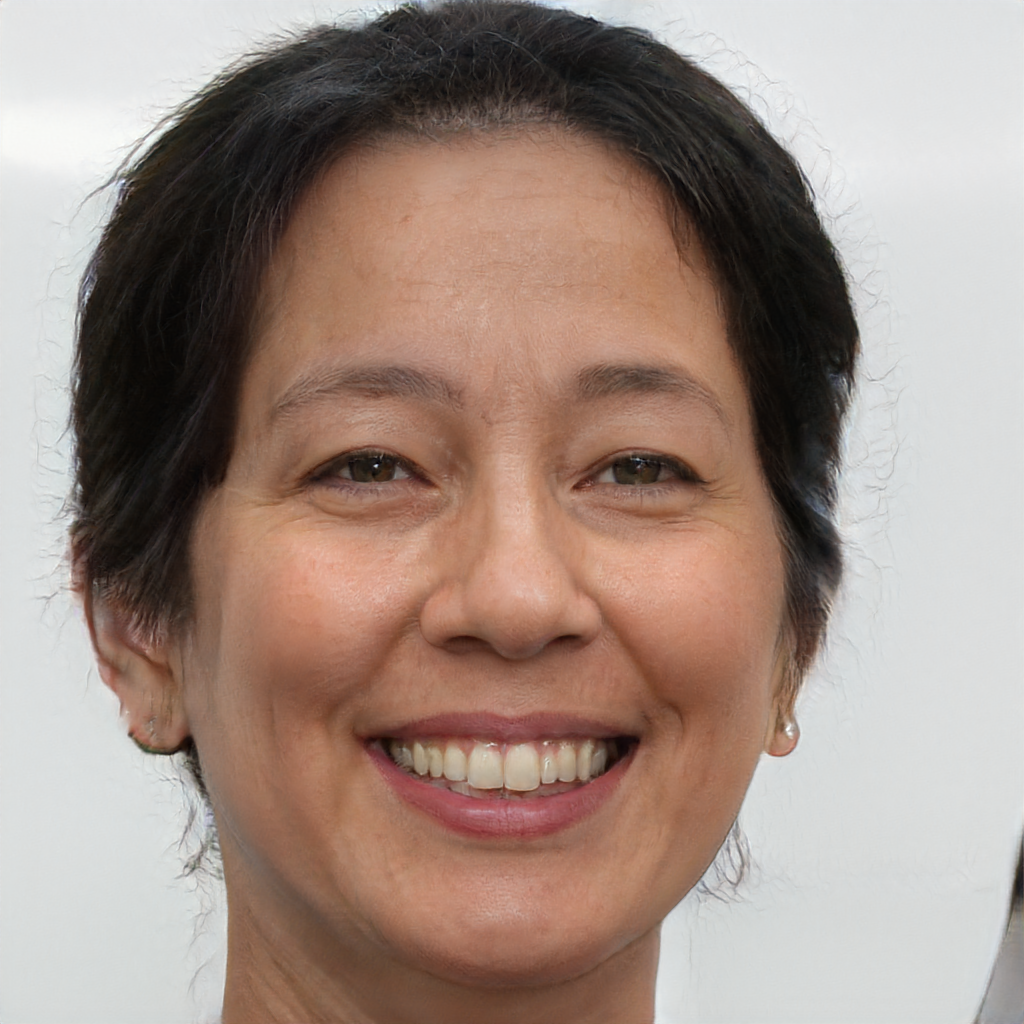}
            & \includegraphics[width=1.7cm,height=1.7cm,keepaspectratio]{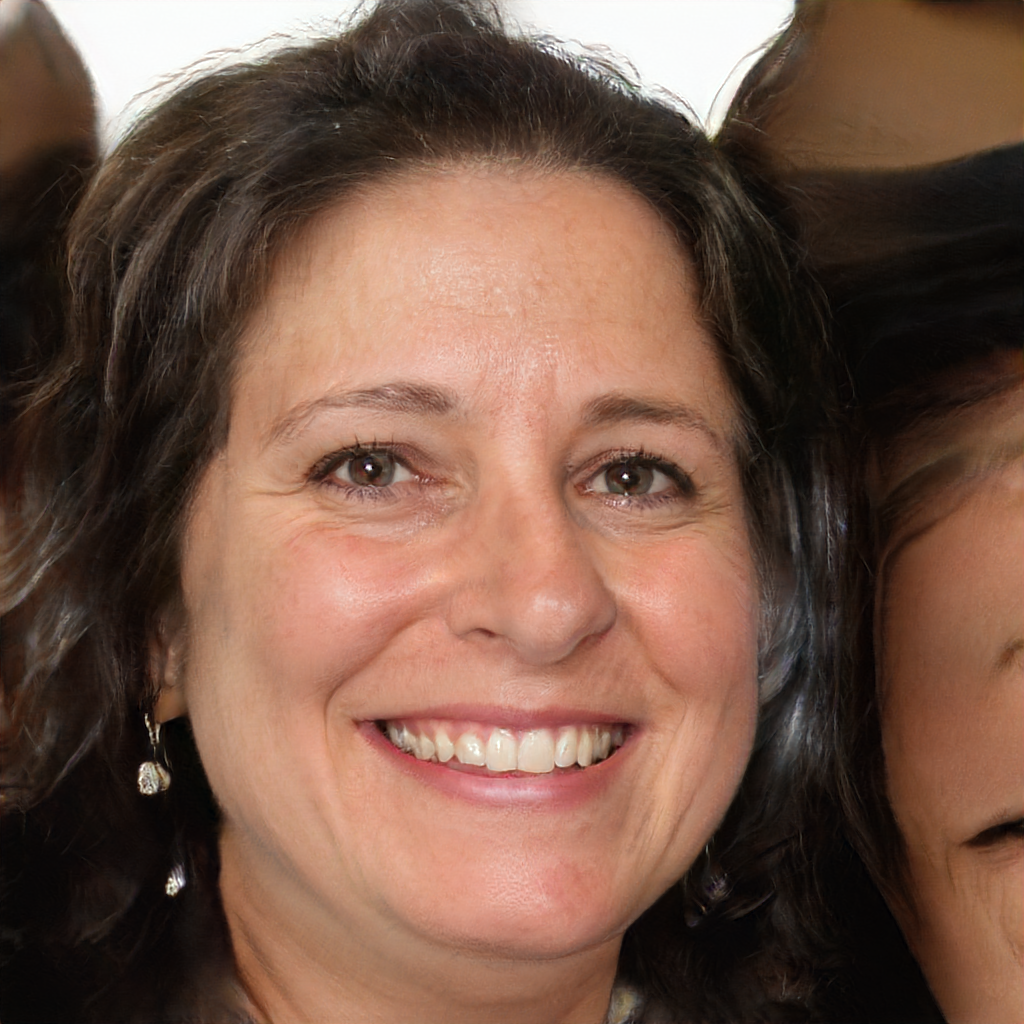}
            & \includegraphics[width=1.7cm,height=1.7cm,keepaspectratio]{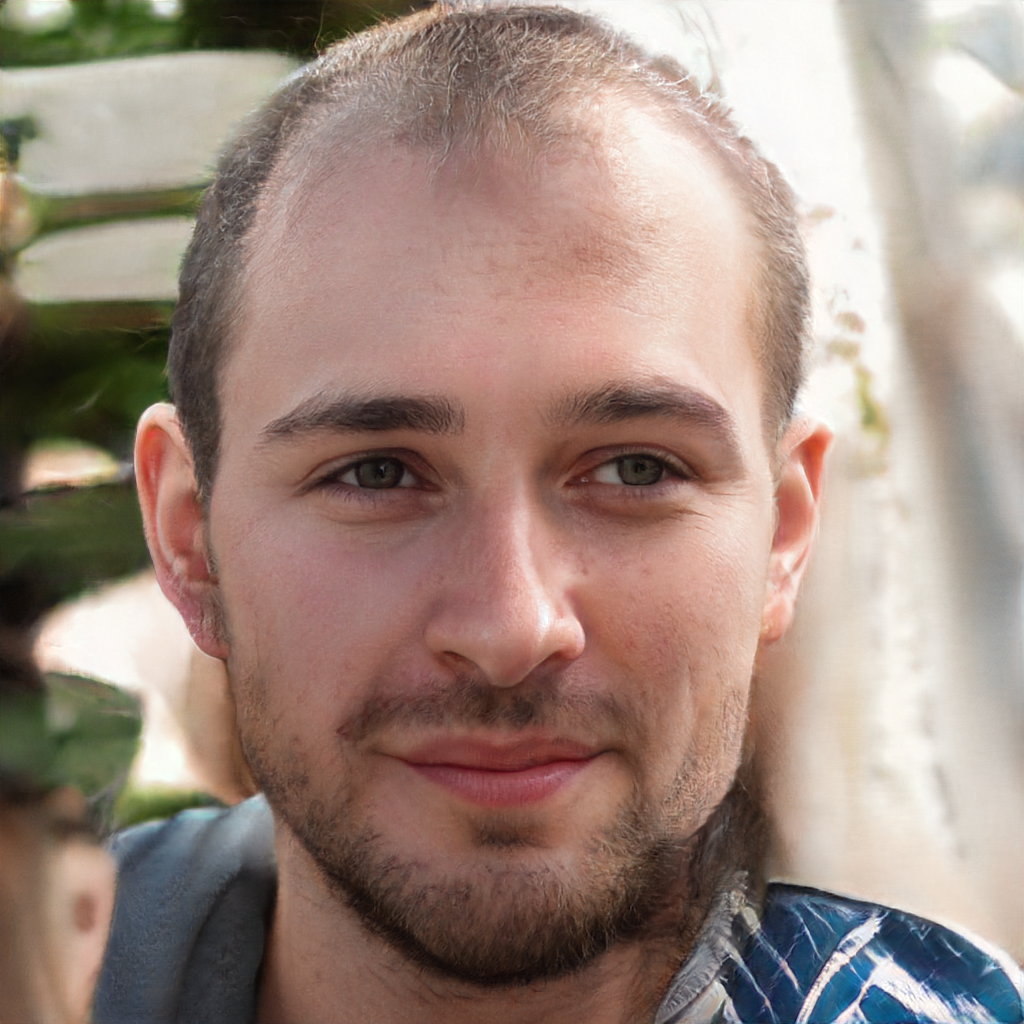} \\
            \rotatebox{90}{\parbox[t]{2cm}{\centering GeoGen}}\hspace*{5pt}
            & \includegraphics[width=1.7cm,height=1.7cm,keepaspectratio]{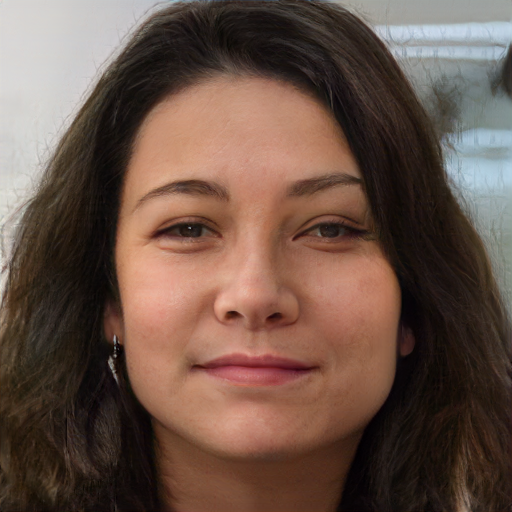}
            & \includegraphics[width=1.7cm,height=1.7cm,keepaspectratio]{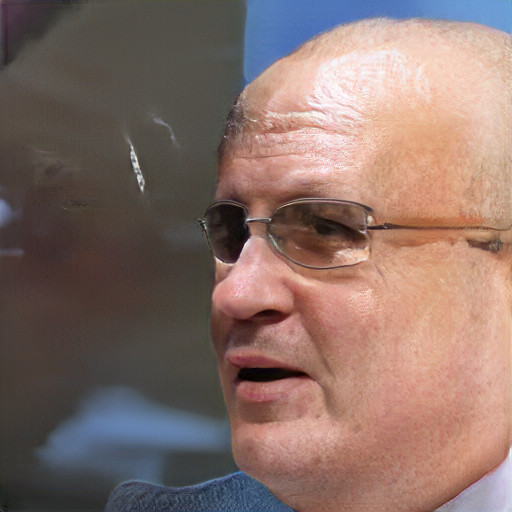}
            & \includegraphics[width=1.7cm,height=1.7cm,keepaspectratio]{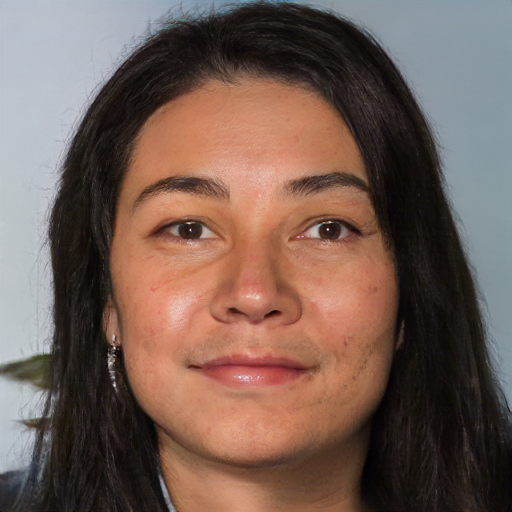}
            & \includegraphics[width=1.7cm,height=1.7cm,keepaspectratio]{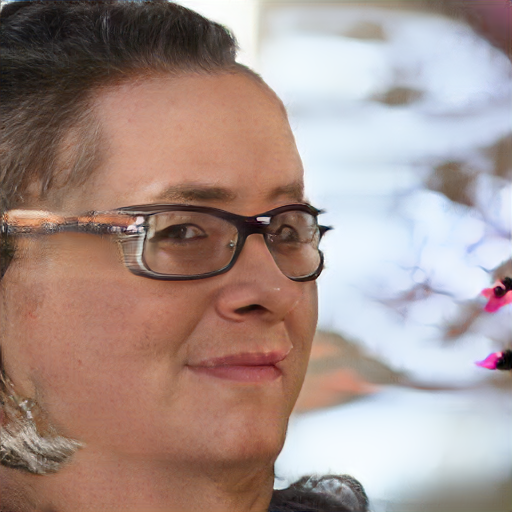}
            & \includegraphics[width=1.7cm,height=1.7cm,keepaspectratio]{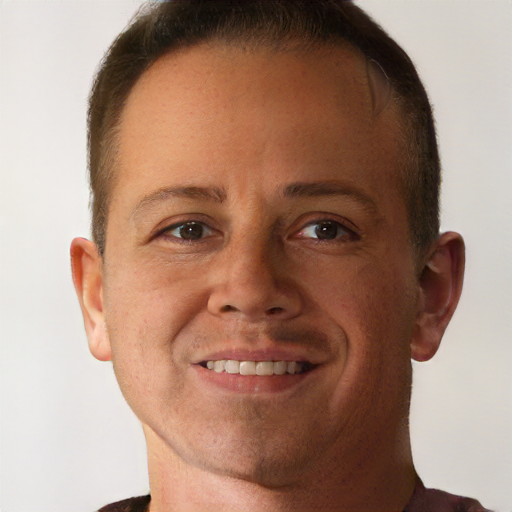} \\
        \end{tabular}
    \end{minipage}%
    \hfill
    \begin{minipage}[t]{0.5\textwidth}
    \centering
    \scriptsize
    \setlength{\tabcolsep}{0pt}
    \begin{tabular}{ccccccc}
        \phantom{\rotatebox{90}{\parbox[t]{2cm}{\centering EG3D}}}\hspace*{5pt}
        & \includegraphics[width=1.7cm,height=1.7cm,keepaspectratio]{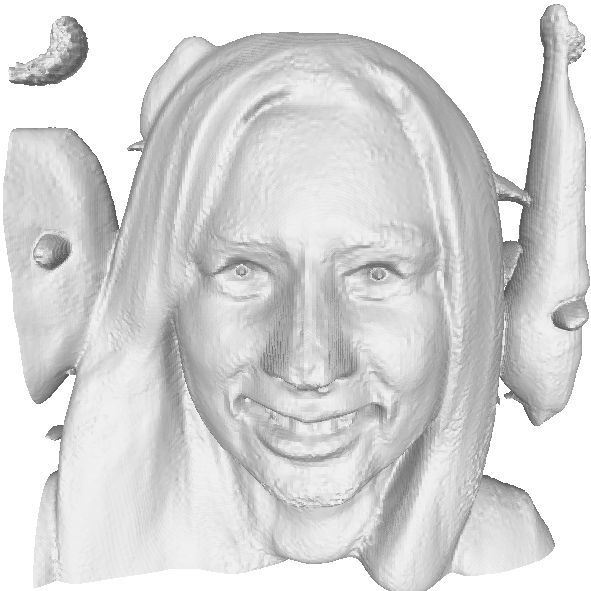}
        & \includegraphics[width=1.7cm,height=1.7cm,keepaspectratio]{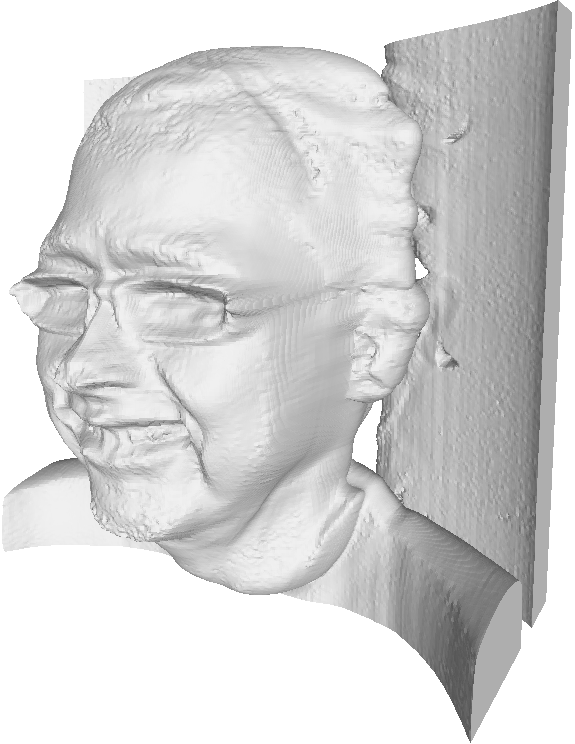}
        & \includegraphics[width=1.7cm,height=1.7cm,keepaspectratio]{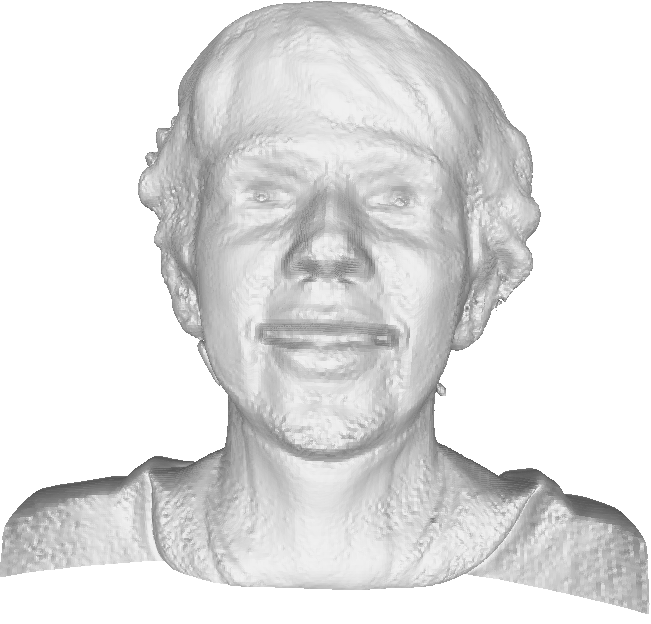}
        & \includegraphics[width=1.7cm,height=1.7cm,keepaspectratio]{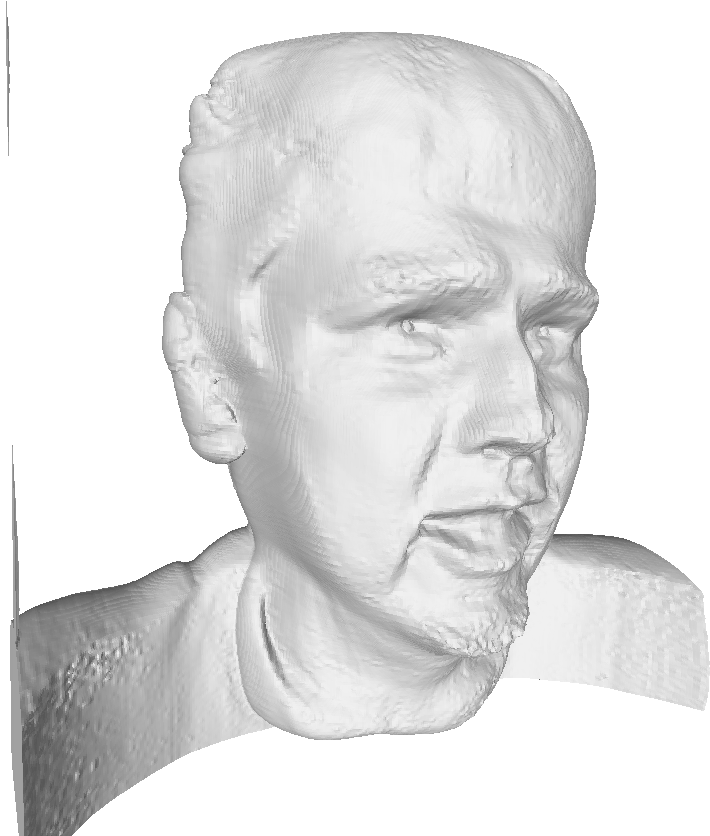}
        & \includegraphics[width=1.7cm,height=1.7cm,keepaspectratio]{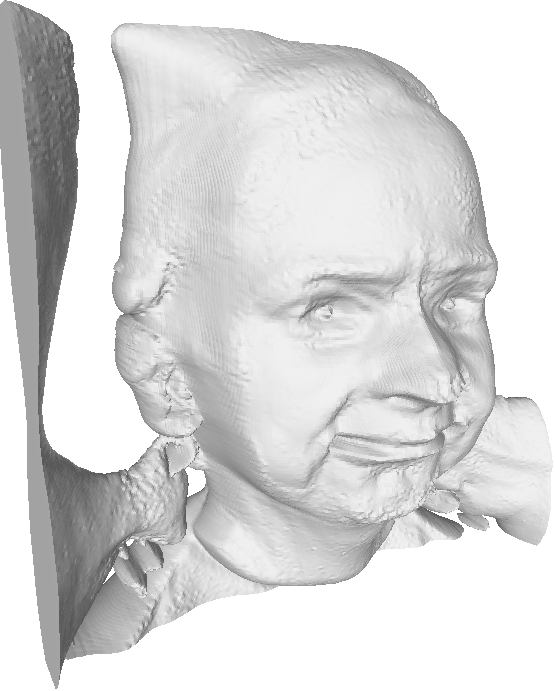} \\
        \phantom{\rotatebox{90}{\parbox[t]{2cm}{\centering GeoGen w/o SDF\&Depth Loss}}}\hspace*{5pt}
        & \includegraphics[width=1.7cm,height=1.7cm,keepaspectratio]{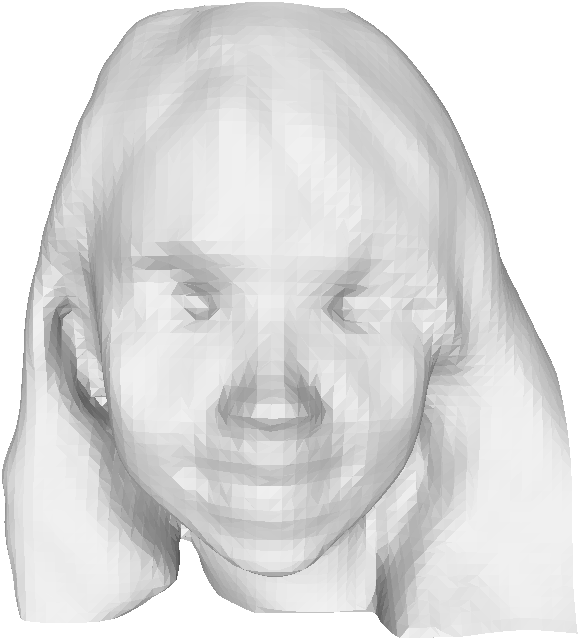}
        & \includegraphics[width=1.7cm,height=1.7cm,keepaspectratio]{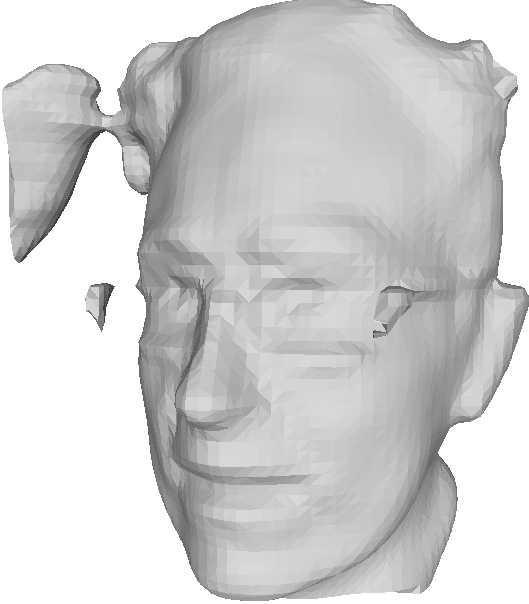}
        & \includegraphics[width=1.7cm,height=1.7cm,keepaspectratio]{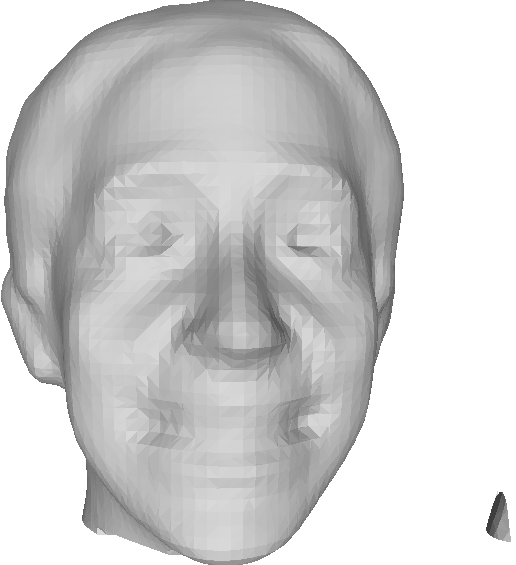}
        & \includegraphics[width=1.7cm,height=1.7cm,keepaspectratio]{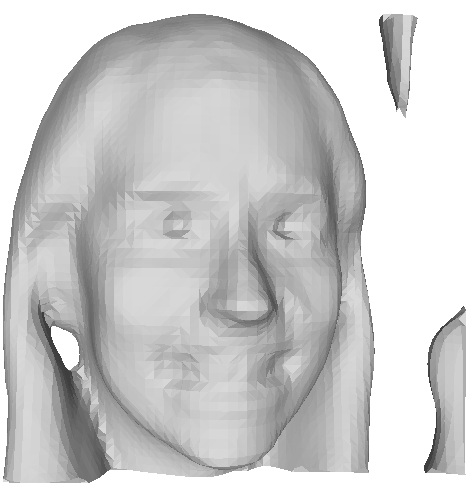}
        & \includegraphics[width=1.7cm,height=1.7cm,keepaspectratio]{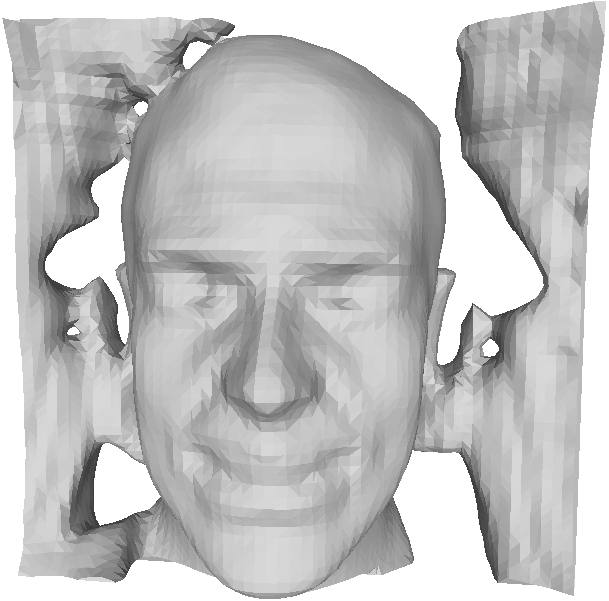} \\
        \phantom{\rotatebox{90}{\parbox[t]{2cm}{\centering GeoGen}}}\hspace*{5pt}
        & \includegraphics[width=1.7cm,height=1.7cm,keepaspectratio]{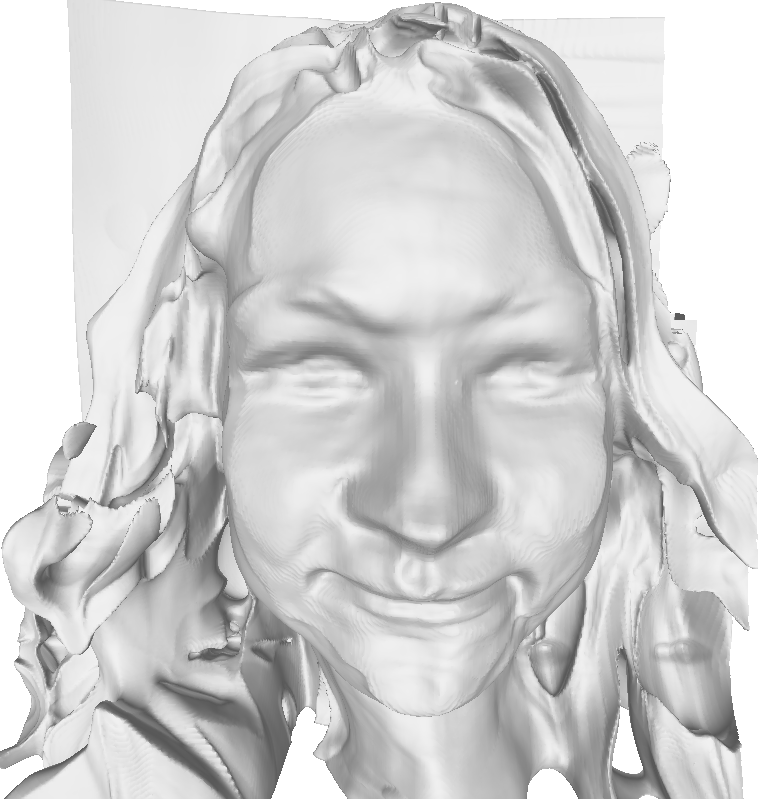}
        & \includegraphics[width=1.7cm,height=1.7cm,keepaspectratio]{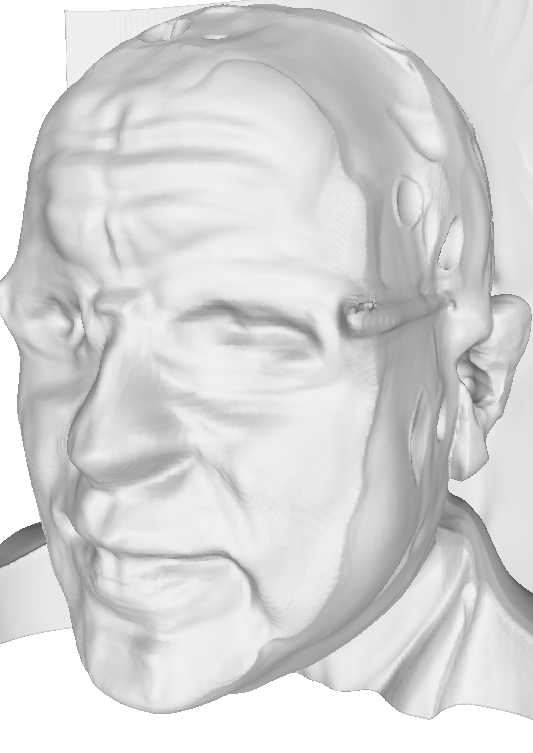}
        & \includegraphics[width=1.7cm,height=1.7cm,keepaspectratio]{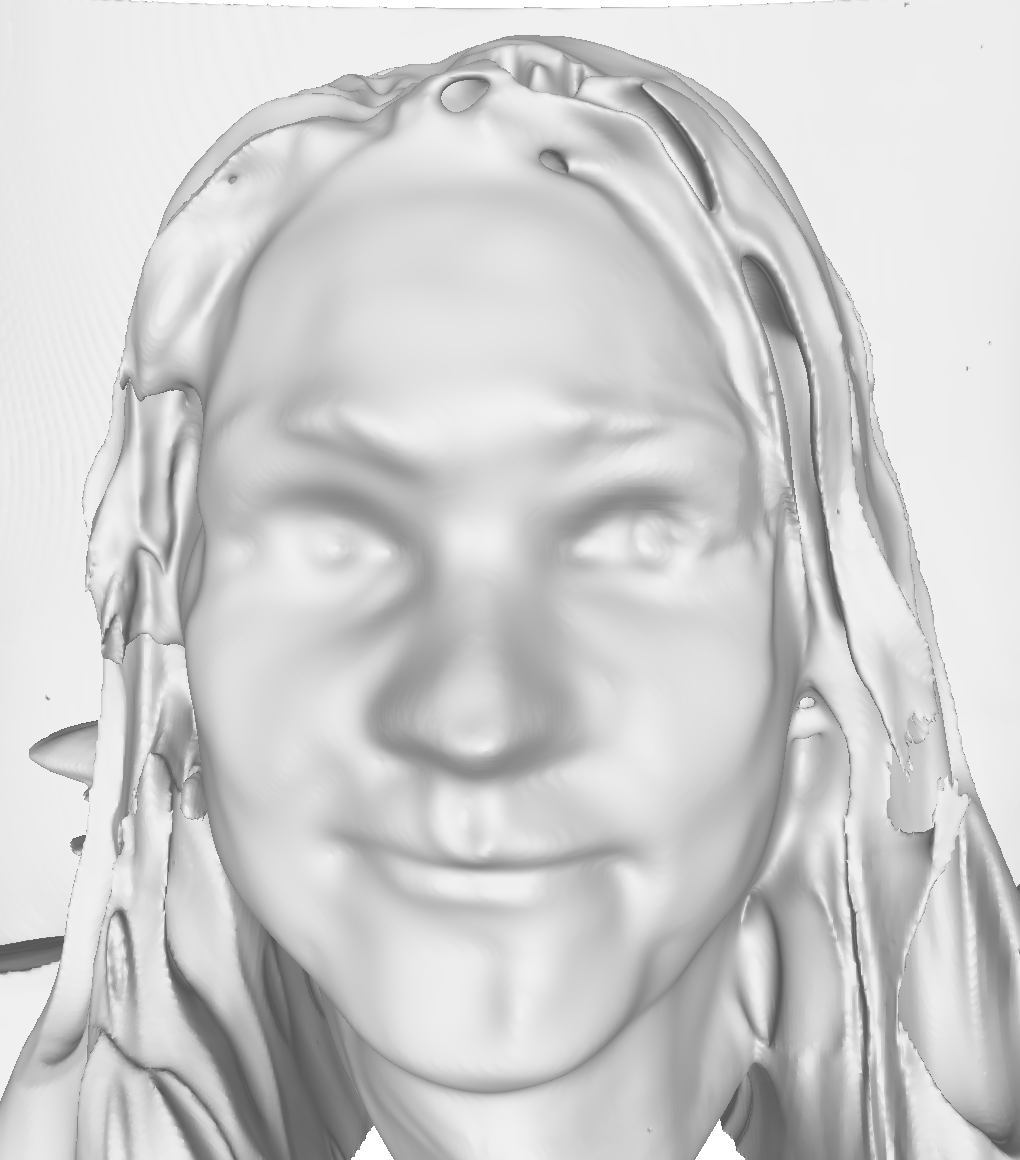}
        & \includegraphics[width=1.7cm,height=1.7cm,keepaspectratio]{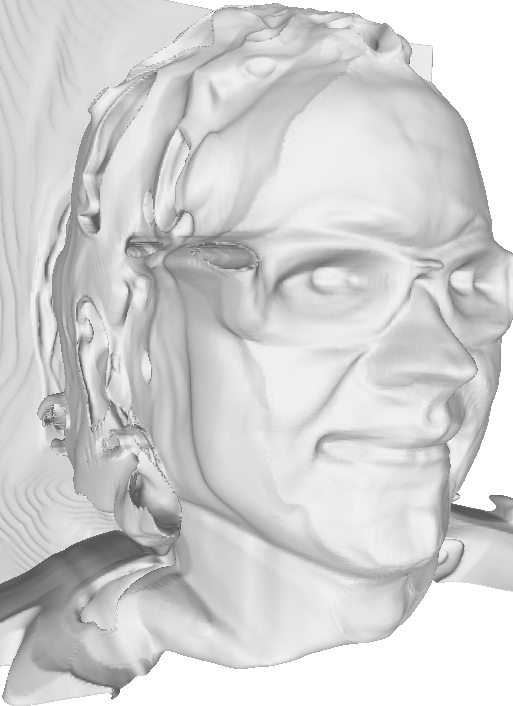}
        & \includegraphics[width=1.7cm,height=1.7cm,keepaspectratio]{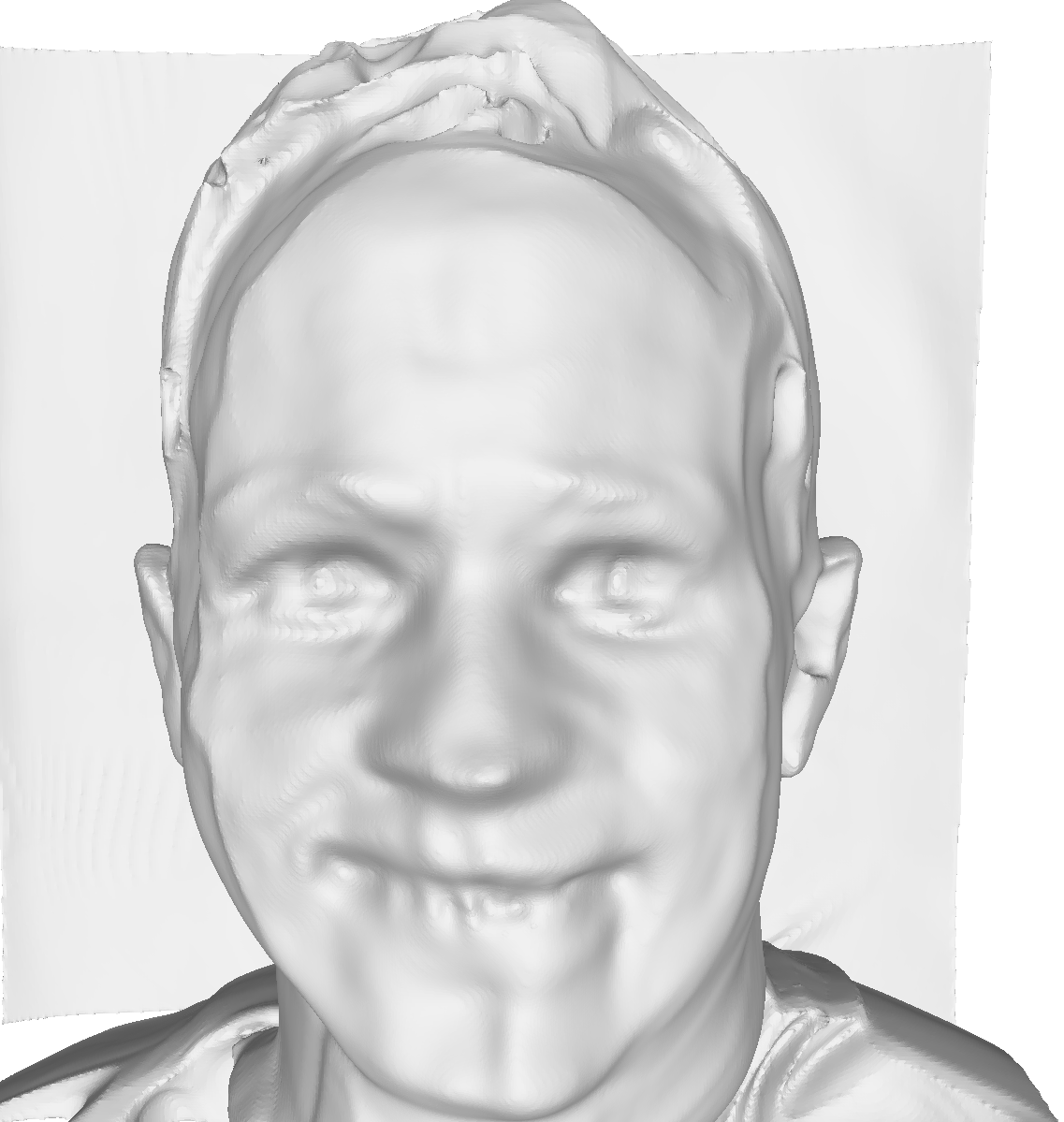} \\
    \end{tabular}
\end{minipage}
    \vspace{-10pt}
    \caption{Sampled images and meshes from EG3D, Style SDF, and our GeoGen approach on FFHQ. GeoGen meshes display smoothness, anatomical accuracy, and detailed facial features. In contrast to EG3D and Style SDF, GeoGen synthesizes finer geometric detail.}
    \label{fig:real_samples}
    \vspace{-5pt}
\end{figure*}

\begin{figure*}[ht]
    \begin{minipage}[t]{0.5\textwidth}
        \centering
        \scriptsize
        \setlength{\tabcolsep}{0pt}
        \begin{tabular}{ccccccc}
            \rotatebox{90}{\parbox[t]{2cm}{\centering EG3D}}\hspace*{5pt}
            & \includegraphics[width=1.7cm,height=1.7cm,keepaspectratio]{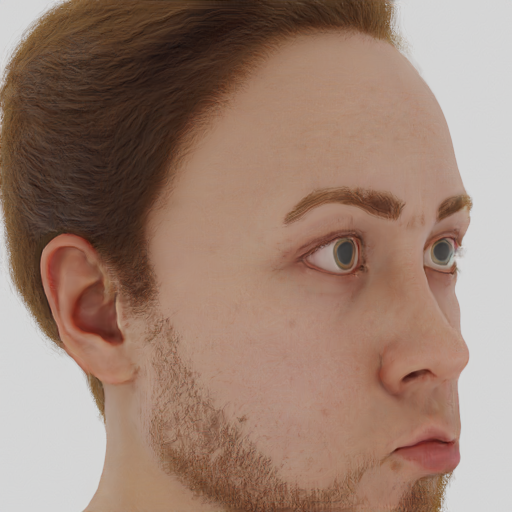}
            & \includegraphics[width=1.7cm,height=1.7cm,keepaspectratio]{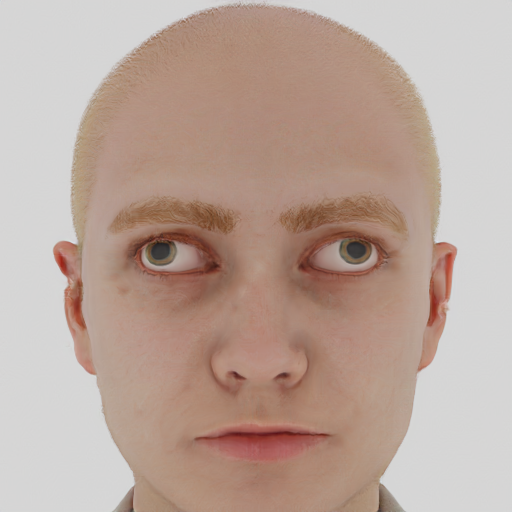}
            & \includegraphics[width=1.7cm,height=1.7cm,keepaspectratio]{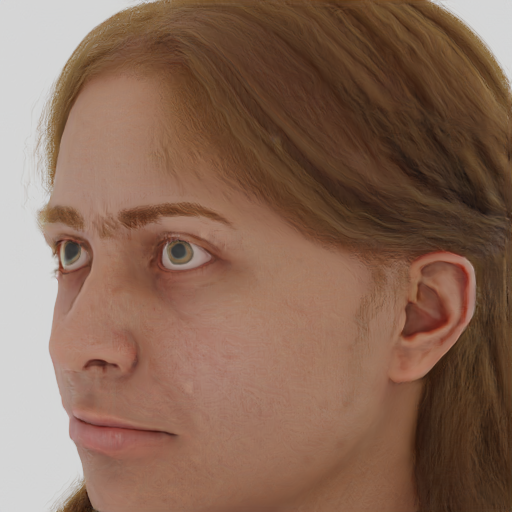}
            & \includegraphics[width=1.7cm,height=1.7cm,keepaspectratio]{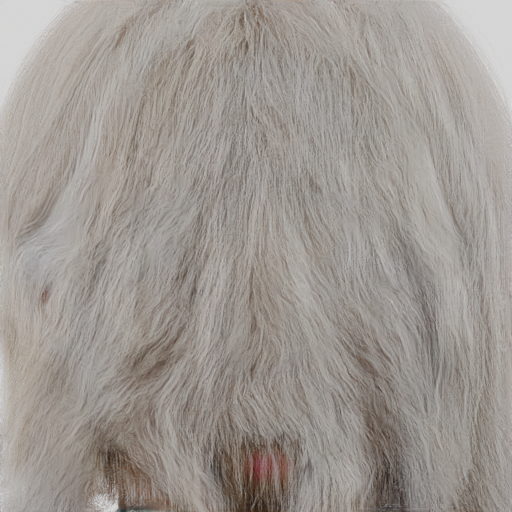}
            & \includegraphics[width=1.7cm,height=1.7cm,keepaspectratio]{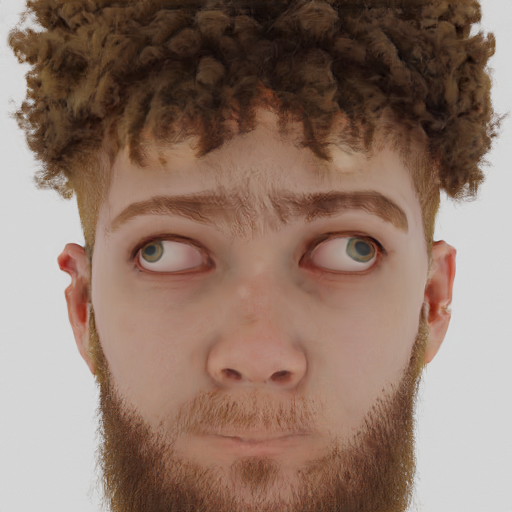} \\
            \rotatebox{90}{\parbox[t]{2cm}{\centering StyleSDF}}\hspace*{5pt}
            & \includegraphics[width=1.7cm,height=1.7cm,keepaspectratio]{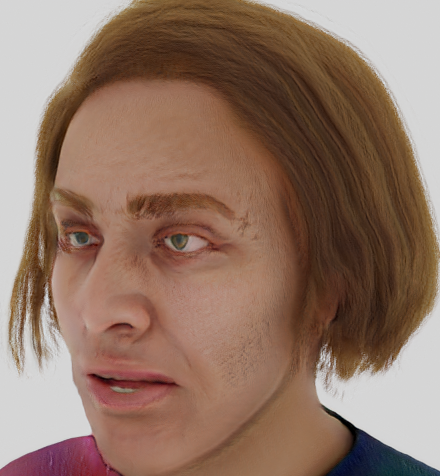}
            & \includegraphics[width=1.7cm,height=1.7cm,keepaspectratio]{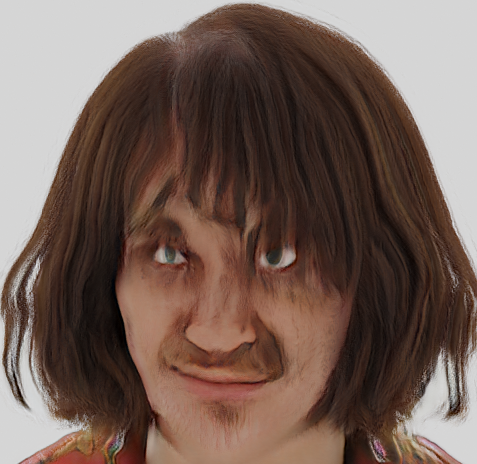}
            & \includegraphics[width=1.7cm,height=1.7cm,keepaspectratio]{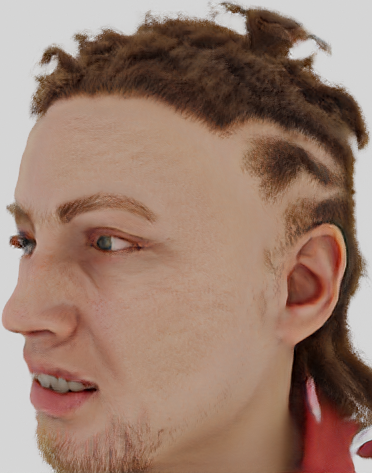}
            & \includegraphics[width=1.7cm,height=1.7cm,keepaspectratio]{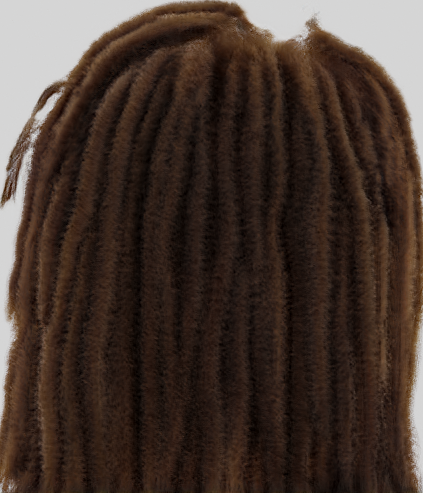}
            & \includegraphics[width=1.7cm,height=1.7cm,keepaspectratio]{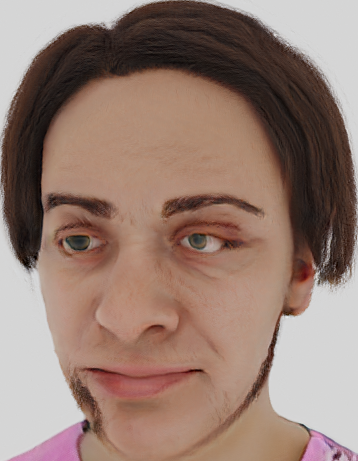} \\
            \rotatebox{90}{\parbox[t]{2cm}{\centering GeoGen}}\hspace*{5pt}
            & \includegraphics[width=1.7cm,height=1.7cm,keepaspectratio]{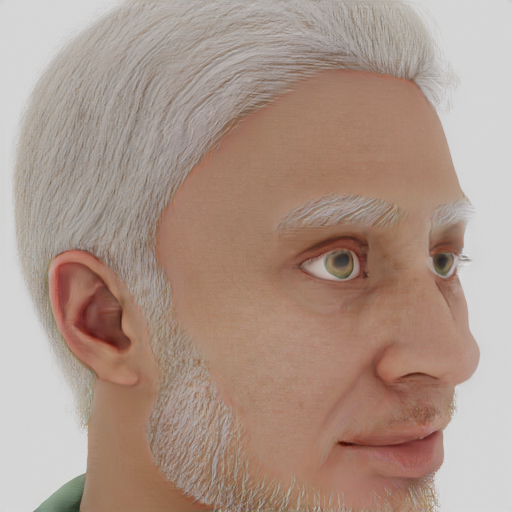}
            & \includegraphics[width=1.7cm,height=1.7cm,keepaspectratio]{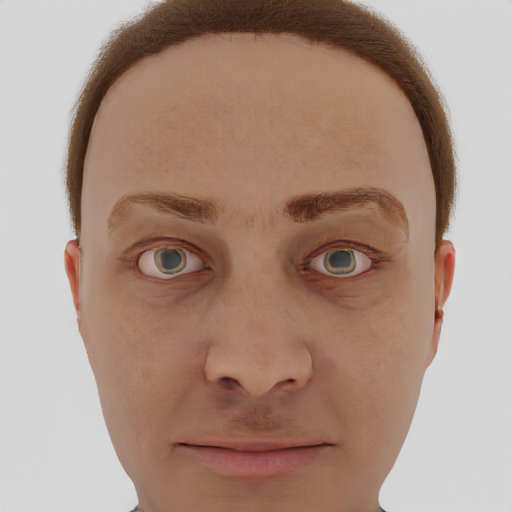}
            & \includegraphics[width=1.7cm,height=1.7cm,keepaspectratio]{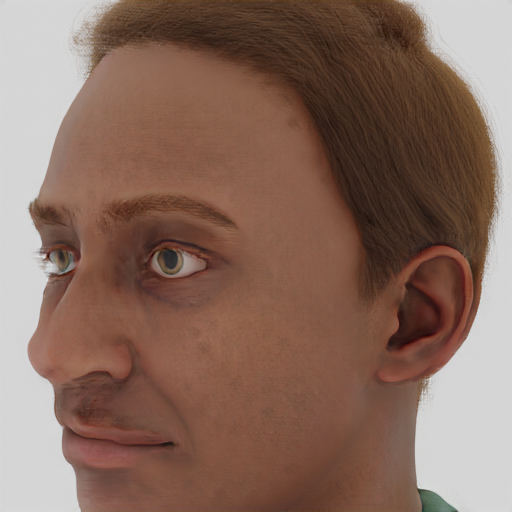}
            & \includegraphics[width=1.7cm,height=1.7cm,keepaspectratio]{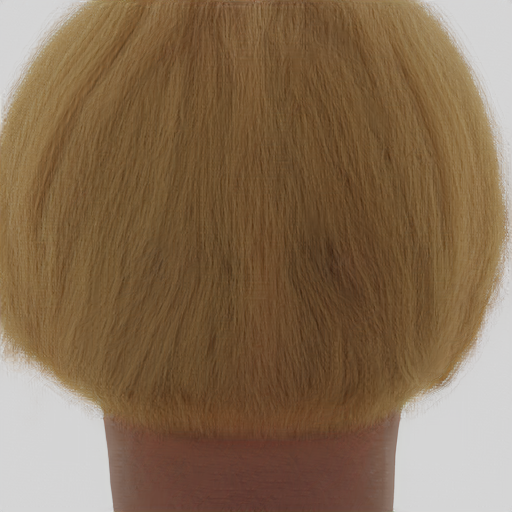}
            & \includegraphics[width=1.7cm,height=1.7cm,keepaspectratio]{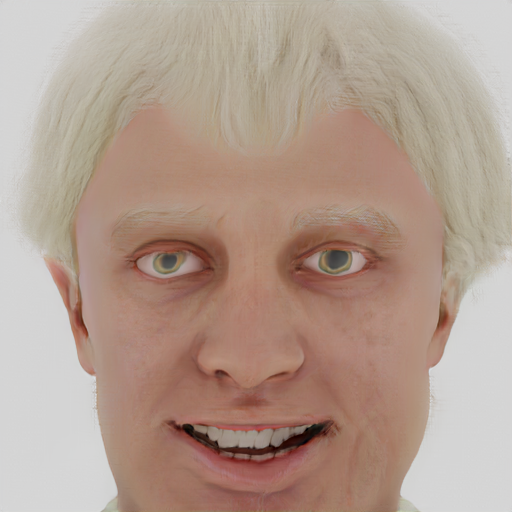} \\
        \end{tabular}
    \end{minipage}%
    \hfill
    \begin{minipage}[t]{0.5\textwidth}
    \centering
    \scriptsize
    \setlength{\tabcolsep}{0pt}
    \begin{tabular}{ccccccc}
        \phantom{\rotatebox{90}{\parbox[t]{2cm}{\centering EG3D}}}\hspace*{5pt}
        & \includegraphics[width=1.7cm,height=1.7cm,keepaspectratio]{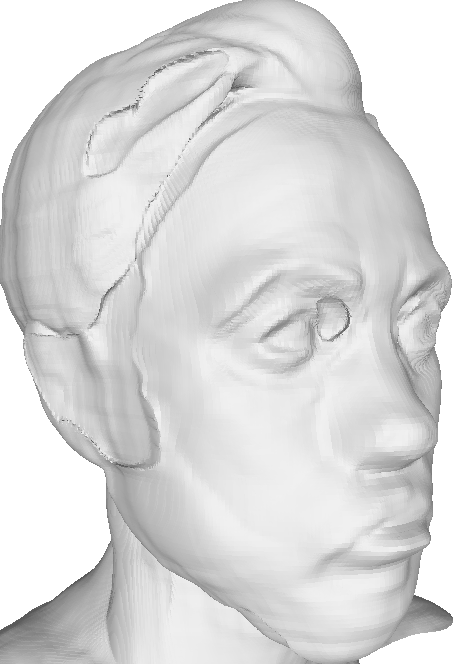}
        & \includegraphics[width=1.7cm,height=1.7cm,keepaspectratio]{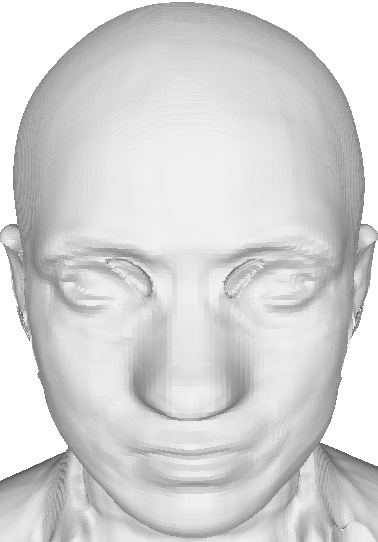}
        & \includegraphics[width=1.7cm,height=1.7cm,keepaspectratio]{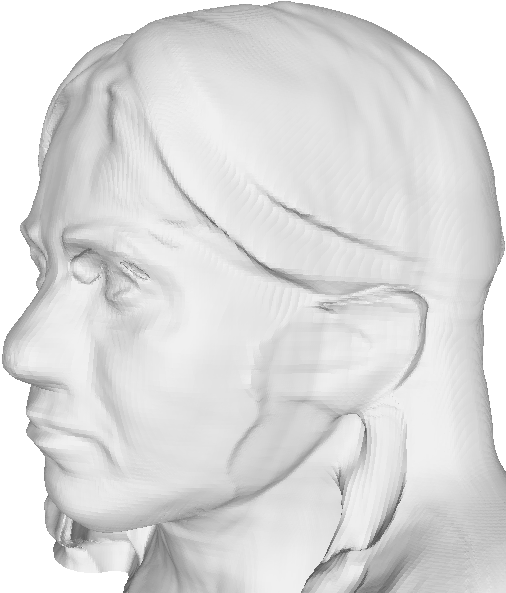}
        & \includegraphics[width=1.7cm,height=1.7cm,keepaspectratio]{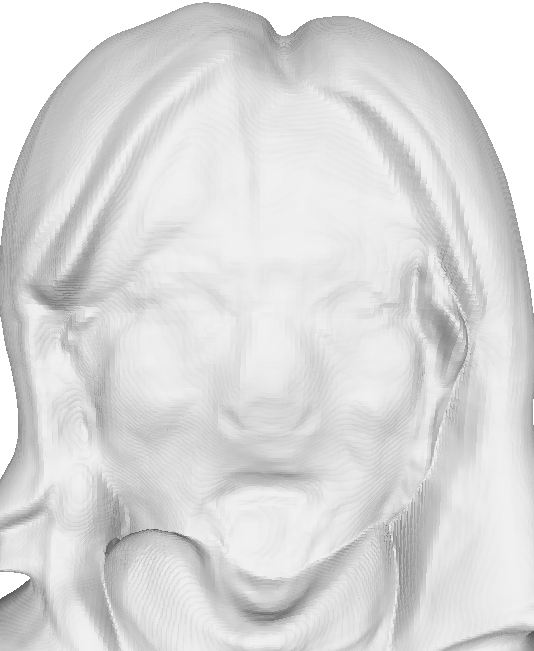}
        & \includegraphics[width=1.7cm,height=1.7cm,keepaspectratio]{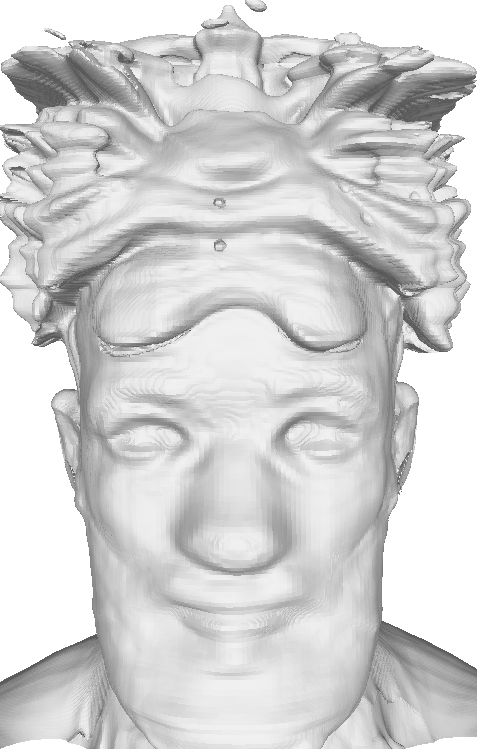} \\
        \phantom{\rotatebox{90}{\parbox[t]{2cm}{\centering GeoGen}}}\hspace*{5pt}
        & \includegraphics[width=1.7cm,height=1.7cm,keepaspectratio]{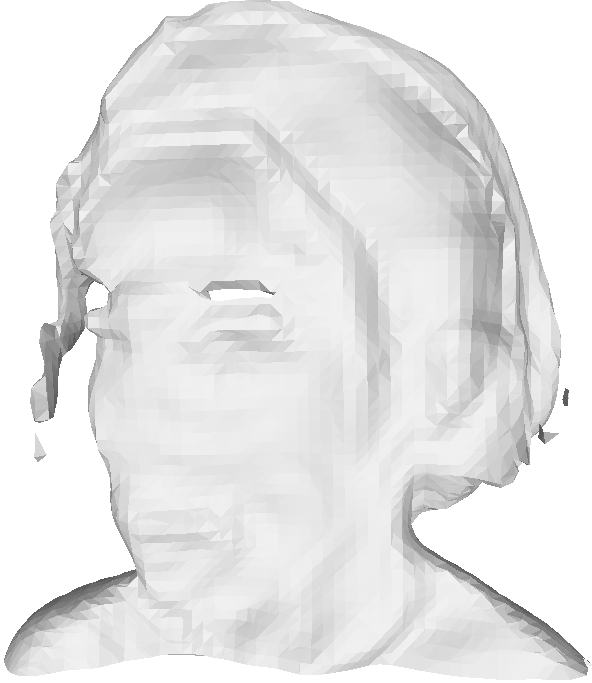}
        & \includegraphics[width=1.7cm,height=1.7cm,keepaspectratio]{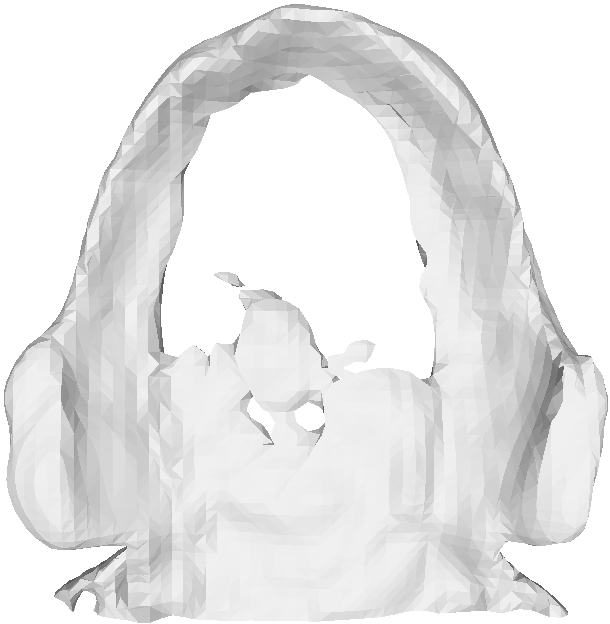}
        & \includegraphics[width=1.7cm,height=1.7cm,keepaspectratio]{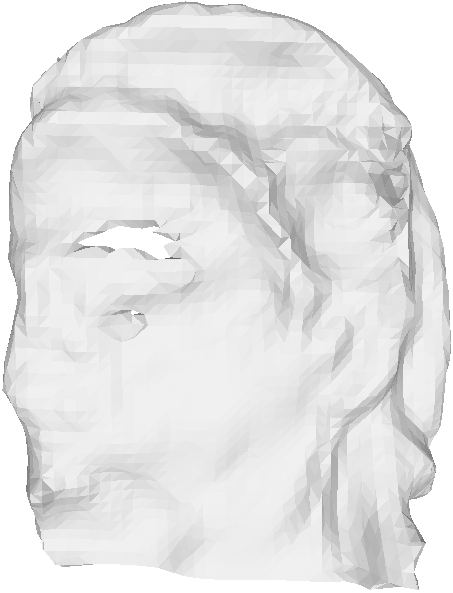}
        & \includegraphics[width=1.7cm,height=1.7cm,keepaspectratio]{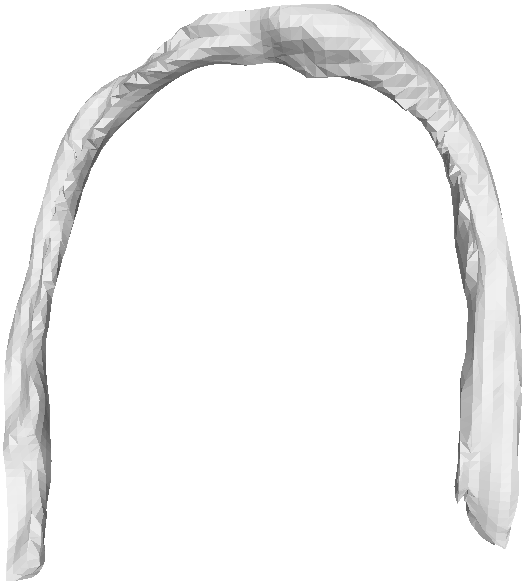}
        & \includegraphics[width=1.7cm,height=1.7cm,keepaspectratio]{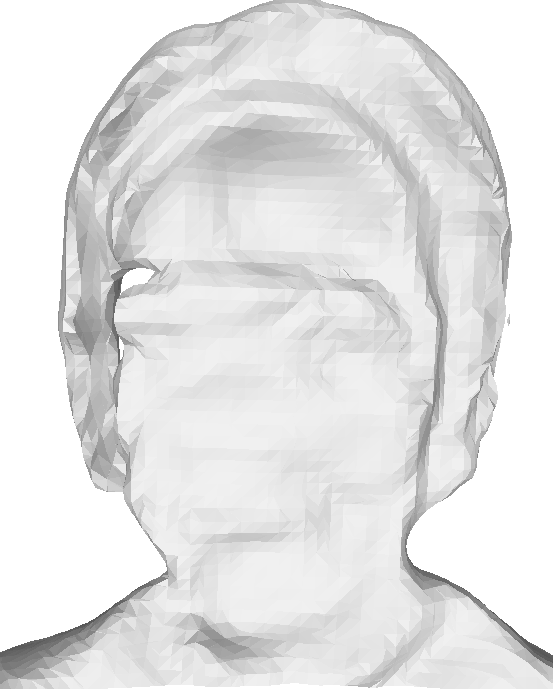} \\
        \phantom{\rotatebox{90}{\parbox[t]{2cm}{\centering GeoGen w/o SDF\&DepthLoss}}}\hspace*{5pt}
        & \includegraphics[width=1.7cm,height=1.7cm,keepaspectratio]{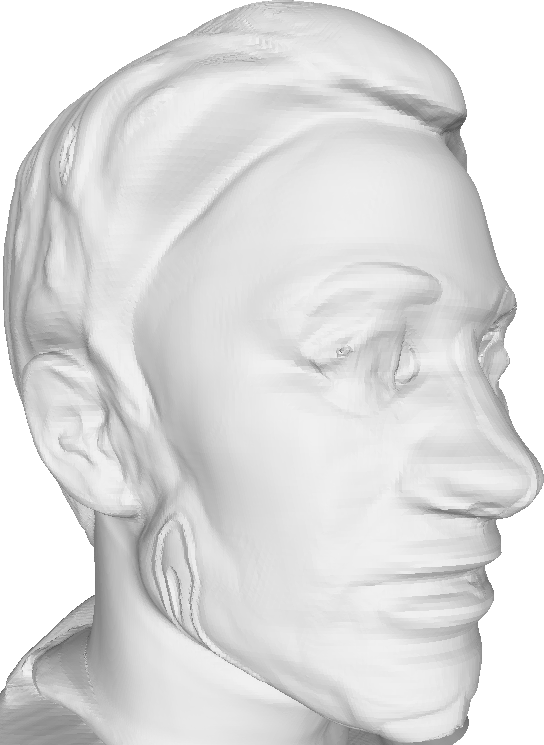}
        & \includegraphics[width=1.7cm,height=1.7cm,keepaspectratio]{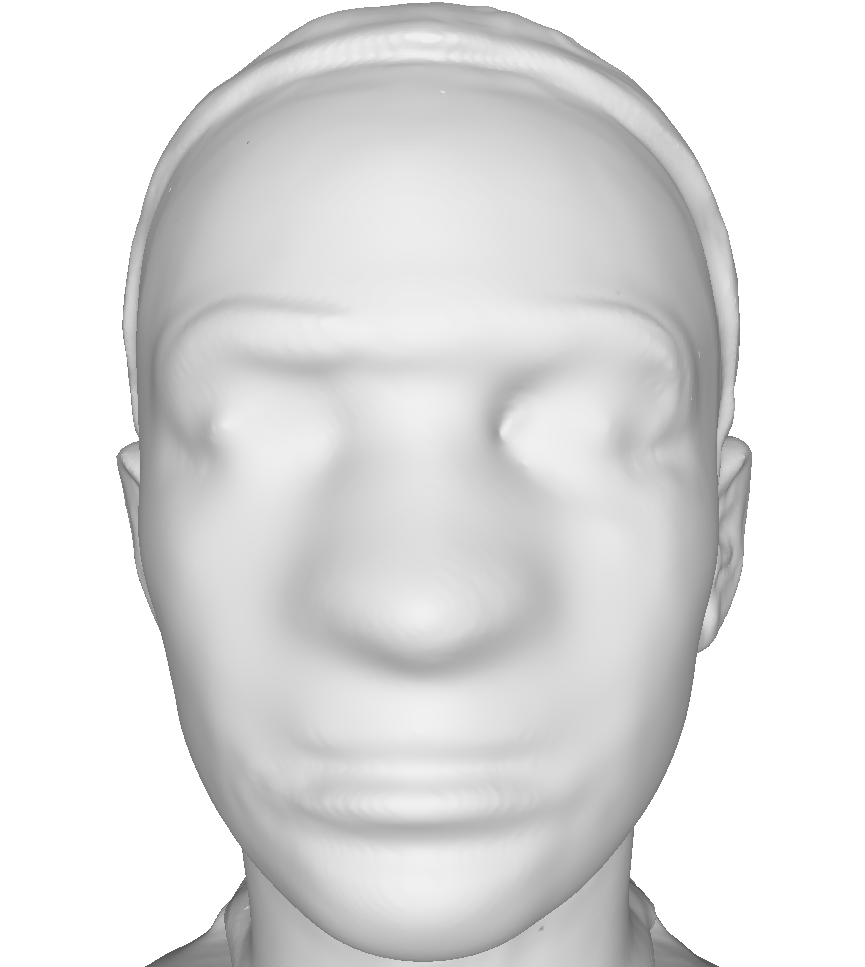}
        & \includegraphics[width=1.7cm,height=1.7cm,keepaspectratio]{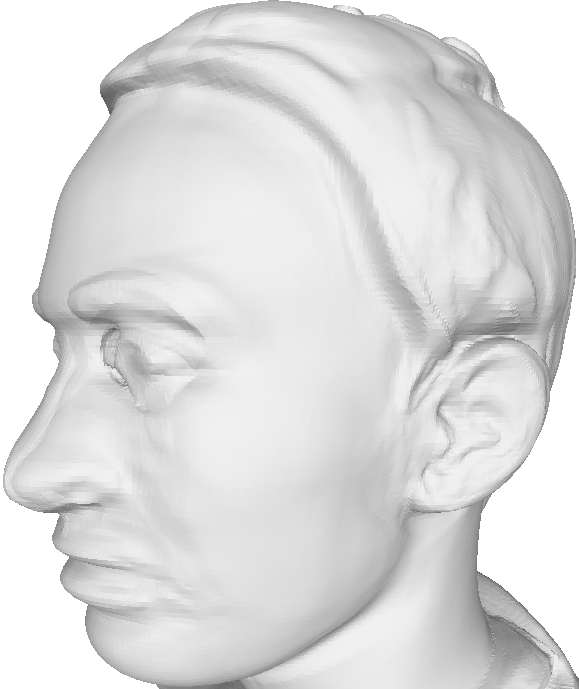}
        & \includegraphics[width=1.7cm,height=1.7cm,keepaspectratio]{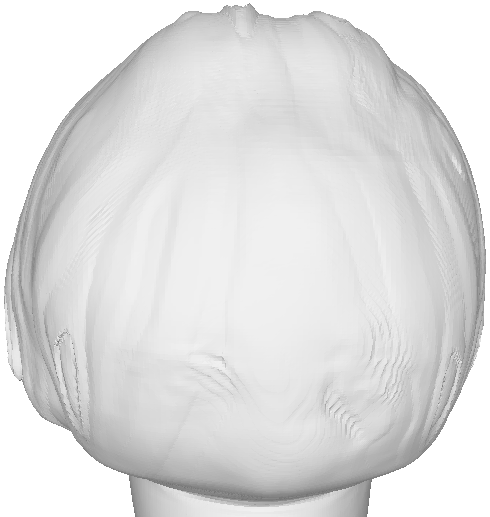}
        & \includegraphics[width=1.7cm,height=1.7cm,keepaspectratio]{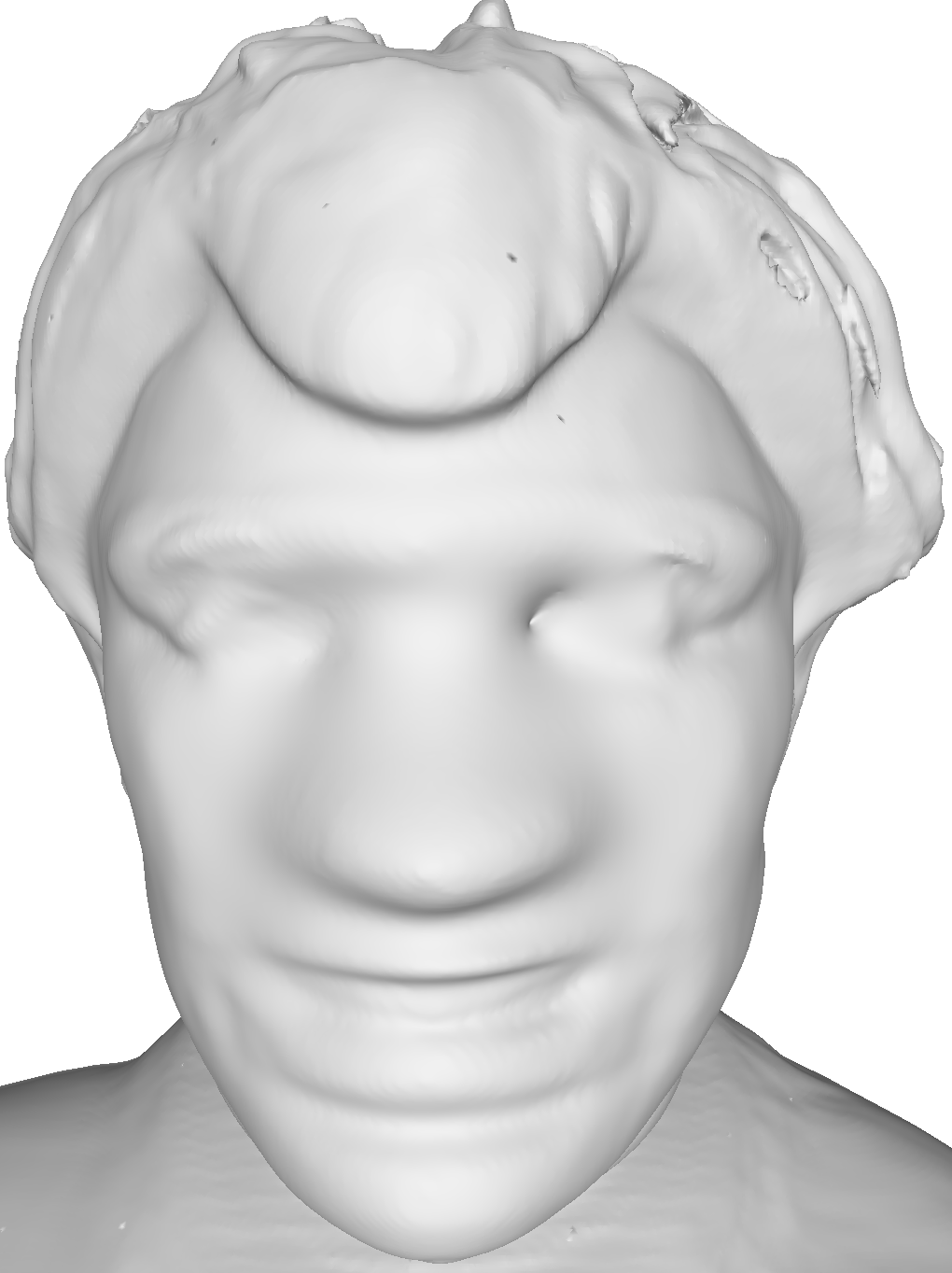} \\

    \end{tabular}
\end{minipage}
    \vspace{-10pt}
    \caption{Sampled images and meshes from EG3D, StyleSDF, and our GeoGen approach trained on our synthetic human head dataset. GeoGen results in fewer overt visual artefacts and more faithfully captures the backs of objects (\eg see second last column). While the 2D images from the competing methods look plausible, the underlying 3D mesh is not always consistent. }
    \label{fig:real_samples_2}
    \vspace{-10pt}
\end{figure*}

\subsection{Datasets} 
We perform experiments on Flickr-Faces-HQ (FFHQ)~\cite{karras2019style}, ShapeNet Cars~\cite{chang2015shapenet}, and our synthetic human dataset described previously. 
Each provide distinct, valuable resources for training and evaluating 3D-aware generative models. 
The FFHQ dataset consists of high-quality real 2D face images. 
It contains over 70,000 1024$\times$1024 resolution images.  
ShapeNet Cars provides images for a variety of car models imaged from different viewpoints.
The dataset we used for training contains 2,100 different car instances, each with 20 images from different viewpoints.

\begin{figure*}[!htbp]
\centering

\newlength{\imgwidth}
\setlength{\imgwidth}{0.12\textwidth}

\newcolumntype{M}[1]{>{\centering\arraybackslash}m{#1}}

\begin{tabular}{@{}m{1em} *{6}{M{\imgwidth}}@{}}

& \textbf{Input image} & \textbf{Pseudo Ground-truth} & \textbf{EG3D} & \textbf{GeoGen} & \textbf{EG3D Zoom} & \textbf{GeoGen Zoom} \\
\midrule

\rotatebox[origin=c]{90}{0\(^{\circ}\)}
& \includegraphics[width=\imgwidth,height=2cm,keepaspectratio]{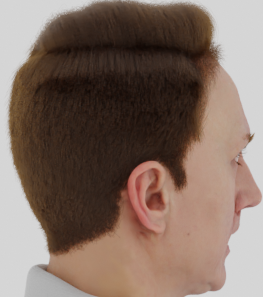}
& \includegraphics[width=\imgwidth,height=2cm,keepaspectratio]{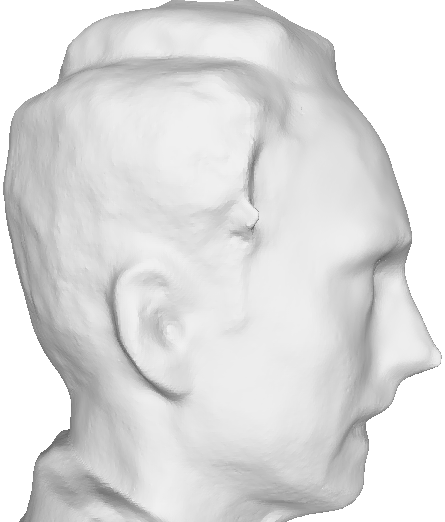}
& \includegraphics[width=\imgwidth,height=2cm,keepaspectratio]{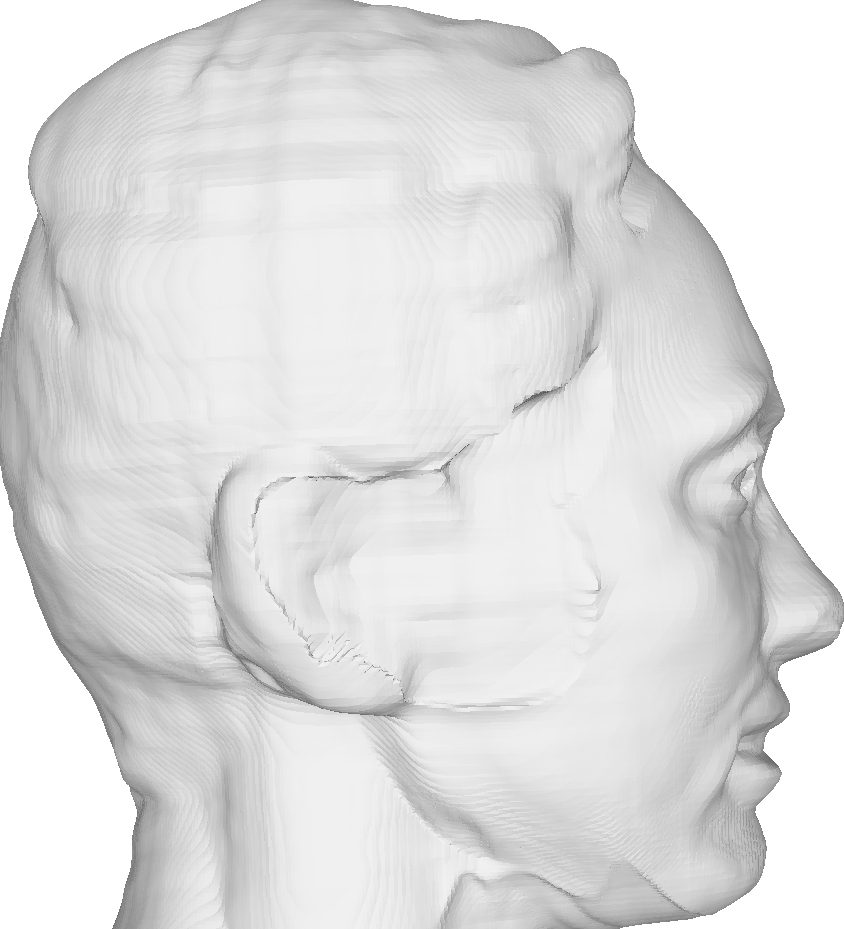}
& \includegraphics[width=\imgwidth,height=2cm,keepaspectratio]{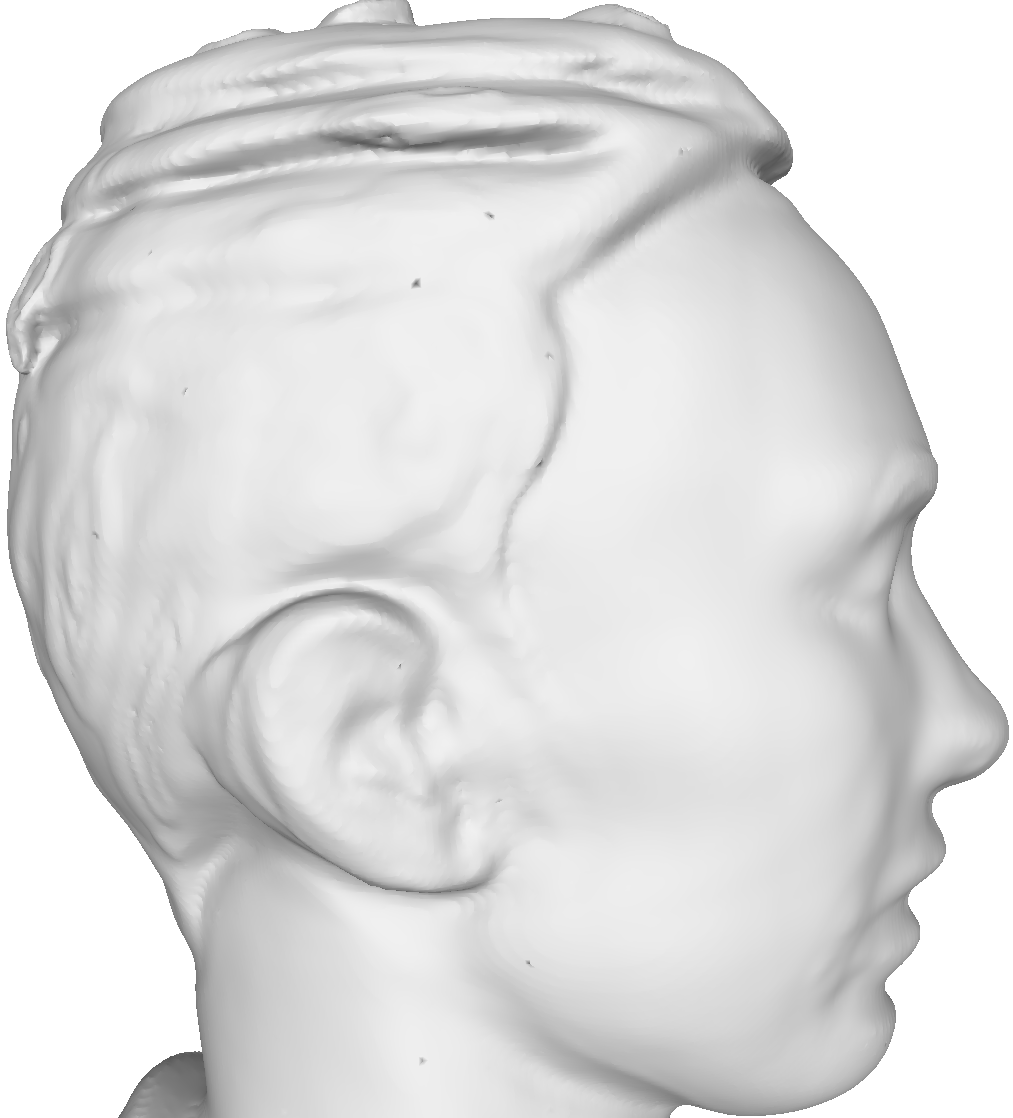}
& \includegraphics[width=\imgwidth,height=2cm,keepaspectratio]{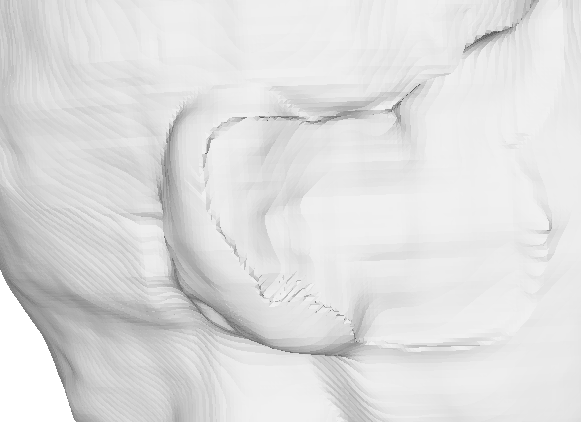}
& \includegraphics[width=\imgwidth,height=2cm,keepaspectratio]{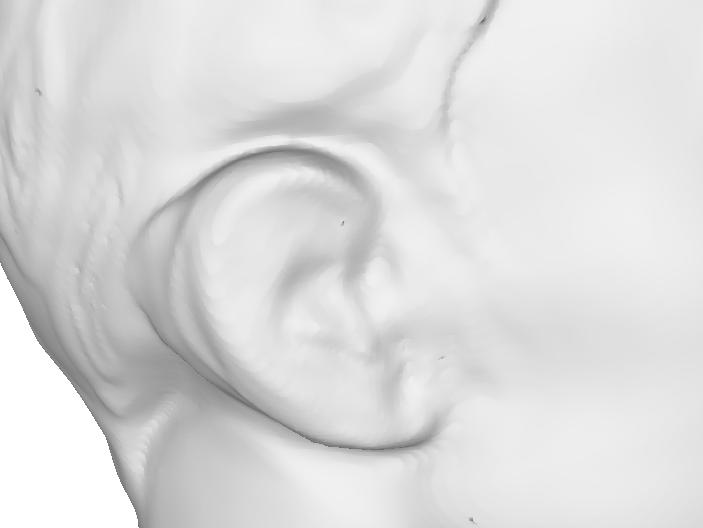} \\
\midrule
\rotatebox[origin=c]{90}{90\(^{\circ}\)}
& \includegraphics[width=\imgwidth,height=2cm,keepaspectratio]{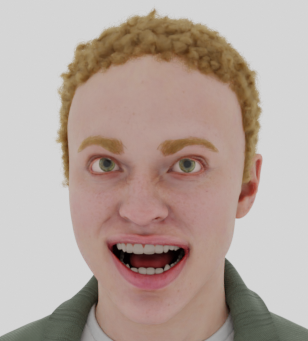}
& \includegraphics[width=\imgwidth,height=2cm,keepaspectratio]{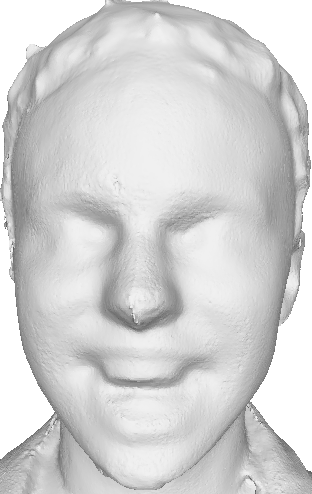}
& \includegraphics[width=\imgwidth,height=2cm,keepaspectratio]{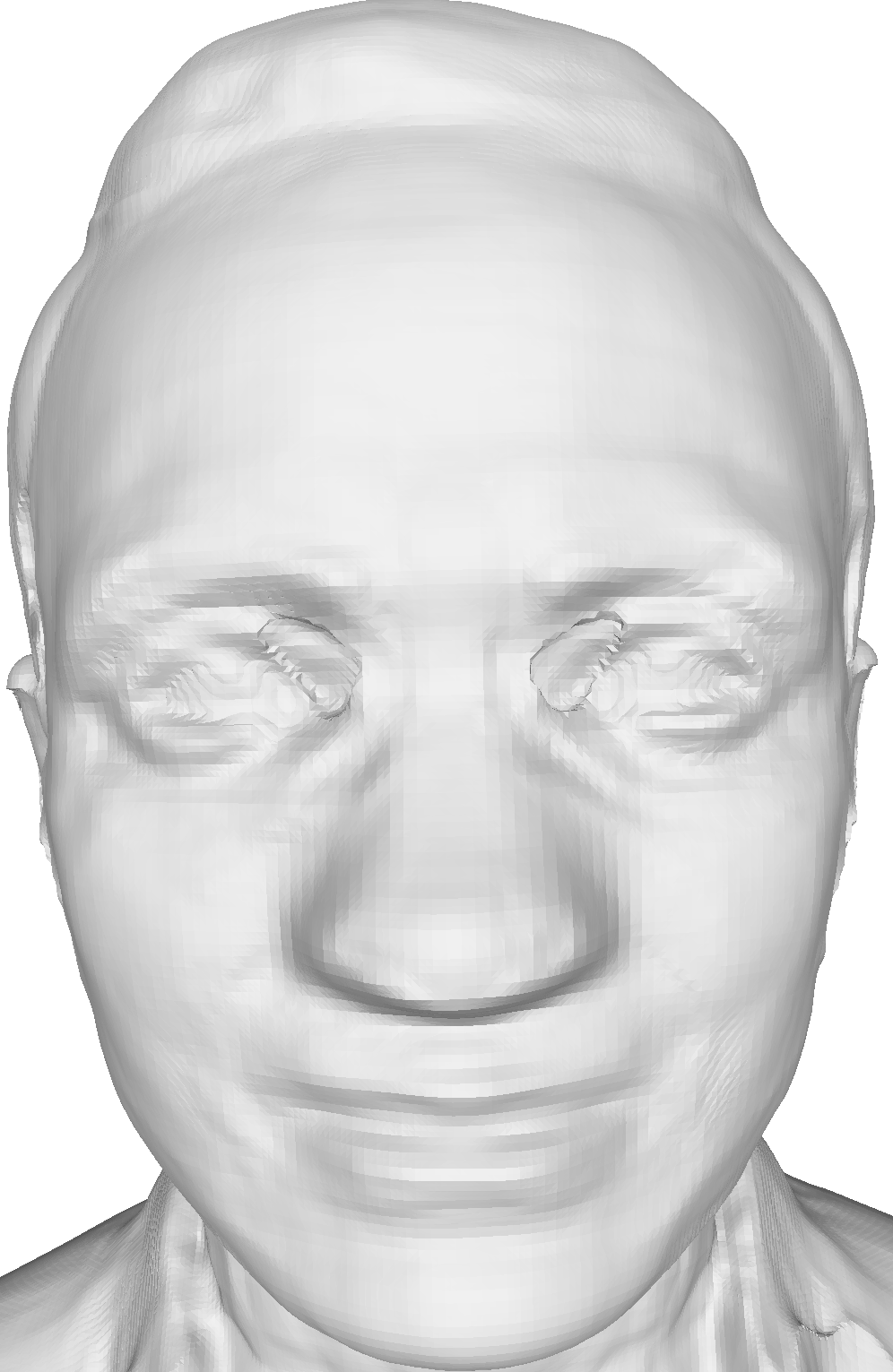}
& \includegraphics[width=\imgwidth,height=2cm,keepaspectratio]{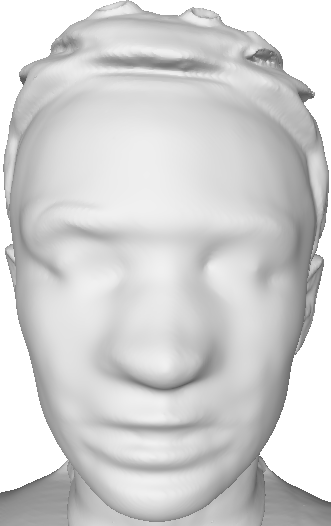}
& \includegraphics[width=\imgwidth,height=2cm,keepaspectratio]{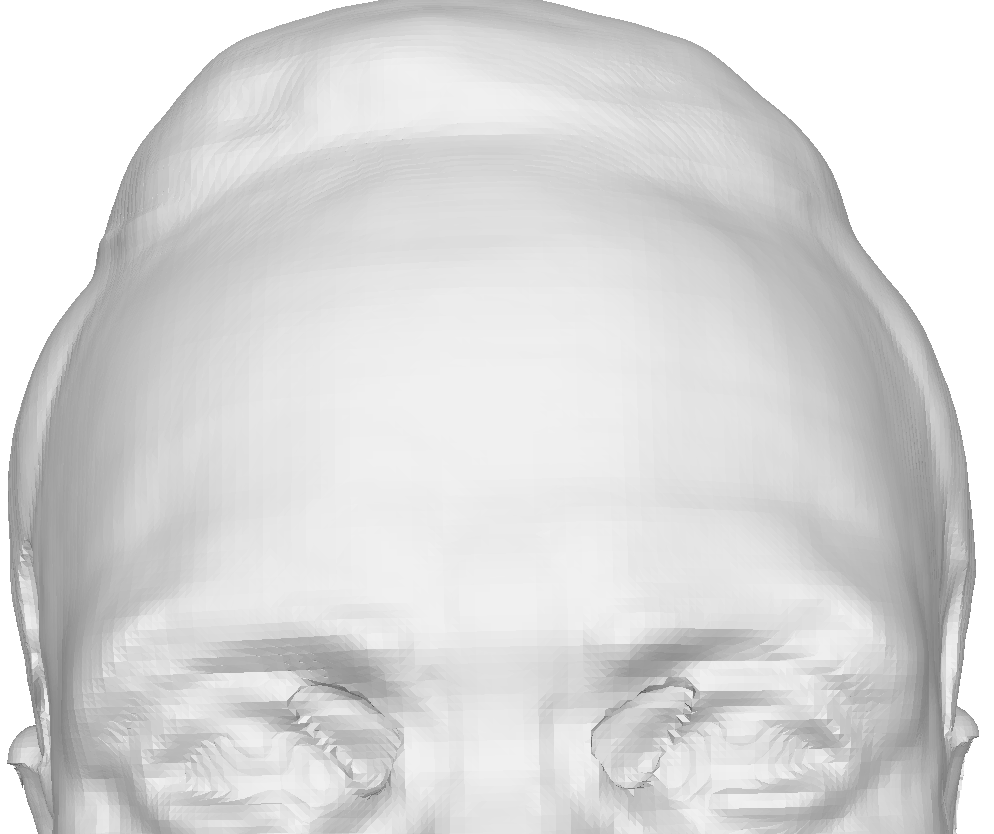}
& \includegraphics[width=\imgwidth,height=2cm,keepaspectratio]{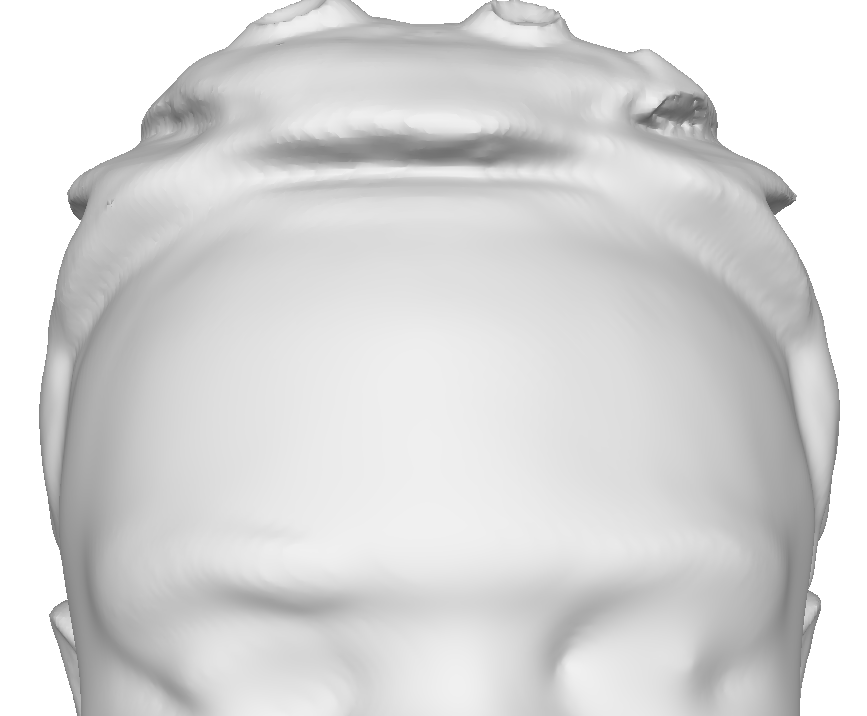}  \\
\bottomrule

\rotatebox[origin=c]{90}{270\(^{\circ}\)}
& \includegraphics[width=\imgwidth,height=2cm,keepaspectratio]{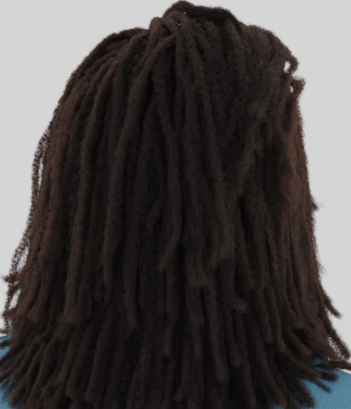}
& \includegraphics[width=\imgwidth,height=2cm,keepaspectratio]{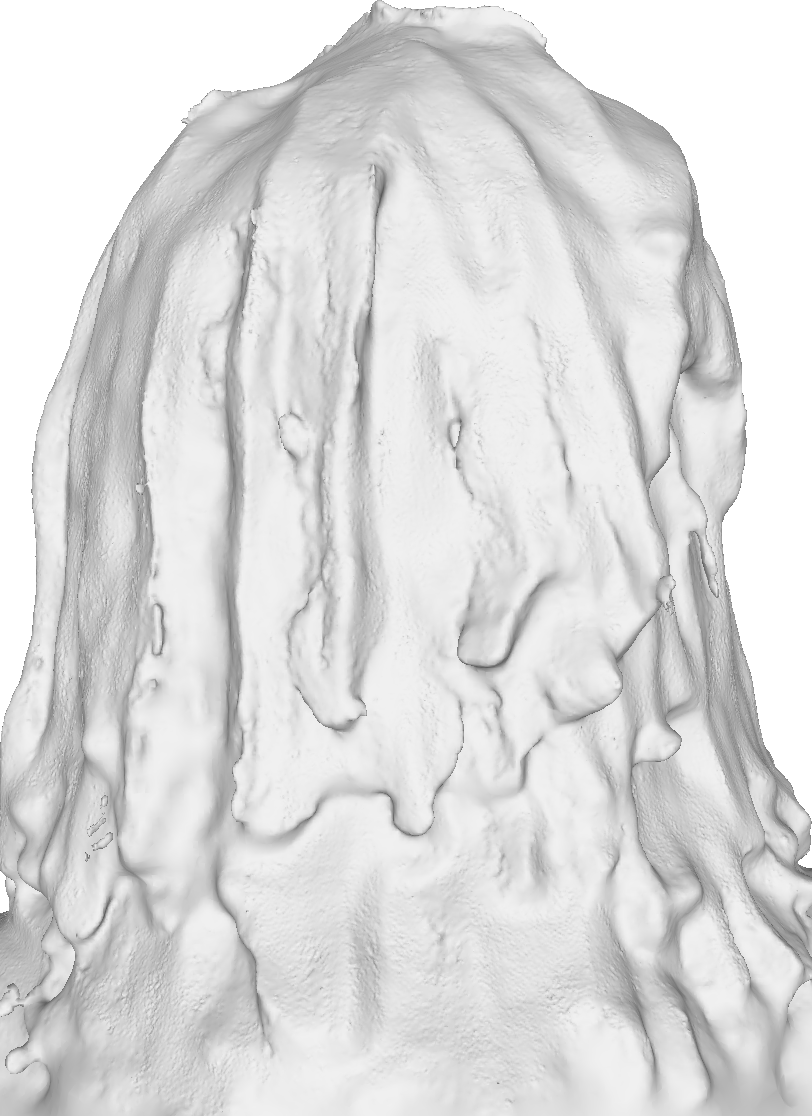}
& \includegraphics[width=\imgwidth,height=2cm,keepaspectratio]{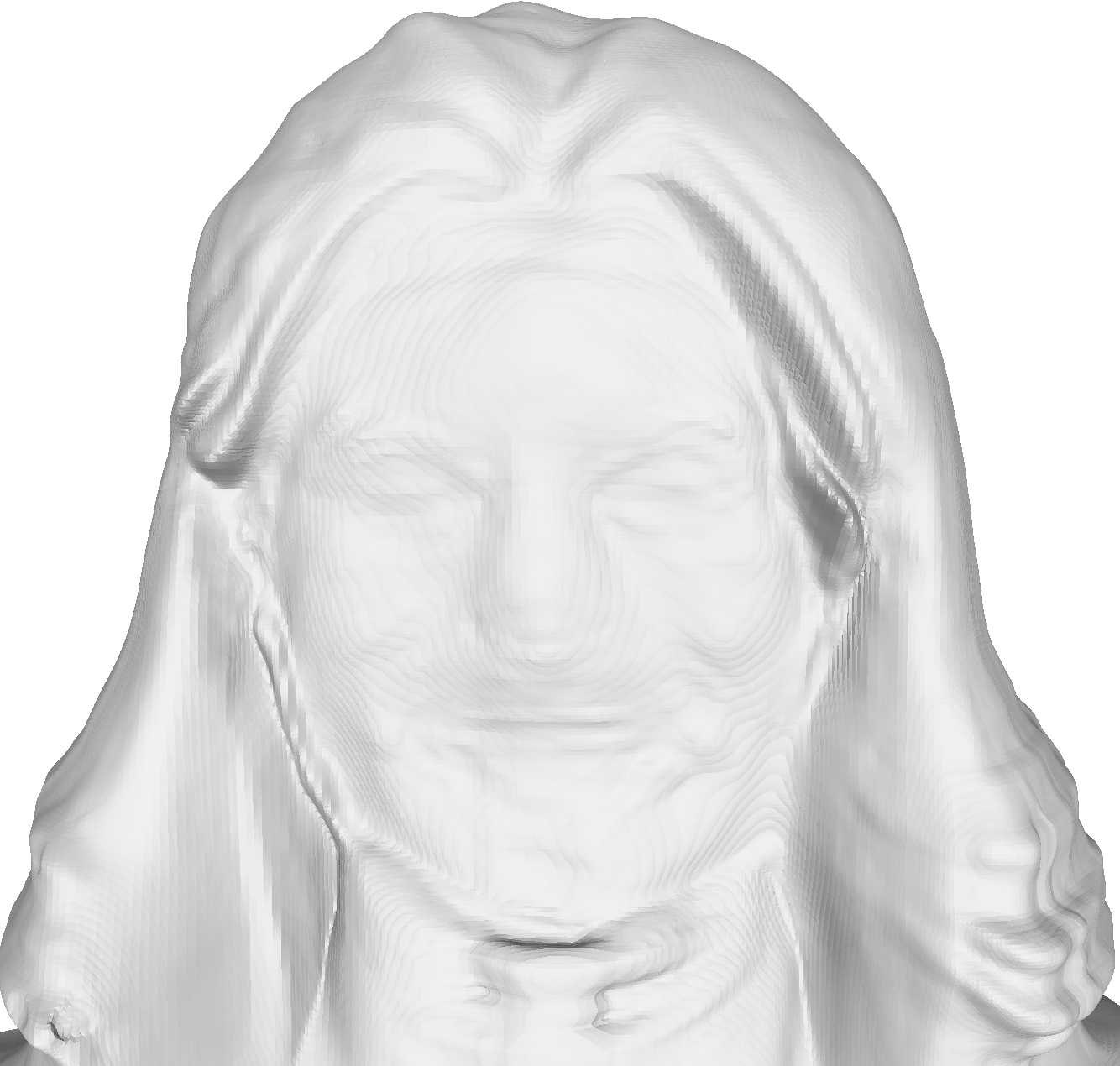}
& \includegraphics[width=\imgwidth,height=2cm,keepaspectratio]{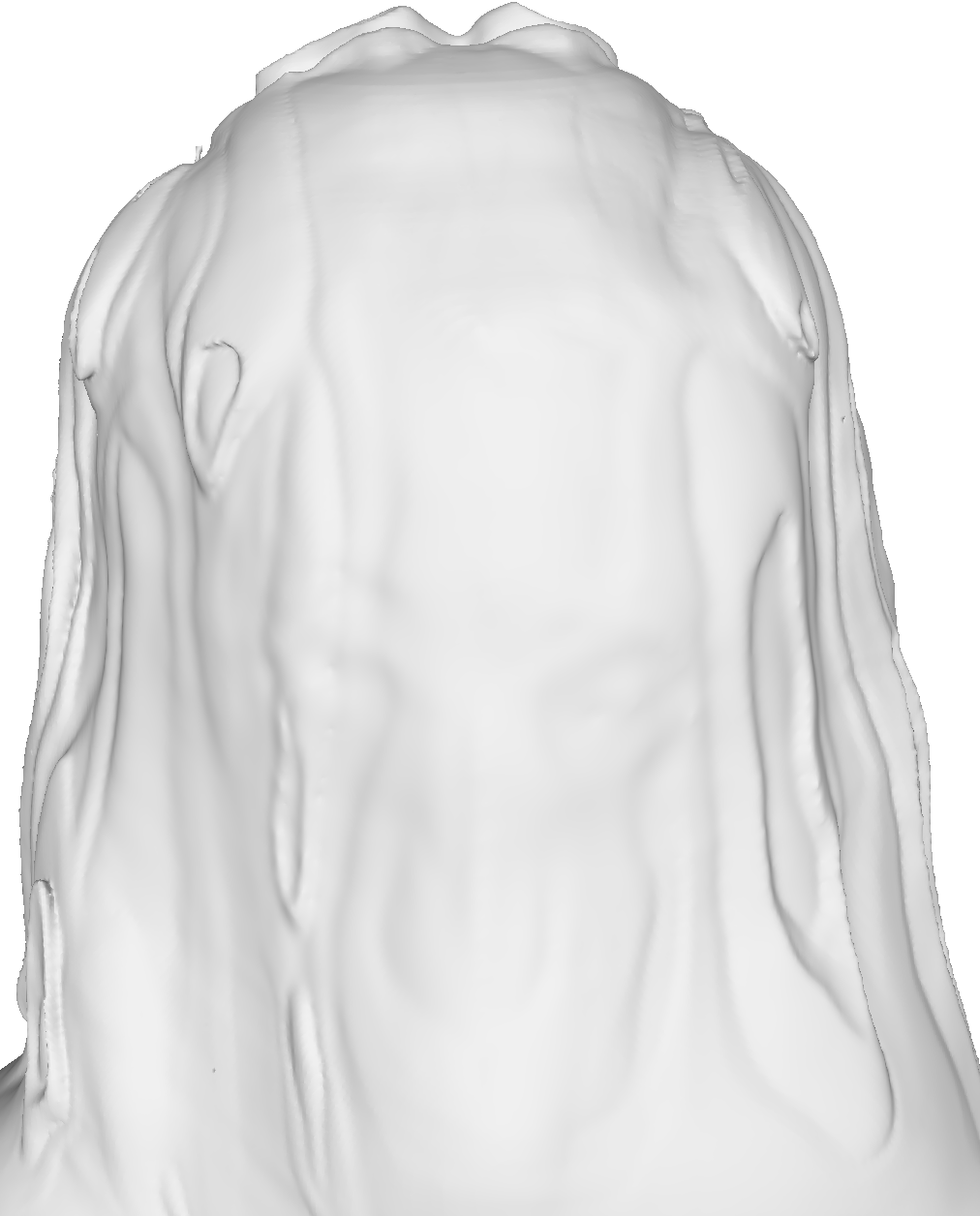}
& \includegraphics[width=\imgwidth,height=2cm,keepaspectratio]{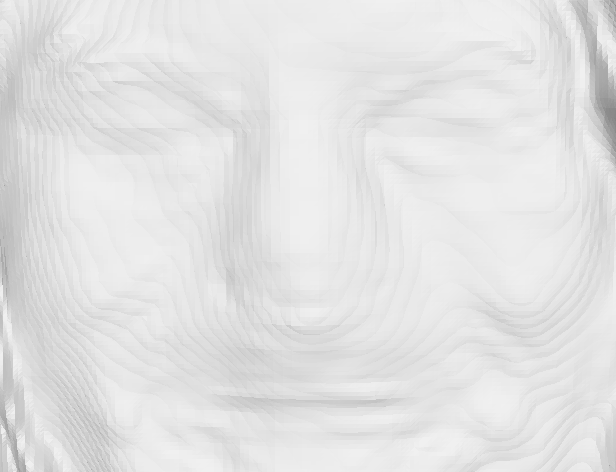}
& \includegraphics[width=\imgwidth,height=2cm,keepaspectratio]{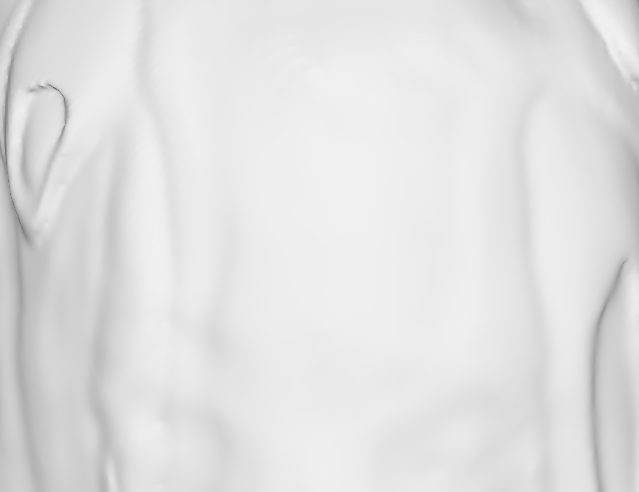}  \\
\bottomrule

\end{tabular}

\caption{Inversion Results for EG3D and GeoGen Models: The figure presents a comparison at 0\(^{\circ}\), 90\(^{\circ}\), and 270\(^{\circ}\) angles to highlight variations in the reconstruction of facial features by the two models.
}
\label{fig:synthetics_inversion}
\end{figure*}

\begin{figure*}[t]
    \begin{minipage}[t]{0.5\textwidth}
        \centering
        \footnotesize
        \setlength{\tabcolsep}{0pt}
        \begin{tabular}{ccccccc}
            & Yaw: 0° & Yaw: 45° & Yaw: 90° & Yaw: 135° & Yaw: 270° \\
            \rotatebox{90}{\parbox[t]{2cm}{\hspace*{\fill}EG3D\hspace*{\fill}}}\hspace*{5pt}
            &  \includegraphics[width=1.6cm,height=1.6cm,keepaspectratio]{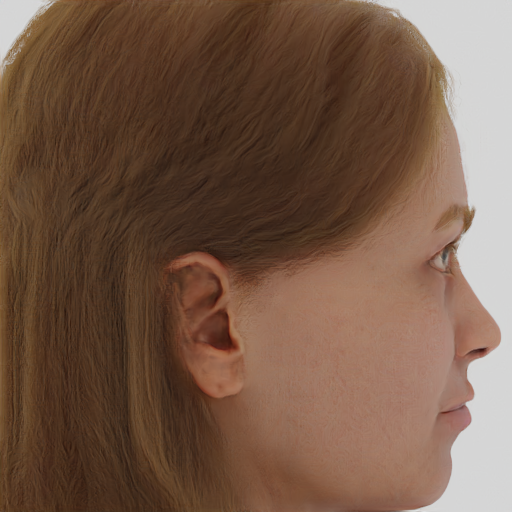}
            &  \includegraphics[width=1.6cm,height=1.6cm,keepaspectratio]{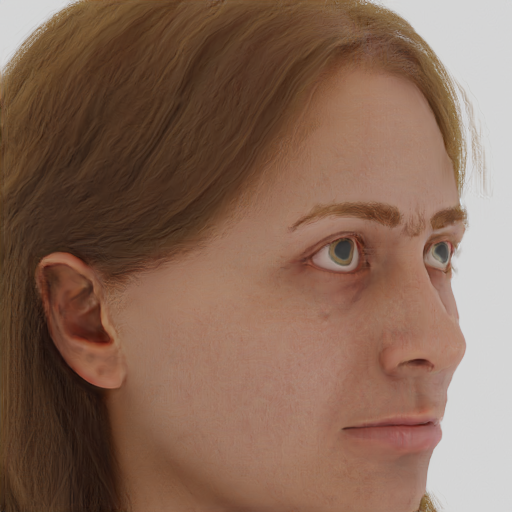}
            &  \includegraphics[width=1.6cm,height=1.6cm,keepaspectratio]{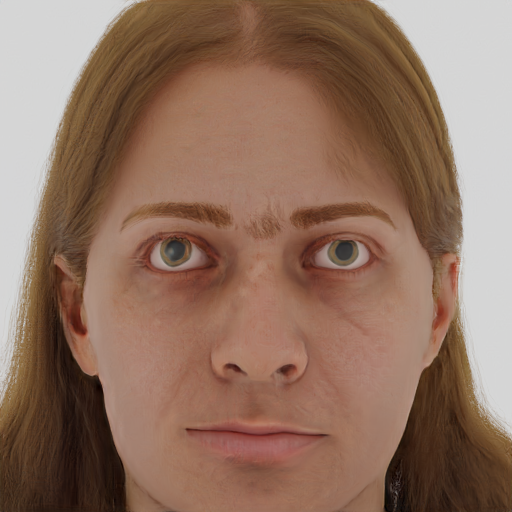}
            &  \includegraphics[width=1.6cm,height=1.6cm,keepaspectratio]{images/supp/syn_meshes/eg3d_syn/seed2/angle135.png}
            &  \includegraphics[width=1.6cm,height=1.6cm,keepaspectratio]{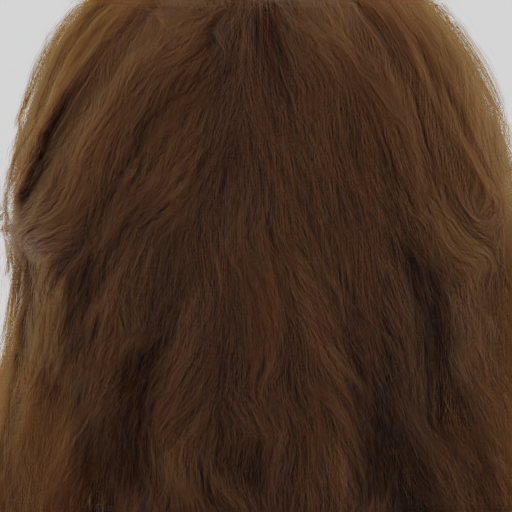} \\
           \rotatebox{90}{\parbox[t]{2cm}{\centering GeoGen w/o SDF DL}}\hspace*{5pt}
            &  \includegraphics[width=1.6cm,height=1.6cm,keepaspectratio]{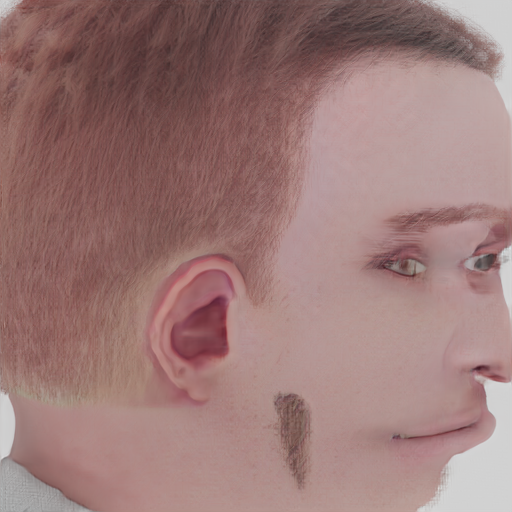}
            &  \includegraphics[width=1.6cm,height=1.6cm,keepaspectratio]{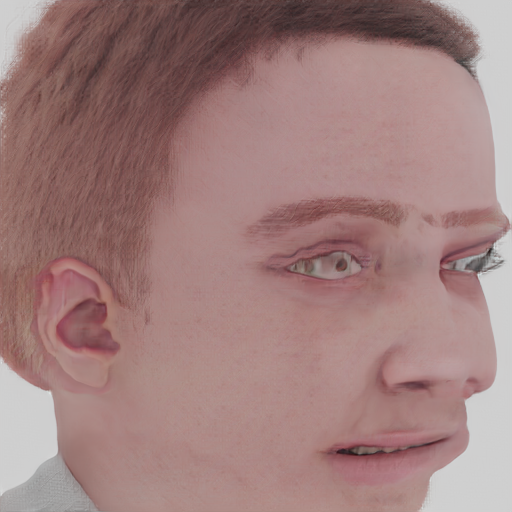}
            &  \includegraphics[width=1.6cm,height=1.6cm,keepaspectratio]{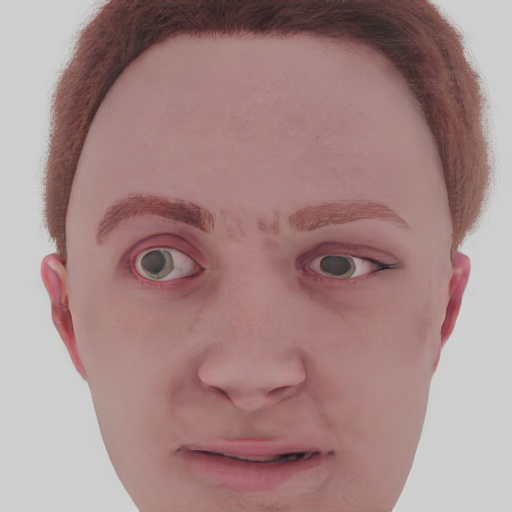}
            &  \includegraphics[width=1.6cm,height=1.6cm,keepaspectratio]{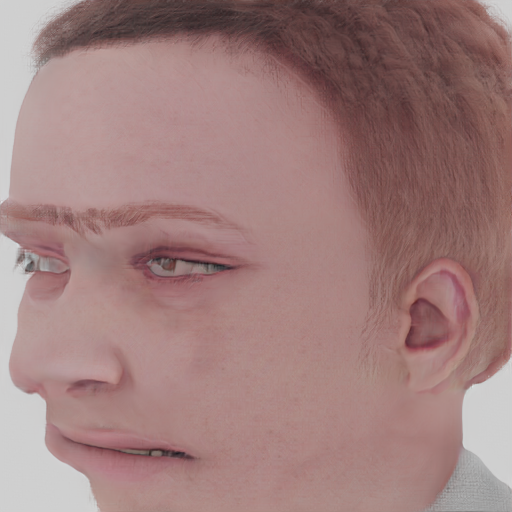}
            &  \includegraphics[width=1.6cm,height=1.6cm,keepaspectratio]{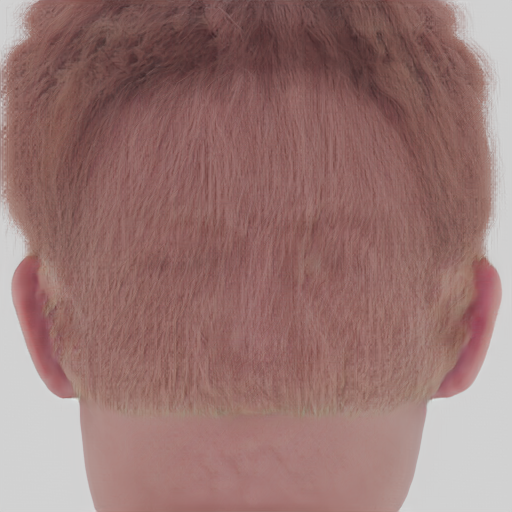} \\
            \rotatebox{90}{\parbox[t]{2cm}{\centering GeoGen}}\hspace*{5pt}
            &  \includegraphics[width=1.6cm,height=1.6cm,keepaspectratio]{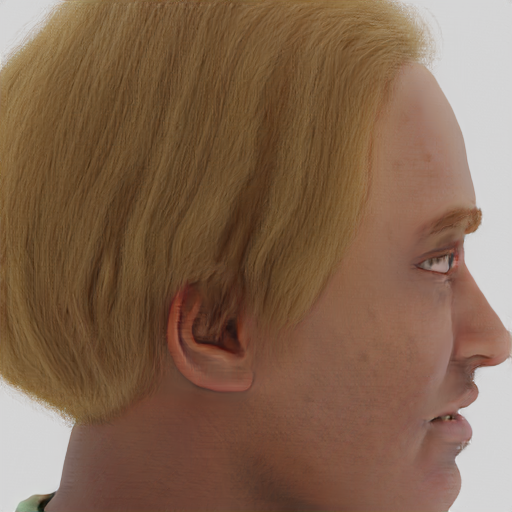}
            &  \includegraphics[width=1.6cm,height=1.6cm,keepaspectratio]{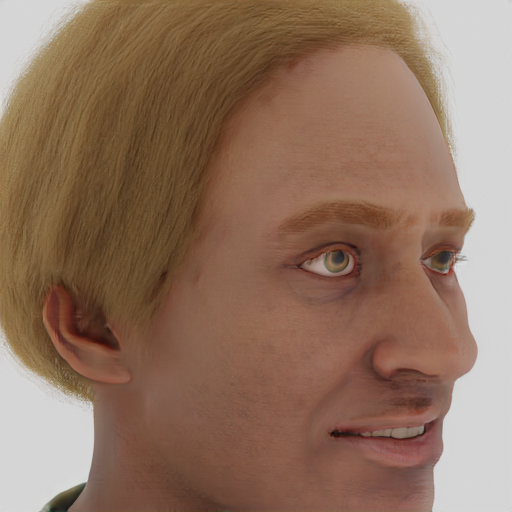}
            &  \includegraphics[width=1.6cm,height=1.6cm,keepaspectratio]{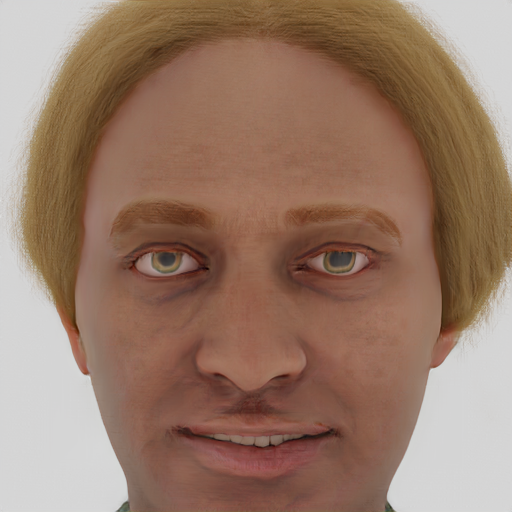}
            &  \includegraphics[width=1.6cm,height=1.6cm,keepaspectratio]{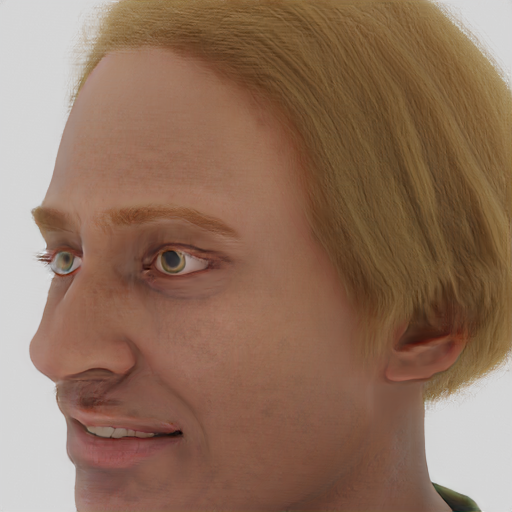}
            &  \includegraphics[width=1.6cm,height=1.6cm,keepaspectratio]{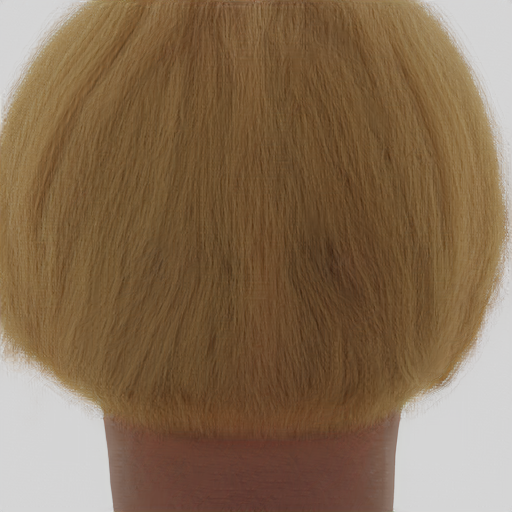} \\
        \end{tabular}
    \end{minipage}%
    \hfill
    \begin{minipage}[t]{0.5\textwidth}
        \centering
        \footnotesize
        \setlength{\tabcolsep}{0pt}
        \begin{tabular}{ccccccc}
            & Yaw: 0° & Yaw: 45° & Yaw: 90° & Yaw: 135° & Yaw: 270° \\
            \rotatebox{90}{\parbox[t]{2cm}{\hspace*{\fill}EG3D\hspace*{\fill}}}\hspace*{5pt}
            &  \includegraphics[width=1.6cm,height=1.6cm,keepaspectratio]{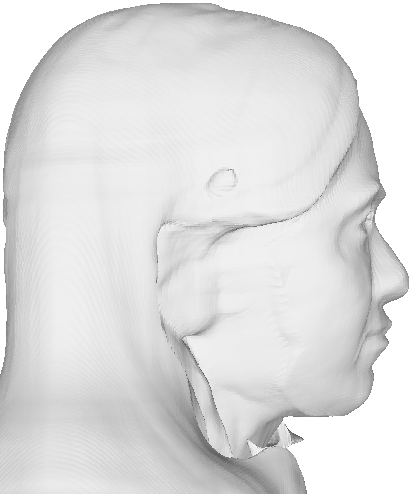}
            &  \includegraphics[width=1.6cm,height=1.6cm,keepaspectratio]{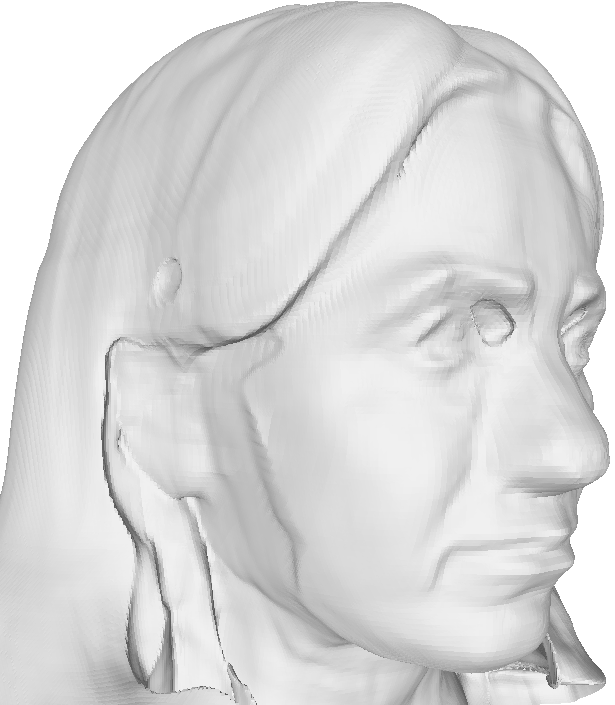}
            &  \includegraphics[width=1.6cm,height=1.6cm,keepaspectratio]{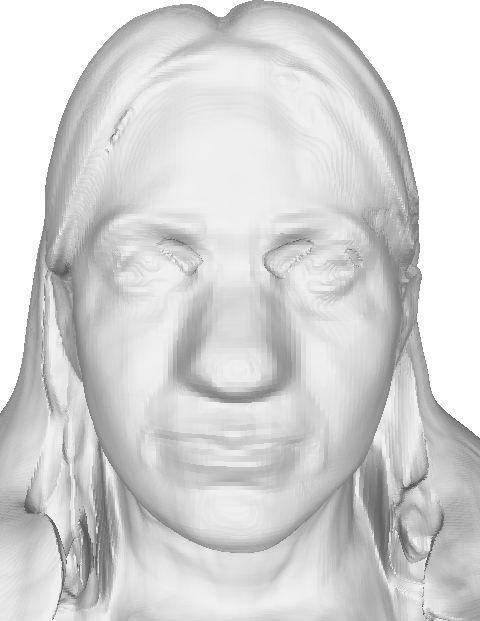}
            &  \includegraphics[width=1.6cm,height=1.6cm,keepaspectratio]{images/supp/syn_meshes/eg3d_syn/seed2/mesh_angle13508.png}
            &  \includegraphics[width=1.6cm,height=1.6cm,keepaspectratio]{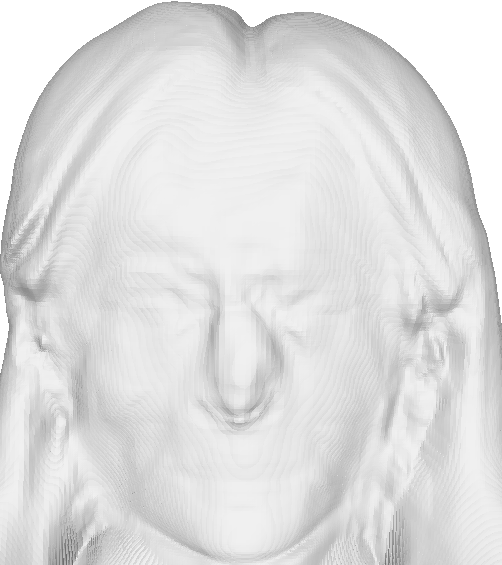} \\
            \rotatebox{90}{\parbox[t]{2cm}{\centering GeoGen w/o SDF DL}}\hspace*{5pt}
            &  \includegraphics[width=1.6cm,height=1.6cm,keepaspectratio]{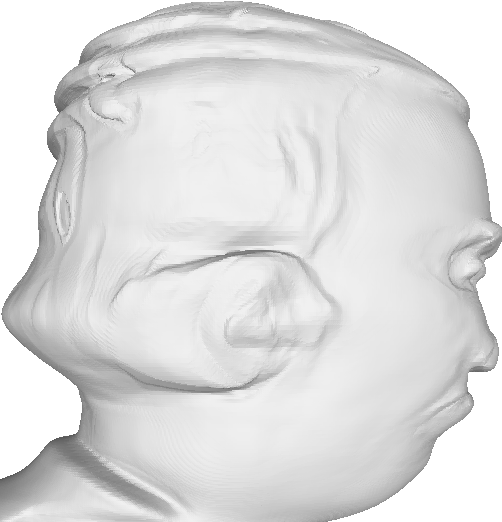}
            &  \includegraphics[width=1.6cm,height=1.6cm,keepaspectratio]{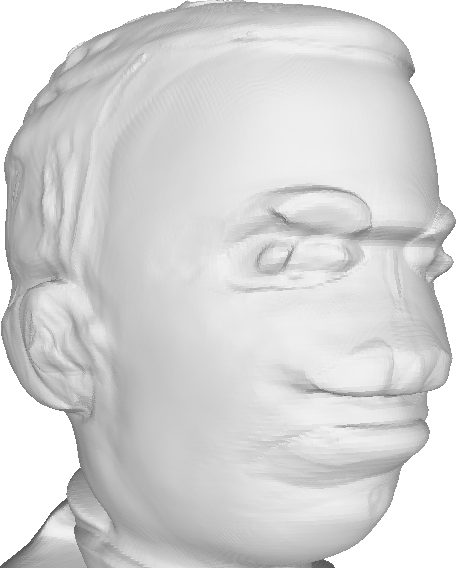}
            &  \includegraphics[width=1.6cm,height=1.6cm,keepaspectratio]{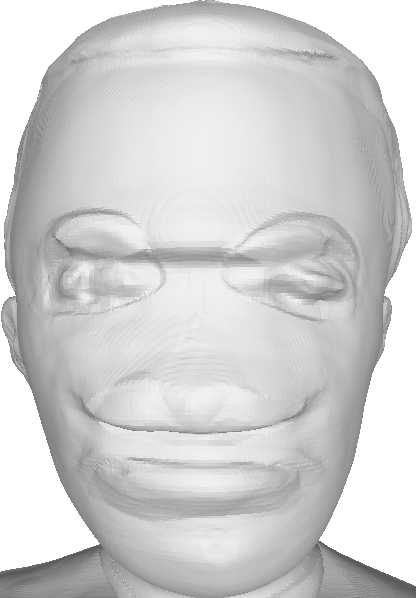}
            &  \includegraphics[width=1.6cm,height=1.6cm,keepaspectratio]{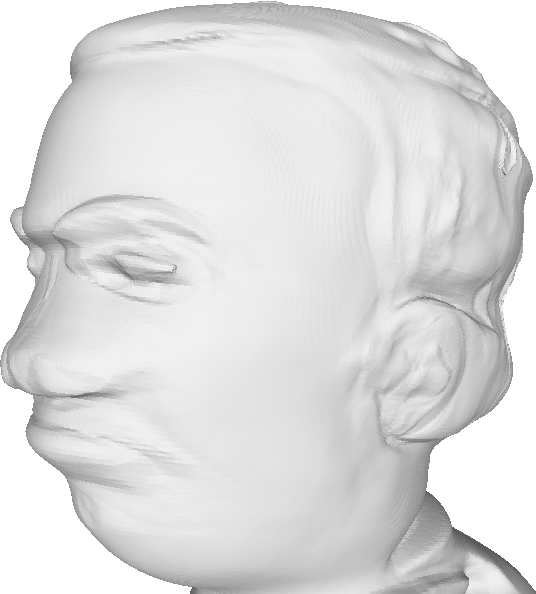}
            &  \includegraphics[width=1.6cm,height=1.6cm,keepaspectratio]{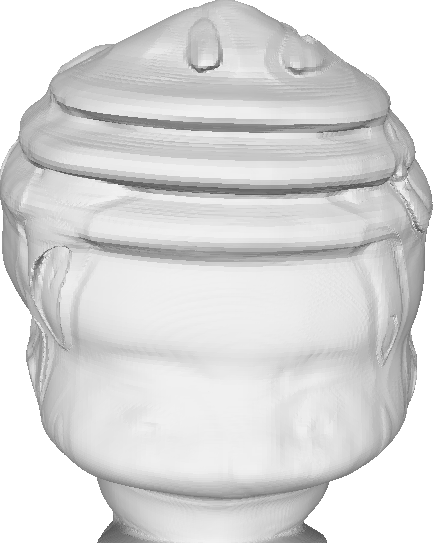} \\
            \rotatebox{90}{\parbox[t]{2cm}{\centering GeoGen}}\hspace*{5pt}
            &  \includegraphics[width=1.6cm,height=1.6cm,keepaspectratio]{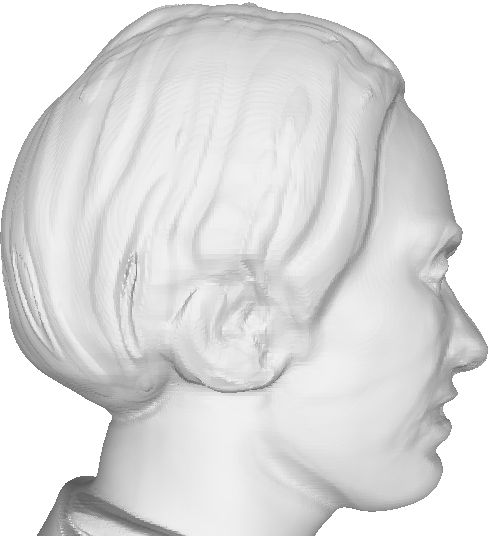}
            &  \includegraphics[width=1.6cm,height=1.6cm,keepaspectratio]{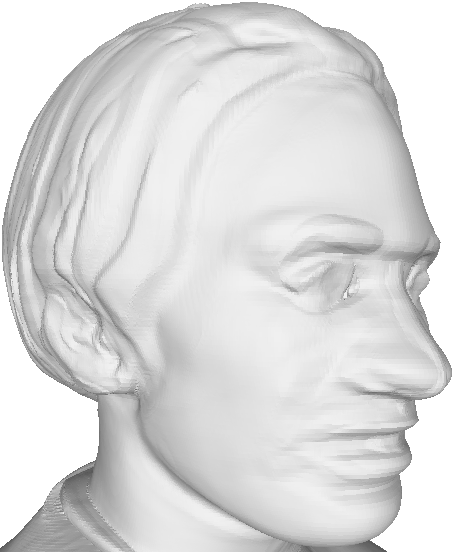}
            &  \includegraphics[width=1.6cm,height=1.6cm,keepaspectratio]{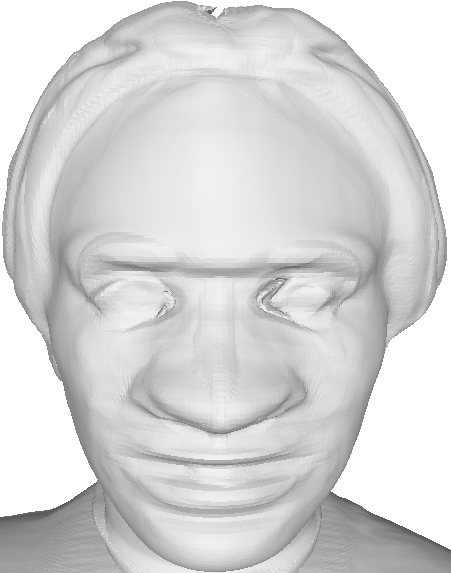}
            &  \includegraphics[width=1.6cm,height=1.6cm,keepaspectratio]{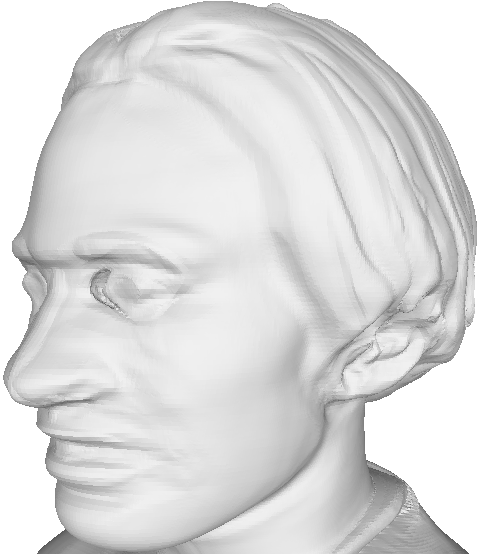}
            &  \includegraphics[width=1.6cm,height=1.6cm,keepaspectratio]{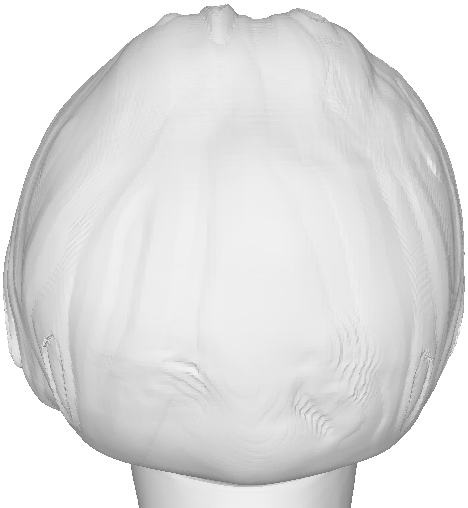} \\
        \end{tabular}
    \end{minipage}
    \vspace{-10pt}
    \caption{Comparison of EG3D and GeoGen, with and without SDF Depth Loss (SDF DL) constraints, showing sampled images from models trained on our synthetic human images. 
    These examples highlight GeoGen's ability to represent finer geometric details, \eg the ears have more detail than those generated by EG3D. We also observe a failure for EG3D in the top right, where the back of the head contains facial geometry. More qualitative results highlighting the differences in the use of the SDF depth loss are shown in the supplementary.}
    \label{fig:synthetics_samples}
    \vspace{-10pt}
\end{figure*}

\subsection{Quantitative results} 

We adopt the widely used Frechet Inception Distance (FID)~\cite{heusel2017gans} and Kernel Inception Distance (KID)~\cite{binkowski2018demystifying} metrics to measure the image synthesis quality of our GeoGen approach.  We also assess multi-view facial identity consistency (ID) by calculating the mean Arcface~\cite{deng2019arcface} cosine similarity score between pairs of views of the same synthesized face rendered from random camera poses. 
We report the results of our retrained EG3D baseline using the same training conditions and our GeoGen model on the three different datasets in Table~\ref{tab:datasets}.  Our improved results show that our GeoGen can achieve better image synthesis results on synthetic humans and ShapeNet Cars datasets. 

An important feature of our approach is its ability to generate accurate meshes from a single image. However, it is difficult to evaluate the \emph{geometric} quality of generative models on real images as ground-truth 3D shape information is challenging to obtain. 
Instead, it is possible to obtain the ground-truth meshes for both synthetic datasets that we use. To evaluate the generated meshes of different methods quantitatively, we leverage the GAN inversion technique PTI~\cite{roich2022pivotal}. 
Then, given an image from the test set dataset, we can estimate the corresponding latent code by PTI. 
With the latent code, we can generate both the synthesized image and mesh. 
In this way, we can compute a range of 3D evaluation metrics that compare the differences between the synthesized mesh and ground-truth mesh to measure the geometry fidelity. 
Results are presented in Table~\ref{tab:metrics}, where we observe that our GeoGen outperforms EG3D.

\begin{figure*}[t]
    \centering
    \includegraphics[width=0.7\linewidth]{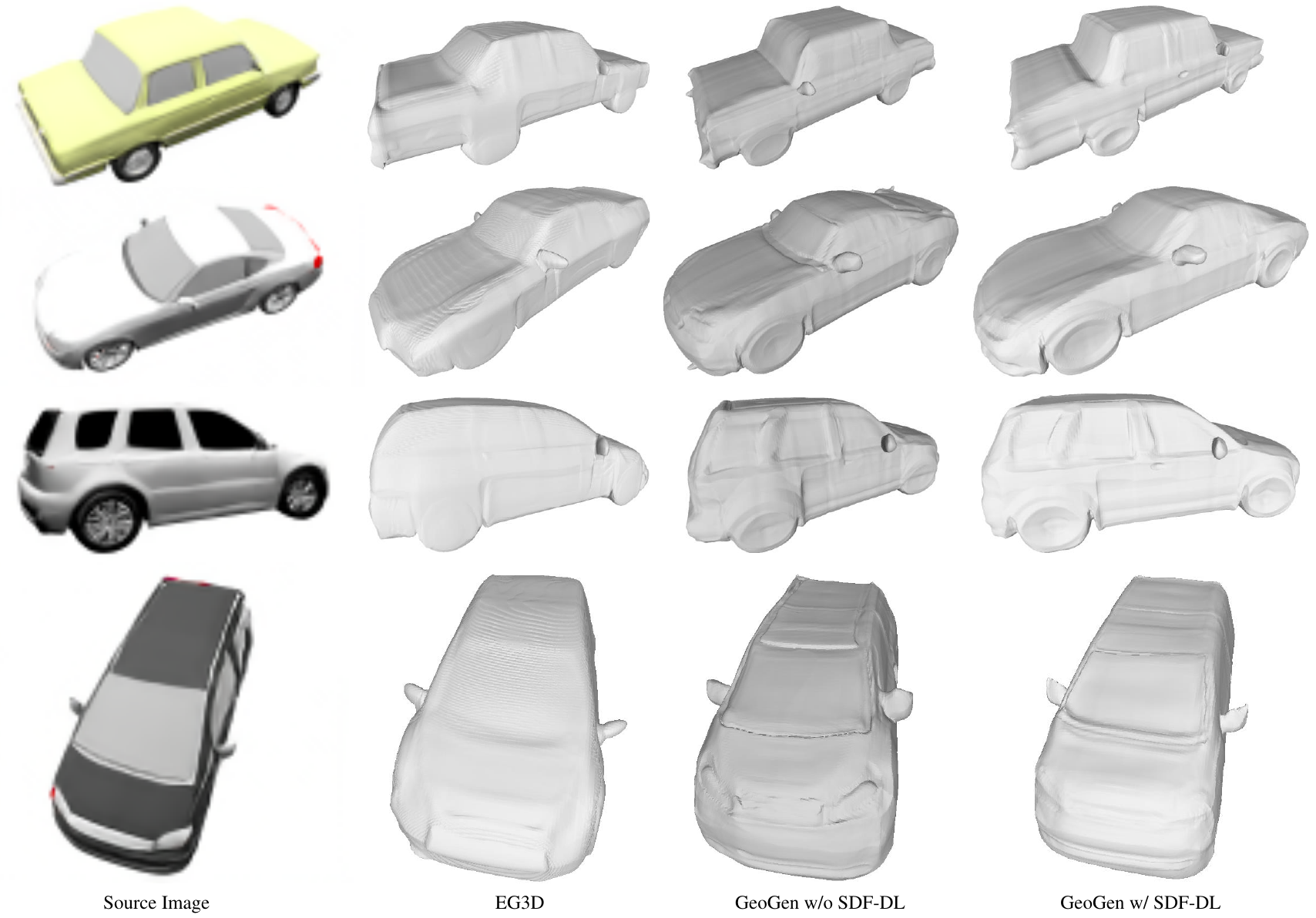}
    \vspace{-5pt}
    \caption{Comparison of mesh predictions on ShapeNet Cars. Meshes are obtained by inverting the source image to derive latent codes. EG3D meshes display diminished shape fidelity and surface detail. Using SDF constraints in GeoGen improves detail, evident around car wheels and windows. Results for GeoGen without SDF constraints are also shown for context.}
    \vspace{-10pt}
    \label{fig:car_inversion_meshes}
\end{figure*}

\subsection{Qualitative results} 
Here we present qualitative results where we compare GeoGen to existing methods. 
In Figures~\ref{fig:synthetics_inversion} and~\ref{fig:car_inversion_meshes} we compare 2D image synthesis of different methods via GAN inversion. We observe that GeoGen results in outputs that more closely match the input image. 
In Figure~\ref{fig:car_inversion_meshes} we observe that GeoGen captures details such as the spacing between the car body and wheel and, in some instances, even the handles on the doors of the cars. Finally, in Figures~\ref{fig:real_samples} and~\ref{fig:real_samples_2} we display sampled outputs (\ie not inversions).

\begin{table}[t]
    \centering
    \renewcommand{\arraystretch}{1.2}
    \resizebox{\columnwidth}{!}{%
    \begin{tabular}{l|c|c|c|c|c}
        \hline
        \multicolumn{6}{c}{\textbf{ShapeNet Cars}} \\
        \hline
        Method & Chamfer$\downarrow$ & MSE$\downarrow$ & HD$\downarrow$ & EMD$\downarrow$ & MSD$\downarrow$ \\
        EG3D & 0.31 & 0.31 & 0.85 & 0.44 & 0.33  \\
        GeoGen w/o SDF\&Depth Loss & 0.27 & 0.28 & \textbf{0.77} & 0.42 & 0.31 \\
        GeoGen & \textbf{0.25} & \textbf{0.27} & \textbf{0.77} & \textbf{0.40} & \textbf{0.29} \\
        \hline
        \multicolumn{6}{c}{\textbf{Synthetic Heads}} \\
        \hline
        Method & Chamfer$\downarrow$ & MSE$\downarrow$ & HD$\downarrow$ & EMD$\downarrow$ & MSD$\downarrow$ \\
        EG3D & 0.21 & 0.29 & 0.65 & 0.54 & 0.35 \\
        GeoGen w/o SDF\& Depth Loss& 0.19 & 0.29 & 0.59 & 0.45 & 0.26 \\
        GeoGen & \textbf{0.17} & \textbf{0.27} & \textbf{0.56} & \textbf{0.43} & \textbf{0.24} \\
        \hline
    \end{tabular}
    }
    \vspace{-5pt}
    \caption{Comparison of different 3D reconstruction metrics for generative models on \emph{ShapeNet Cars} and our \emph{Synthetic Heads} dataset. 
    We report averages for MSE, HD, and MSD metrics. Variations of GeoGen without the SDF and Depth Loss constraints are also shown. Best methods for each dataset are bolded.}
    \vspace{-10pt}
    \label{tab:metrics}
\end{table}

\section{Discussion}
\vspace{-5pt}
Our evaluation shows the competitive performance of our proposed GeoGen model, both qualitatively and quantitatively. 
To gain deeper insight into the effectiveness of our approach, we employed a suite of metrics that assess both the 2D and 3D aspects of the images and meshes generated by our model.
Two quantitative performance areas are of particular note: the synthesis of high-quality 2D images and precise 3D geometric predictions. 
Our model competes closely with EG3D~\cite{mildenhall2020nerf} in terms of 2D metrics, outperforming both StyleSDF~\cite{park2019deepsdf} and GRAF~\cite{schwarz2020graf}. 
This demonstrates our model's ability to generate high-fidelity 2D images.

Table~\ref{tab:metrics} showcases a systematic comparison between GeoGen and EG3D, revealing the advantages of incorporating Signed Distance Functions (SDF) and SDF depth constraints during training. 
The lower Chamfer Distance for GeoGen compared to EG3D for both Cars and synthetic human heads is indicative of a more precise alignment between the reconstructed points and corresponding points in the ground-truth. This highlights an improved precision in point-to-point correspondence which is an essential part of 3D reconstruction.  The Earth Mover's Distance, another vital metric in understanding the geometrical congruence between shapes, is also consistently lower for GeoGen. This indicates that the shapes are more similar, requiring fewer alterations to match the ground-truth, thus showing an underlying efficiency in GeoGen’s modeling approach. 
Finally, the Mean Surface Distance adds to the evidence of GeoGen's superiority, as it also yields consistently lower values. The implication here is a closer similarity between the reconstructed and target shapes, providing further evidence for GeoGen's effectiveness. 

The utilization of the SDF in GeoGen ensures better geometric consistency in the reconstruction, as it leverages the implicit representation of the mesh's surface. GeoGen, with its additional depth constraints, preserves topology and fine details that are often overlooked with conventional generative techniques like EG3D (see Figure~\ref{fig:synthetics_samples}). It is also noteworthy that these numerical advantages, though significant, do not fully represent the perceptual quality of the reconstructed models. Qualitative evaluations indicate that models generated by GeoGen often appear more realistic and accurate, underscoring GeoGen's advantage in bridging quantitative performance with perceptual realism.

\noindent{\textbf{Limitations.}} 
Our GAN-based approach, like others, requires posed images for training. 
Camera poses can be estimated similar to methods used in FFHQ. While we aim to align the expected depth with the SDF's zero-level set, extending the SDF consistency loss to other points along the ray could theoretically enhance geometric accuracy. However, this would substantially increase computational load. There are also inherent limitations in learning-based methods, such as potential bias from unrepresentative training data, notably in web-scraped human face images. 

\vspace{-5pt}
\section{Conclusion}
\vspace{-5pt}
We presented GeoGen, a novel 3D-aware generative model for synthesizing high-quality 2D images with associated accurate 3D geometry, that is trained from 2D images. 
GeoGen outperforms established methods on several performance metrics. By harnessing the power of neural implicit representations and neural signed distance functions, we have developed a solution that delivers both quality and versatility in the context of 3D representation learning. 
In addition, we presented a new synthetic human head dataset for training and quantitatively evaluating 3D generative models.  
GeoGen moves us closer to the goal of enriching fields such as character animation, gaming, and virtual reality with plausible 3D geometry from single input images.  
Our results affirm the potential of our approach and its relevance in this rapidly evolving field.

{
    \small
    \bibliographystyle{ieeenat_fullname}
    \bibliography{main}
}

\clearpage
\appendix
\setcounter{table}{0}
\renewcommand{\thetable}{A\arabic{table}}
\setcounter{figure}{0}
\renewcommand{\thefigure}{A\arabic{figure}}

\input{supp_content}

\end{document}

%% file: supp_content.tex
\noindent{\huge Appendix}

The foundation of our model relies on the official implementation of Enhanced Generative 3D Models (EG3D) \cite{chan2023eg3d}. We utilized R1 regularization, assigning a gamma = 1 for the synthetic humans and FFHQ dataset based on the input image size of 512 x 512 and batch size of 32 across 8 v100 GPUs, following the same hyperparameter tuning of EG3D. For ShapeNet Cars, we adopted a gamma value of 0.3 based on the 128 x 128 resolution and batch size of 32 \cite{fu2022geo}. Our model employs the same architecture as StyleGAN2 \cite{or2022stylesdf}, composed of a mapping network with 8 hidden layers, and output convolutions yielding 96 feature maps. Following the EG3D protocol, these are then reshaped into 3 planes of 256 x 256 x 32 \cite{chan2023eg3d}.

\subsection{GeoGen training}
During the initial training of GeoGen for the FFHQ and Synthetics dataset, the model was trained end-to-end, a process that necessitated unique handling of the SDF depth consistency loss. For the first 10,000 epochs, we set the beta value for the Laplace density distribution to 0.1 and refrained from making it learnable, as our end-to-end model would not have been able to learn the best beta value at this stage \cite{fu2022geo}. This approach allowed the model to first learn the optimal geometry and SDF depth map. In contrast, StyleSDF had to introduce a two-stage training process precisely because their pipeline was not trained end-to-end. They consistently used a learnable beta parameter for the Laplace density distribution throughout their training, as their method required more flexibility in the control of the SDF consistency loss.

The Laplace beta value plays a crucial role in the SDF network as it controls the shape of the Laplace distribution, influencing how the model penalizes deviations from the expected SDF values. A lower beta value produces a wider distribution, allowing for a larger spread of SDF values, and a higher beta value tightens the distribution, constraining the SDF values more strictly. This ability to control the distribution of SDF values enables fine-tuning of the model's sensitivity to inconsistencies in the SDF depth, a key aspect of the learning process.
After the generator in our model showed improvement in rendering, depth maps, and underlying geometry, we activated the SDF constraint for depth map regularization and introduced the learnable beta parameter for the remaining 10,000 epochs. This allowed us to dynamically adapt the SDF consistency loss and fine-tune the model's learning of SDF depth.

Both EG3D and GeoGen models underwent training for 20,000 epochs for the FFHQ and Synthetics data, while for the ShapeNet dataset, training was conducted for 10,000 epochs. The batch size for all models was 18, with the discriminator's learning rate at 0.002 and the generator's at 0.0025. The training was carried out using 4 NVIDIA P100, while an RTX 2080 and RTX 4090 were used for inference during inversions and sample generation. Our end-to-end training approach, including the specific handling of the Laplace beta value, was central to our method's effectiveness in learning SDF depth. It allowed us to combine the flexibility needed in the early stages of learning with the precision required in later stages, reflecting a sophisticated understanding of the role that SDF plays in the generative process.

\subsection{SDF and color network and surface rendering}

The resulting embedding from the augmented spatial representation is fed into the SDF (Signed Distance Function) network. This network utilizes the embedded position to query the SDF value at a specific point, which gives precise information regarding the distance to the nearest surface within the 3D space. The understanding of these distances is crucial in the reconstruction of 3D objects, as it provides detailed insights into the geometry and the underlying complexities of the data being modeled.

Once the SDF network receives and processes the embedded position, the computed SDF values are further handled by the color network. This auxiliary network takes the SDF values and translates them into the corresponding color values for the rendered 3D object. The direct utilization of SDF values as input for the color network establishes a coherent link between the geometric structure and visual appearance of the object. Both the SDF and color networks are built with a single hidden layer comprising 64 hidden units and leverage a soft plus activation function. This structure ensures smooth transitions and optimal gradient flow within the networks. For the transformation of the SDF into tangible density, a specific surface rendering technique has been applied. The sampling strategy is carefully chosen and tailored to different datasets, such as using 48 uniformly spaced samples and 48 importance samples per ray for the FFHQ dataset, and 64 of each for ShapeNet cars and Synthetics data.

In combination, these elements forge an intricate pipeline that integrates spatial features and coordinates, through a positional encoder, with the SDF and color networks. The methodology's architecture ensures a nuanced and true-to-life representation across a multitude of datasets. The implementation of a positional encoder has further enhanced the SDF network's capacity to grasp and replicate complex 3D geometries. The employment of SDF networks for surface rendering has led to a more sophisticated and resilient interpretation of various datasets.

\begin{figure*}[th]
    \centering
    \tiny %
    \setlength{\tabcolsep}{0pt}
    \resizebox{0.95\textwidth}{!}{%
        \begin{tabular}{cccccc}
            & car1 & car2 & car3 & car \\
            \rotatebox{90}{\parbox[t]{2cm}{\hspace*{\fill}Source Image\hspace*{\fill}}}\hspace*{5pt}
                & \includegraphics[width=1.6cm]{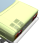}
                & \includegraphics[width=1.8cm]{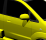}
                & \includegraphics[width=1.5cm]{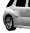}
                & \includegraphics[width=2cm]{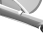} \\
            \rotatebox{90}{\parbox[t]{2cm}{\hspace*{\fill}EG3D\hspace*{\fill}}}\hspace*{5pt}
                & \includegraphics[width=2cm]{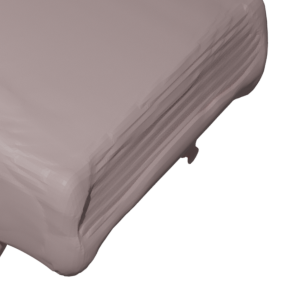}
                & \includegraphics[width=2cm]{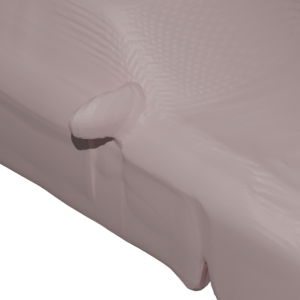}
                & \includegraphics[width=2cm]{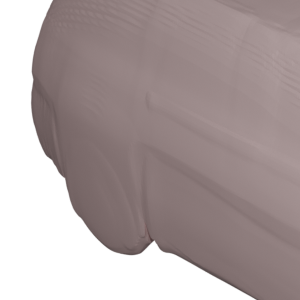}
                & \includegraphics[width=2cm]{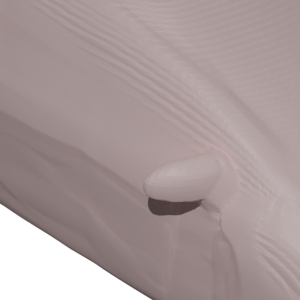} \\
            \rotatebox{90}{\parbox[t]{2cm}{\hspace*{\fill}GeoGen w/0 SDF-DL\hspace*{\fill}}}\hspace*{5pt}
                & \includegraphics[width=2cm]{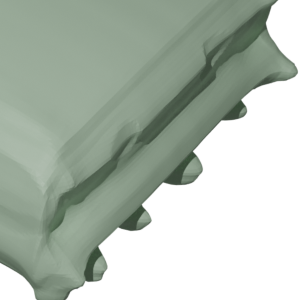}
                & \includegraphics[width=2cm]{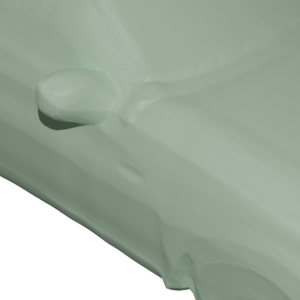}
                & \includegraphics[width=2cm]{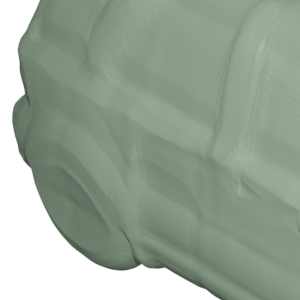}
                & \includegraphics[width=2cm]{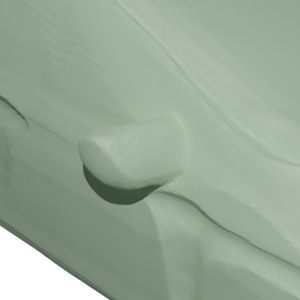} \\
            \rotatebox{90}{\parbox[t]{2cm}{\hspace*{\fill}GeoGen\hspace*{\fill}}}\hspace*{5pt}
                & \includegraphics[width=2cm]{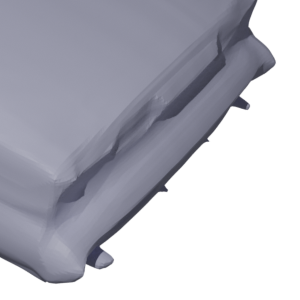}
                & \includegraphics[width=2cm]{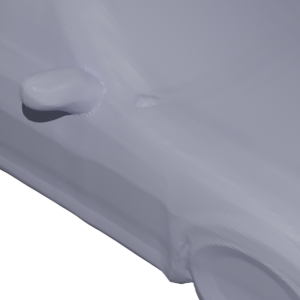}
                & \includegraphics[width=2cm]{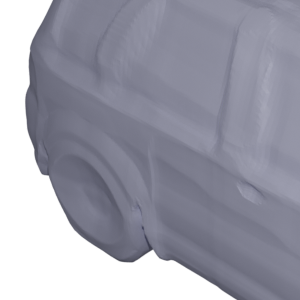}
                & \includegraphics[width=2cm]{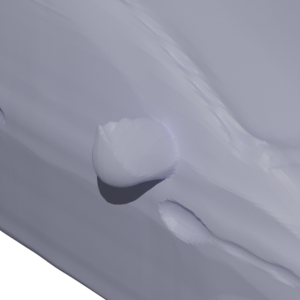}
        \end{tabular}
    }
    \caption{A detailed comparison between EG3D and GeoGen in the context of ShapeNet cars inversion of meshes, emphasizing the differences in the geometric representation and rendering capabilities of both methods. The samples underscore the advanced efficacy of GeoGen in capturing and reconstructing intricate geometric details within the car models, even at granular levels. This superiority is attributed to the integration of the Signed Distance Function (SDF) network along with the SDF depth consistency loss within GeoGen's architecture. The SDF approach provides a continuous and differentiable representation of the car's surface, enabling more precise and robust alignment with the observed data. This contributes to better capturing of fine geometrical nuances and results in more accurate reconstructions. Conversely, the EG3D~\cite{chan2023eg3d} method's rendered meshes reveal a deficiency in portraying granular details, leading to a more approximate and less nuanced depiction of the vehicles.}
    \label{fig:car_zoom}
\end{figure*}

\subsection{Reconstruction of pseudo ground truth meshes}
To reconstruct pseudo ground truth meshes we use Planar Prior Assisted PatchMatch Multi-View Stereo (ACMP)~\cite{xu2020planar} and Poisson surface reconstruction~\cite{poisson_surface}. 
Example pseudo ground truth meshes are shown in Figure~\ref{fig:synthetics_inversion_supp}. 
Recognizing the challenge of depth estimation in low-textured areas, which typically exhibit strong planarity, ACMP makes use of planar models in conjunction with the PatchMatch algorithm. By embedding planar models into PatchMatch MVS via a probabilistic graphical model, our approach introduces a multi-view aggregated matching cost. This novel cost function takes both photometric consistency and planar compatibility into consideration~\cite{xu2020planar}, thus accommodating both non-planar and planar regions. This method has demonstrated its capability to recover depth information in areas of extremely low texture, efficiently leading to high completeness in 3D models.

The problem of surface reconstruction from oriented points is cast as a spatial Poisson problem using Poisson surface reconstruction. This formulation's advantage is its simultaneous consideration of all points without the need for heuristic spatial partitioning or blending, which enhances resilience to data noise \cite{poisson_surface}. The use of a hierarchy of locally supported basis functions and the reduction of the solution to a well-conditioned sparse linear system makes this approach computationally efficient.

By seamlessly integrating ACMP with Poisson surface reconstruction, we've crafted a novel method for 3D model reconstruction. The fusion of these techniques allows us to address the complexities and subtleties of 3D modeling, particularly in challenging scenarios where noise and low texture might otherwise impede reconstruction. The reconstructed pseudo-ground truth meshes generated by this combined approach are a testament to its effectiveness, signifying an exciting advancement in the realm of 3D modeling and a promising avenue for further exploration and optimization.

\begin{figure*}[t]
    \begin{minipage}[t]{0.5\textwidth}
        \centering
        \footnotesize
        \setlength{\tabcolsep}{0pt}
        \begin{tabular}{ccccccc}
            & Yaw: 0° & Yaw: 90° & Yaw: 180°  \\
            \rotatebox{90}{\parbox[t]{2cm}{\centering{GeoGen w/o PE\&DL}}}\hspace*{5pt}
            &  \includegraphics[width=2cm,height=2cm,keepaspectratio]{images/supp/syn_meshes/geogen_noloss/seed1/angle000.png}
            &  \includegraphics[width=2cm,height=2cm,keepaspectratio]{images/supp/syn_meshes/geogen_noloss/seed1/angle045.png}
            &  \includegraphics[width=2cm,height=2cm,keepaspectratio]{images/supp/syn_meshes/geogen_noloss/seed1/angle090.png}\\
            \rotatebox{90}{\parbox[t]{2cm}{\hspace*{\fill}GeoGen with PE\hspace*{\fill}}}\hspace*{5pt}
            &  \includegraphics[width=2cm,height=2cm,keepaspectratio]{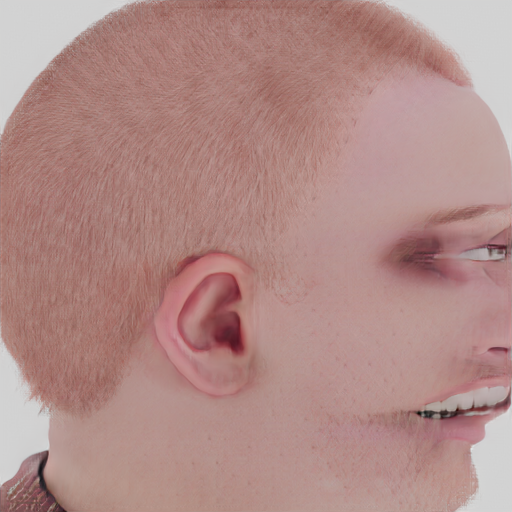}
            &  \includegraphics[width=2cm,height=2cm,keepaspectratio]{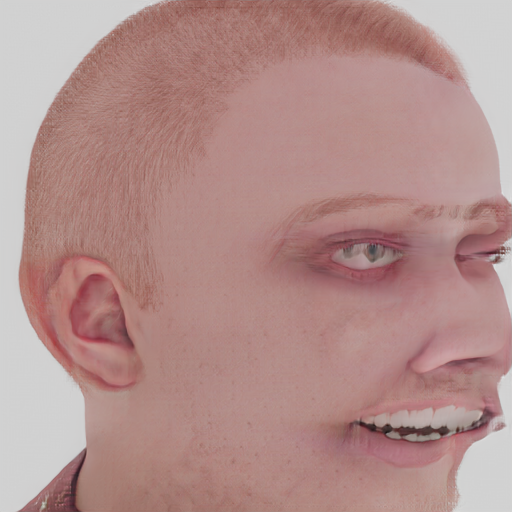}
            &  \includegraphics[width=2cm,height=2cm,keepaspectratio]{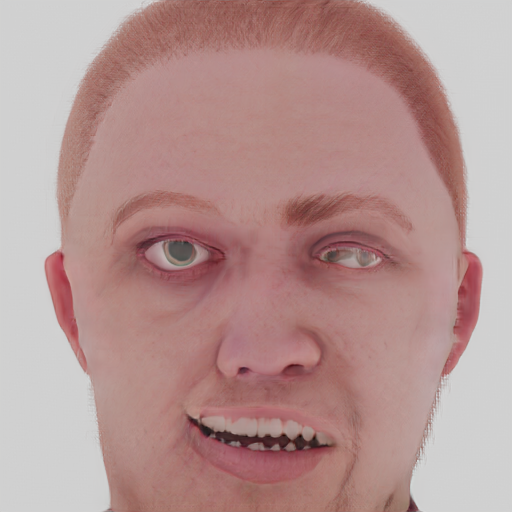}\\
        \end{tabular}
    \end{minipage}%
    \hfill
    \begin{minipage}[t]{0.5\textwidth}
        \centering
        \footnotesize
        \setlength{\tabcolsep}{0pt}
        \begin{tabular}{ccccccc}
            & Yaw: 0° & Yaw: 90° & Yaw: 180°  \\
            \rotatebox{90}{\parbox[t]{2cm}{\hspace*{\fill}GeoGen w/0 PE\hspace*{\fill}}}\hspace*{5pt}
            &  \includegraphics[width=2cm,height=2cm,keepaspectratio]{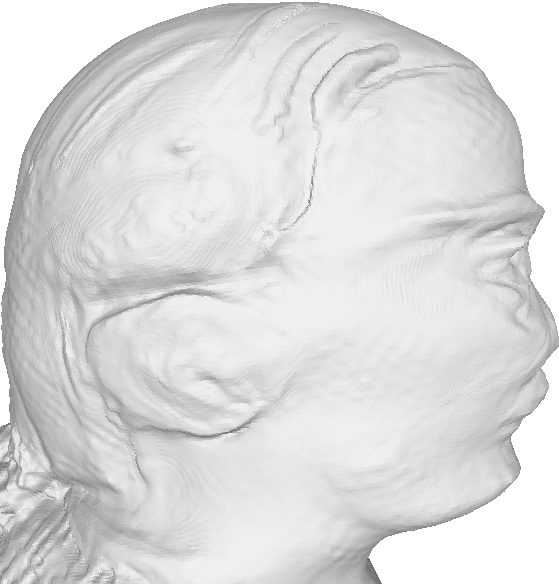}
            &  \includegraphics[width=2cm,height=2cm,keepaspectratio]{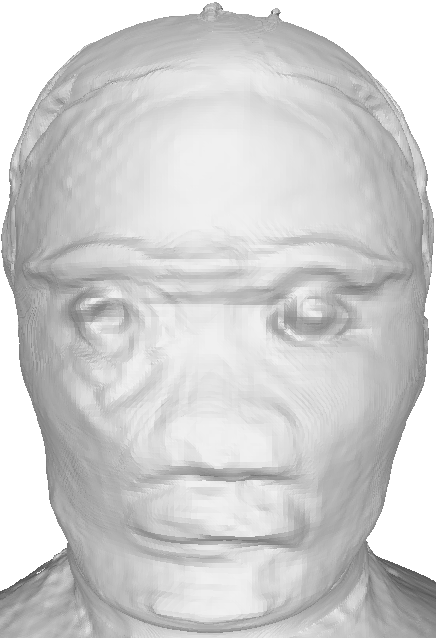}
            &  \includegraphics[width=2cm,height=2cm,keepaspectratio]{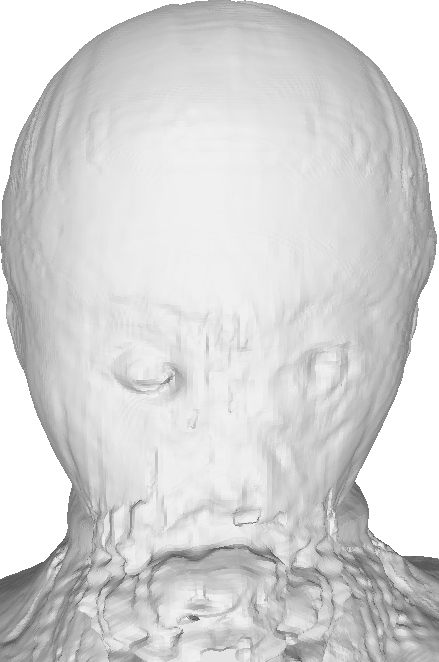}\\
            \rotatebox{90}{\parbox[t]{2cm}{\hspace*{\fill}GeoGen w/0 PE\hspace*{\fill}}}\hspace*{5pt}
            &  \includegraphics[width=2cm,height=2cm,keepaspectratio]{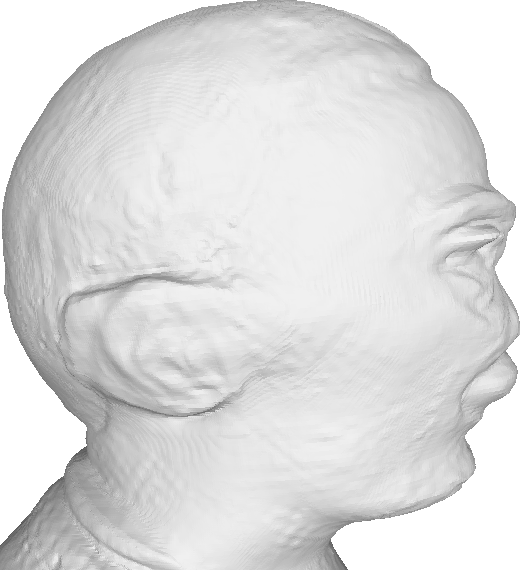}
            &  \includegraphics[width=2cm,height=2cm,keepaspectratio]{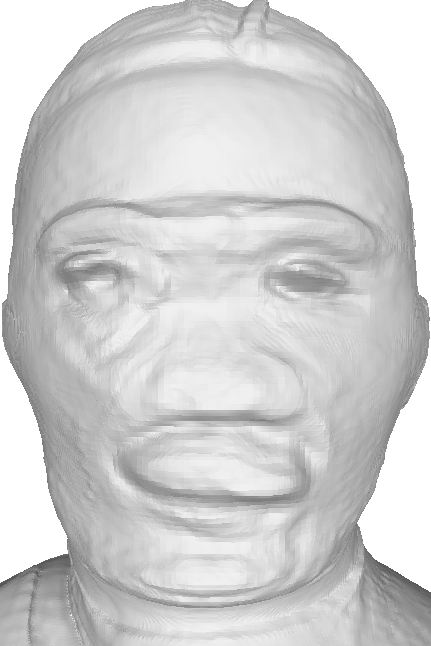}
            &  \includegraphics[width=2cm,height=2cm,keepaspectratio]{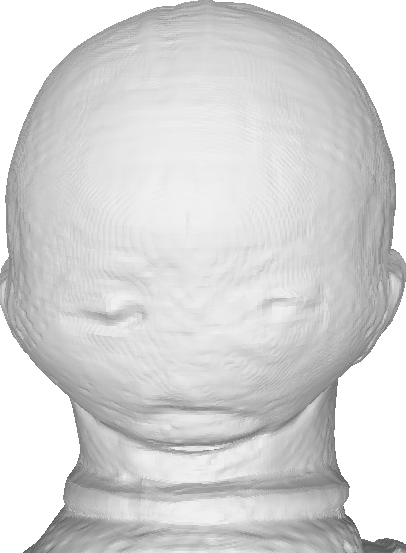}\\
        \end{tabular}
    \end{minipage}
    \vspace{-10pt}
    \caption{This caption accompanies a series of synthetic images generated by the GeoGen model operating without a positional encoder. The figures on the left illustrate the model's output at different yaw angles, showcasing its ability to render facial features from various perspectives. On the right, the corresponding mesh structures are displayed, providing a deeper insight into the model's geometric rendering capabilities. These results were captured prior to the point of model collapse, highlighting the model's performance and limitations in the absence of positional encoding. This comparison not only demonstrates the visual output of the model but also underscores the critical role of positional encoding in maintaining structural integrity and realism in the generated images and meshes.} 
    \label{fig:synthetics_samples_supp}
    \vspace{-10pt}
\end{figure*}

\subsection{Results without positional encoder}
Here we explore causes behind the collapse of the GeoGen model, specifically when trained without the aid of positional encoding in the context of Neural Radiance Fields (NeRF) and GAN training. The absence of positional encoding can lead to several critical issues (see Figure~\ref{fig:synthetics_samples_supp}). 
Firstly, in GAN training, the phenomenon of mode collapse becomes more pronounced. This is where the generator starts producing a limited variety of outputs, failing to capture the complex data distribution. Secondly, the intrinsic characteristics of NeRF, which rely heavily on precise spatial information to render 3D scenes accurately, are compromised without positional encoding. This results in the model's inability to effectively learn and represent high-frequency details, leading to a loss of detail and realism in the generated images. Lastly, positional encoding plays a vital role in stabilizing the training process by providing a more detailed and nuanced understanding of spatial relationships in the data. Its absence can result in unstable training dynamics, ultimately causing the model to collapse, particularly evident in our observations post epoch 11000. This highlights the essential nature of positional encoding in maintaining the stability and efficacy of models like GeoGen, especially in complex applications involving synthetic human images and 3D rendering.

\section{Datasets}

\subsection{FFHQ and rebalanced FFHQ}

Our modeling framework originally utilized the "in-the-wild" version of the FFHQ dataset, a comprehensive collection of uncropped, original PNG human images sourced from Flickr, as documented by Karras et al. (2019) \cite{karras2019style}. To adapt these images for our purposes, we employed a sophisticated face detection and pose-extraction system \cite{chan2023eg3d}, allowing us to determine the face area and label each image with its corresponding pose. The images were then cropped to approximate the dimensions of the original FFHQ dataset. We assumed fixed camera intrinsics for all images, with a focal length 4.26 times the image width, mimicking a standard portrait lens \cite{chan2023eg3d}. After removing a small number of images where face detection proved unsuccessful, our final dataset comprised 69,957 images.

In our reporting, we include the 2D performance metrics of models trained on the Rebalanced FFHQ dataset, particularly focusing on the outcomes from NVIDIA-trained models. The Rebalanced FFHQ dataset, known for its broader diversity in facial orientations, plays a crucial role in enhancing the model's capability to understand and replicate human facial features from various angles. This dataset is especially valuable for models that need to handle a wide range of facial geometries, such as those used in advanced image generation and recognition tasks.

While we present these metrics to showcase the performance improvements facilitated by the Rebalanced FFHQ dataset, it's important to note a limitation in the available data. NVIDIA, the entity responsible for training these models, has not provided detailed information regarding the number of epochs, specific training methodologies, or other intricate details of the training process. This lack of detailed training information could potentially impact the reproducibility and further optimization of these models.

Understanding the training duration (measured in epochs) and the specific methodologies employed is crucial for comprehensively evaluating a model's performance and for making informed comparisons with other models. The absence of this information leaves a gap in fully understanding how the Rebalanced FFHQ dataset impacts model performance compared to the original FFHQ dataset. Despite this, the reported 2D metrics still offer valuable insights into the enhanced capabilities of models trained on the Rebalanced FFHQ dataset, highlighting their improved proficiency in handling diverse facial features and orientations.

\begin{figure*}[t]
  \centering
  \begin{tabular}{lcccc}
    
    \rotatebox{90}{\parbox[t]{2cm}{\hspace*{\fill}GT\hspace*{\fill}}}\hspace*{5pt}
    & \includegraphics[width=0.21\linewidth, keepaspectratio]{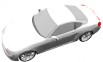}
    & \includegraphics[width=0.21\linewidth, keepaspectratio]{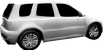}
    & \includegraphics[width=0.12\linewidth, keepaspectratio]{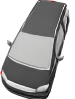}
    & \includegraphics[width=0.2\linewidth, keepaspectratio]{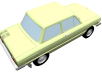} \\[-4pt]

    \rotatebox{90}{\parbox[t]{2cm}{\hspace*{\fill}EG3D\hspace*{\fill}}}\hspace*{5pt}
    & \includegraphics[width=0.21\linewidth, keepaspectratio]{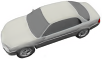}
    & \includegraphics[width=0.21\linewidth, keepaspectratio]{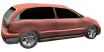}
    & \includegraphics[width=0.12\linewidth, keepaspectratio]{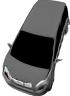}
    & \includegraphics[width=0.2\linewidth, keepaspectratio]{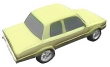} \\[-4pt]

    \rotatebox{90}{\parbox[t]{2cm}{\hspace*{\fill}GeoGen\hspace*{\fill}}}\hspace*{5pt}
    & \includegraphics[width=0.21\linewidth, keepaspectratio]{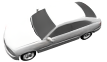}
    & \includegraphics[width=0.21\linewidth, keepaspectratio]{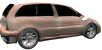}
    & \includegraphics[width=0.12\linewidth, keepaspectratio]{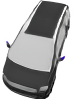}
    & \includegraphics[width=0.2\linewidth, keepaspectratio]{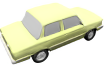} \\[6pt]
  \end{tabular}
  \caption{Comparison of EG3D and GeoGen inversion results using held-out images from the ShapeNet Car test set. GeoGen results more closely resemble the input ground truth image (GT).}
  \label{fig:inversion_cars}
\end{figure*}

\subsection{ShapeNet V1}

We utilized the ShapeNet V1 Cars dataset for additional validation, rigorously comparing methodologies on a specific subset that includes 128 renderings of synthetic cars \cite{chang2015shapenet}. This carefully curated dataset offers a robust platform for assessing performance across various viewing angles, enabling a comprehensive evaluation of 3D reconstruction and rendering techniques.

The ShapeNet dataset, as employed in our setup, builds on prior research and consists of 2,100 car images captured from 50 different perspectives \cite{chang2015shapenet}. The multi-angle images provide an ideal scenario to analyze geometric consistency, shadow rendering, and surface texturing. Similar to the preprocessing applied to the FFHQ dataset, our approach to the ShapeNet data followed established protocols, maintaining the integrity and original characteristics of the images. Unlike other methodologies that might use augmentation or mirror images, we consciously chose not to apply these techniques to preserve the authenticity of the data and ensure a more accurate assessment of the models' performance \cite{chang2015shapenet}.

\subsection{Synthetic humans}

Our training model also harnessed our proprietary synthetic human dataset. This extensive collection encompasses 200,000 images, representing 20,000 unique identities. Each of these identities is portrayed from only 10 viewpoints, a stark contrast to the Rodin model where each identity was rendered from 300 diverse viewpoints~\cite{wang2023rodin}. Despite the significant reduction in viewpoints per identity in our dataset, our model produces high quality outputs in terms of geometry and rendering \cite{an2023panohead}. Our training approach proves that strong performance can be achieved with a more limited number of viewpoints.

\subsection{Pivotal tuning inversion}

In the context of our work with Pivotal Tuning Inversion (PTI), a specialized process to invert generative models like StyleGAN, we adopt a meticulous procedure to enhance the accuracy and efficiency of the inversion.

Initially, we utilize an off-the-shelf face detection solution to accurately locate and extract face regions within the test images. This process allows for precise alignment and ensures that the features of interest are adequately centered and scaled. The extracted regions are then cropped and resized to a consistent resolution of 512x512 pixels, facilitating uniform processing and analysis across different images.

Following this preprocessing stage, we implement the PTI methodology as delineated by Tov et al. \cite{tov2023pivotal}. This approach consists of two main stages:

\begin{enumerate}

\item \textbf{Fine-tuning of generator weights.} Subsequent to the initial latent code optimization, we proceed with an additional 500 iterations dedicated to fine-tuning the generator's weights. This phase is pivotal in refining the subtle details and enhancing the realism of the generated images. By adjusting the generator's parameters, we align the synthetic outputs more closely with the underlying distribution of the real data, improving both the fidelity and the perceptual quality of the inversions.

\item \textbf{Latent code optimization.} For the first 500 iterations, we focus on the optimization of the latent code, a compact representation within the model's latent space that encodes the essential features of the target image. Utilizing gradient-based optimization techniques, we iteratively refine the latent code to minimize the discrepancy between the generated image and the target. This stage ensures that the inverted model captures the essential characteristics of the face.

\end{enumerate}

The combination of these two stages offers a robust and precise inversion process, enabling us to generate high-quality, detailed images that faithfully represent the original inputs. The PTI methodology, by explicitly separating the optimization of the latent code and the fine-tuning of the generator, provides a nuanced control over the inversion process, yielding superior results in terms of both accuracy and visual appeal.

\subsection{Justifying the limitations in GAN inversion}

In the field of Generative Adversarial Networks (GANs), particularly with advanced models like EG3D, the accuracy of GAN inversion can be inconsistent. This inconsistency can be attributed to several factors, encompassing both the inherent characteristics of the generative model and the methodologies used in the inversion process.

Firstly, the architecture and complexity of the GAN model play a crucial role. A model with limitations in its design may not capture a broad range of features effectively, leading to challenges in accurately reproducing certain types of images during inversion. For example, if the model's architecture does not account for a wide variety of facial orientations, it may struggle with accurately inverting images that fall outside of its trained norm.

Additionally, the scope and diversity of the training data are critical. A model trained on a dataset with limited variety, such as one primarily consisting of front-facing images, may not perform well in inverting images with diverse or unusual orientations. The quality and diversity of the training data directly influence the model's ability to handle a wide range of inversion tasks.

Furthermore, the model's resolution and detail capabilities are also significant. Models that generate lower-resolution images or lack fine detail might fail to accurately capture nuances in the inversion process, resulting in less precise or realistic inversions.

On the side of inversion methodologies, the efficiency of the algorithm and its approach to navigating and manipulating the latent space of the GAN are key factors. The choice of loss functions and regularization techniques within the inversion method can greatly affect the match quality between the inverted image and the original. Computational constraints can also limit the effectiveness of more resource-intensive, yet potentially more accurate, inversion methods.

In summary, the limitations in GAN inversion accuracy can be attributed to a complex interplay of factors related to both the generative model's characteristics and the inversion techniques used. Understanding and addressing these factors is crucial for improving the accuracy and reliability of GAN inversions.

\begin{figure*}[h]
  \centering
  \begin{minipage}{0.4\textwidth}
    \centering
    \textbf{Without GeoGen SDF Constraint}\\  %
    \includegraphics[width=\linewidth]{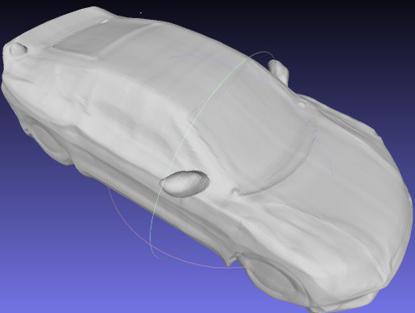}  %
  \end{minipage}%
  \begin{minipage}{0.4\textwidth}
    \centering
    \textbf{With GeoGen SDF Constraint}\\  %
    \includegraphics[width=\linewidth]{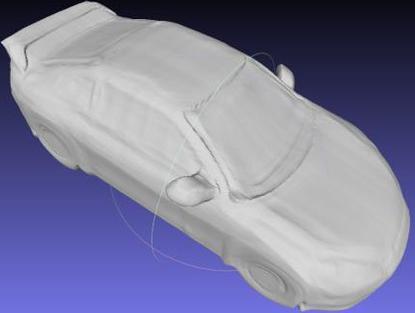}  %
  \end{minipage}
  \caption{Comparison of models without (left) and with (right) our GeoGen SDF constraint.}
\label{fig:car_zoom1}
\end{figure*}

\subsection{Evaluation metrics}

Evaluating the quality and performance of generated images is paramount in understanding the effectiveness of generative models. To this end, we employed the Fréchet Inception Distance (FID) and Kernel Inception Distance (KID), calculating these metrics for 50,000 generated images against all training images for both FFHQ and synthetic humans datasets. The calculations were performed using the implementation provided in the StyleGAN2 codebase~\cite{karras2019style}, ensuring consistency with commonly accepted standards.

Our GeoGen model's KID scores were found to be 100 times lower than those of comparative models, an unexpected result that warrants careful consideration. One possible hypothesis for this abnormality might be an alignment of specific features or particularities in the convergence behavior during the training of our model. It could also be related to the choice of hyperparameters or the data preprocessing steps that were unique to our experiment. However, these hypotheses are subject to further investigation, and the exact reason behind the unusually low KID score remains an intriguing question for future research.

Alongside the 2D image quality evaluation, we also assessed 3D geometry comparisons, adopting the Efficient Geometry Aware 3D Network (EG3D) \cite{chan2023eg3d} for evaluation. Our GeoGen model showed promising results relative to the EG3D model, as indicated by these metrics, both in terms of 2D image quality and 3D Chamfer distance metrics. The overall evaluation paints a comprehensive picture of our model's capabilities, but the abnormally low KID score serves as a reminder that there may always be underlying complexities and subtleties that require further exploration and understanding.

\begin{figure*}[t]
    \centering
    \footnotesize
    \resizebox{0.75\textwidth}{!}{%
    \setlength{\tabcolsep}{0pt}
    \begin{tabular}{ccccccc}
        \rotatebox{90}{\parbox[t]{3cm}{\hspace*{\fill}Source Image\hspace*{\fill}}}\hspace*{5pt}
        &  \includegraphics[width=2.5cm,height=2.5cm,keepaspectratio]{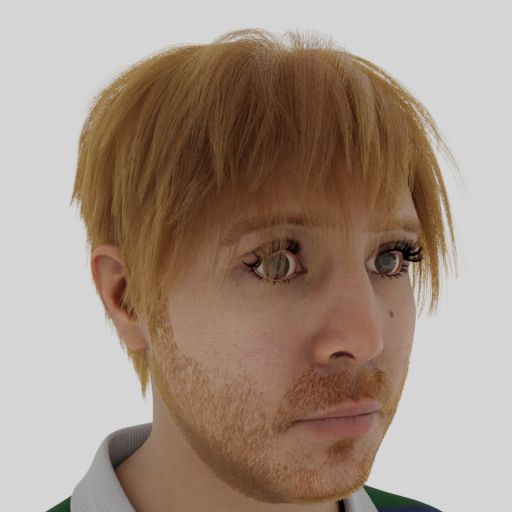}
        &  \includegraphics[width=2.5cm,height=2.5cm,keepaspectratio]{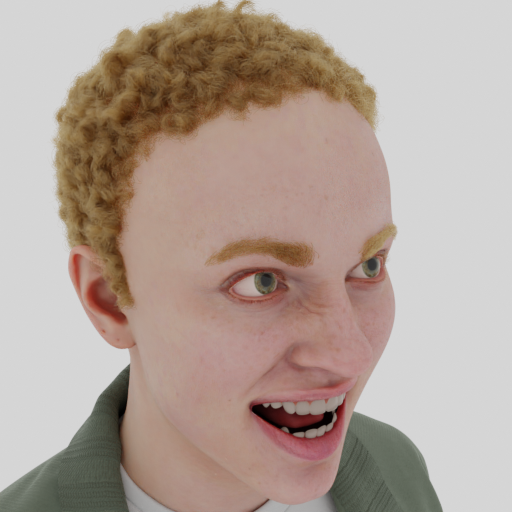}
        &  \includegraphics[width=2.5cm,height=2.5cm,keepaspectratio]{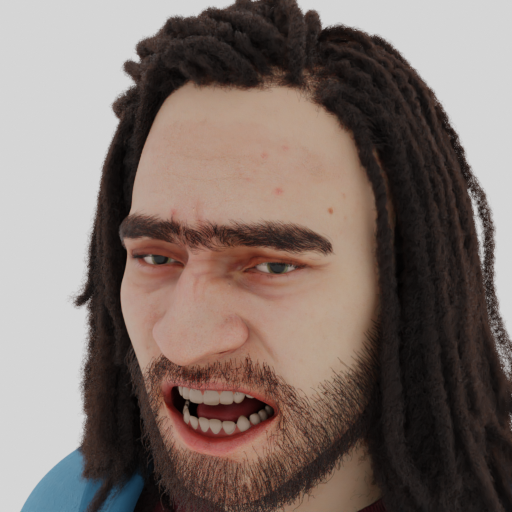}
        &  \includegraphics[width=2.5cm,height=2.5cm,keepaspectratio]{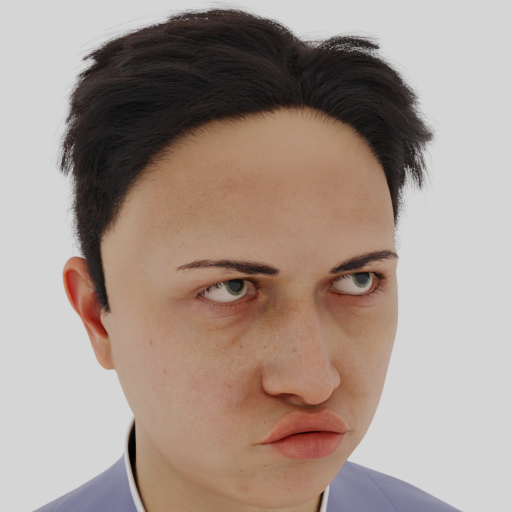} \\
        \rotatebox{90}{\parbox[t]{3cm}{\hspace*{\fill}Pseudo GT\hspace*{\fill}}}\hspace*{5pt}
        &  \includegraphics[width=2.5cm,height=2.5cm,keepaspectratio]{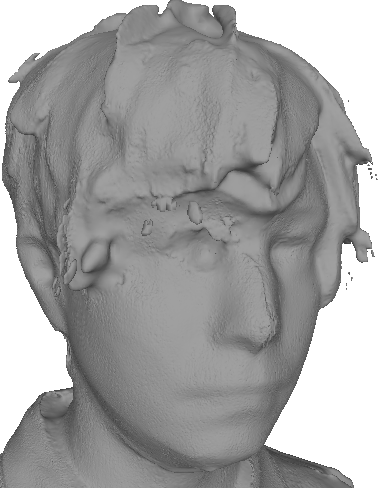}
        &  \includegraphics[width=2.5cm,height=2.5cm,keepaspectratio]{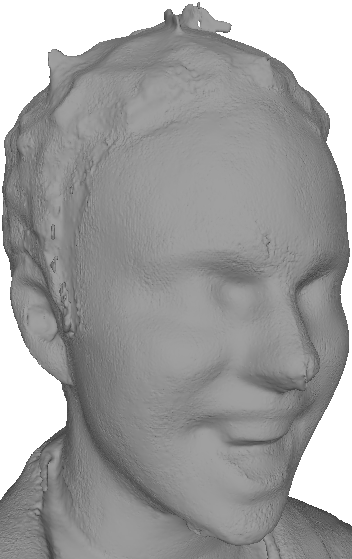}
        &  \includegraphics[width=2.5cm,height=2.5cm,keepaspectratio]{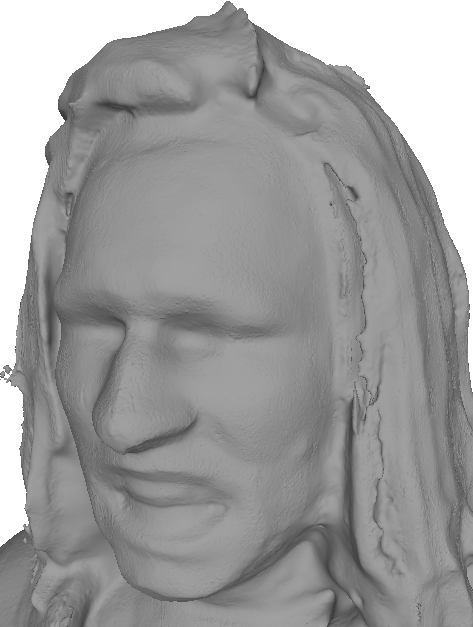}
        &  \includegraphics[width=2.5cm,height=2.5cm,keepaspectratio]{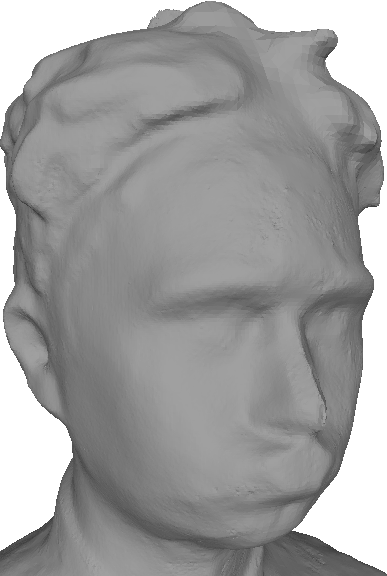} \\
        \rotatebox{90}{\parbox[t]{3cm}{\hspace*{\fill}EG3D Inversion\hspace*{\fill}}}\hspace*{5pt}
        &  \includegraphics[width=2.5cm,height=2.5cm,keepaspectratio]{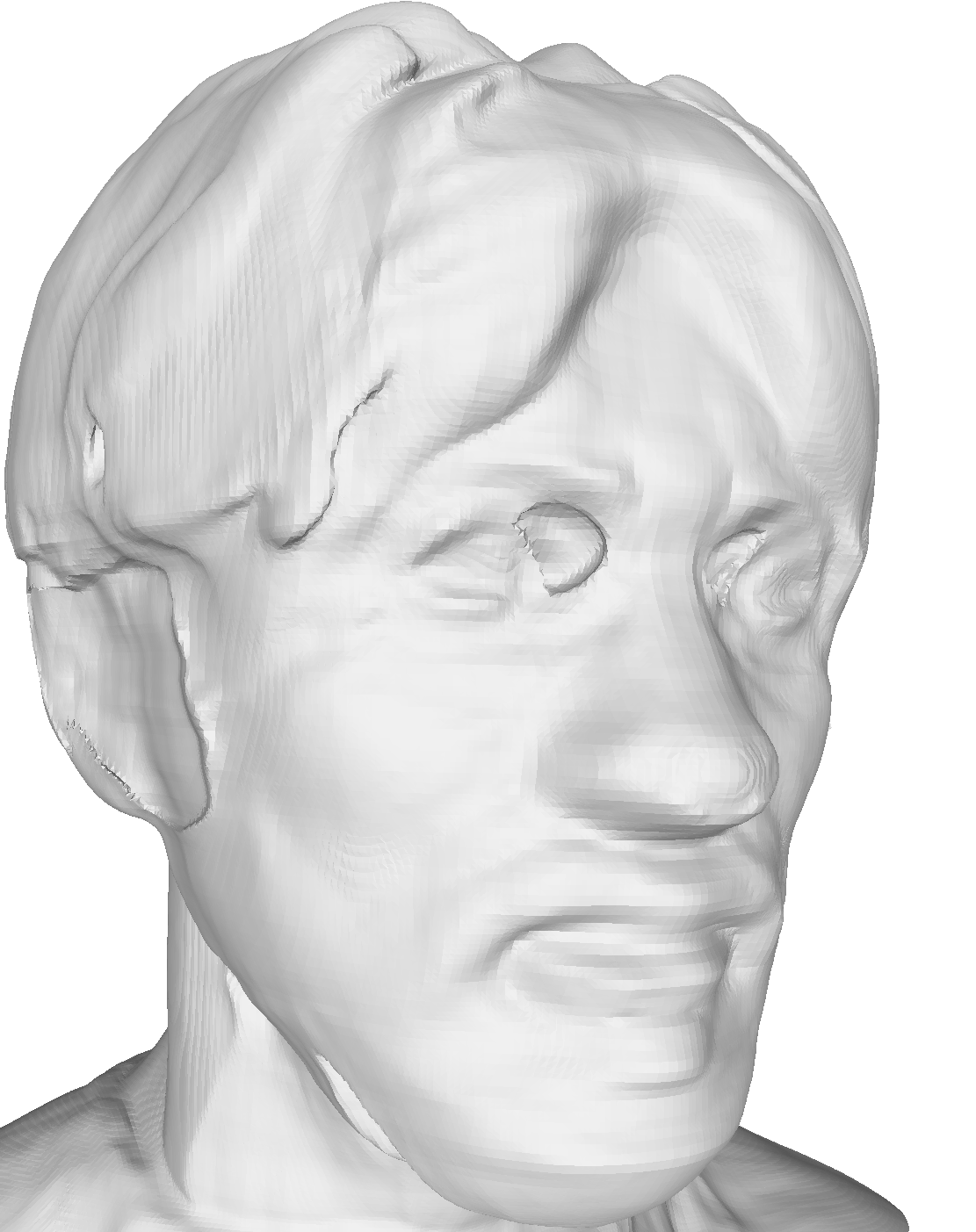}
        &  \includegraphics[width=2.5cm,height=2.5cm,keepaspectratio]{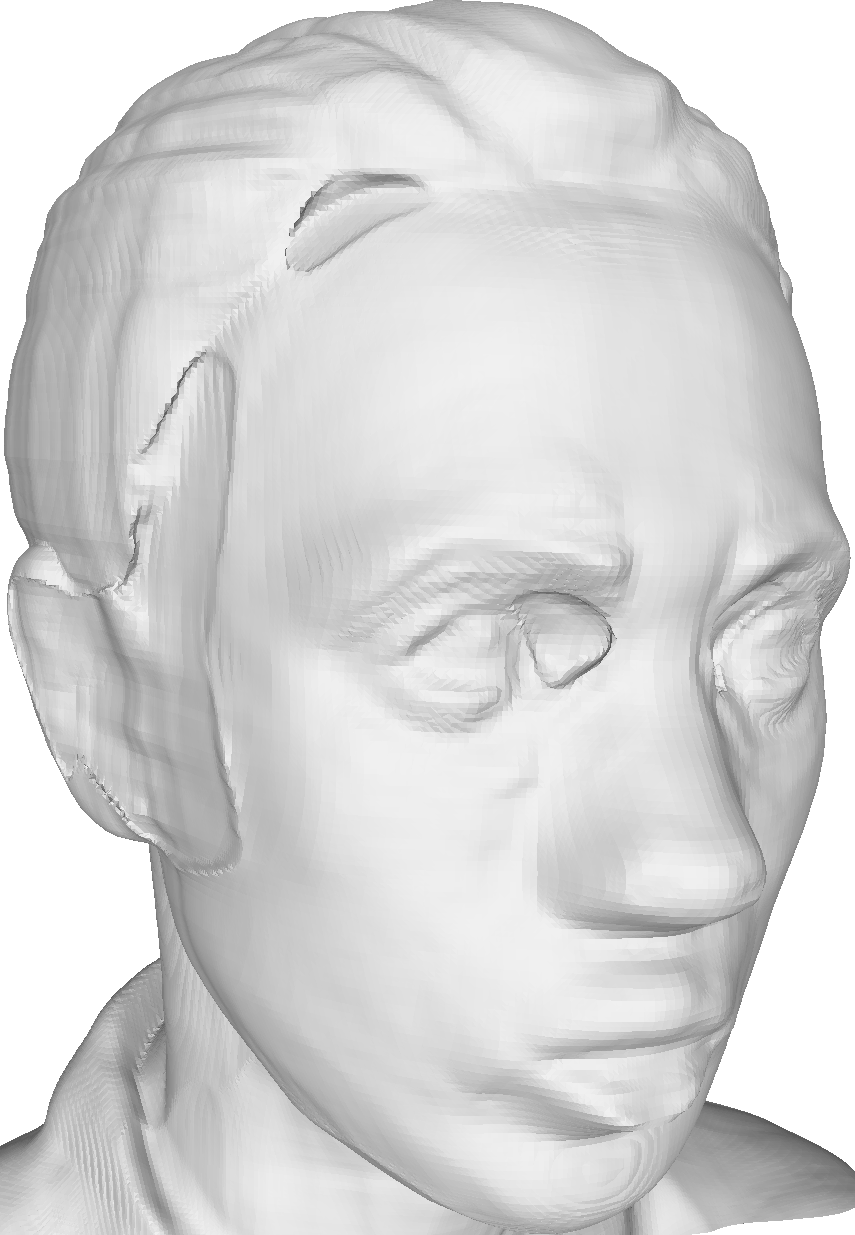}
        &  \includegraphics[width=2.5cm,height=2.5cm,keepaspectratio]{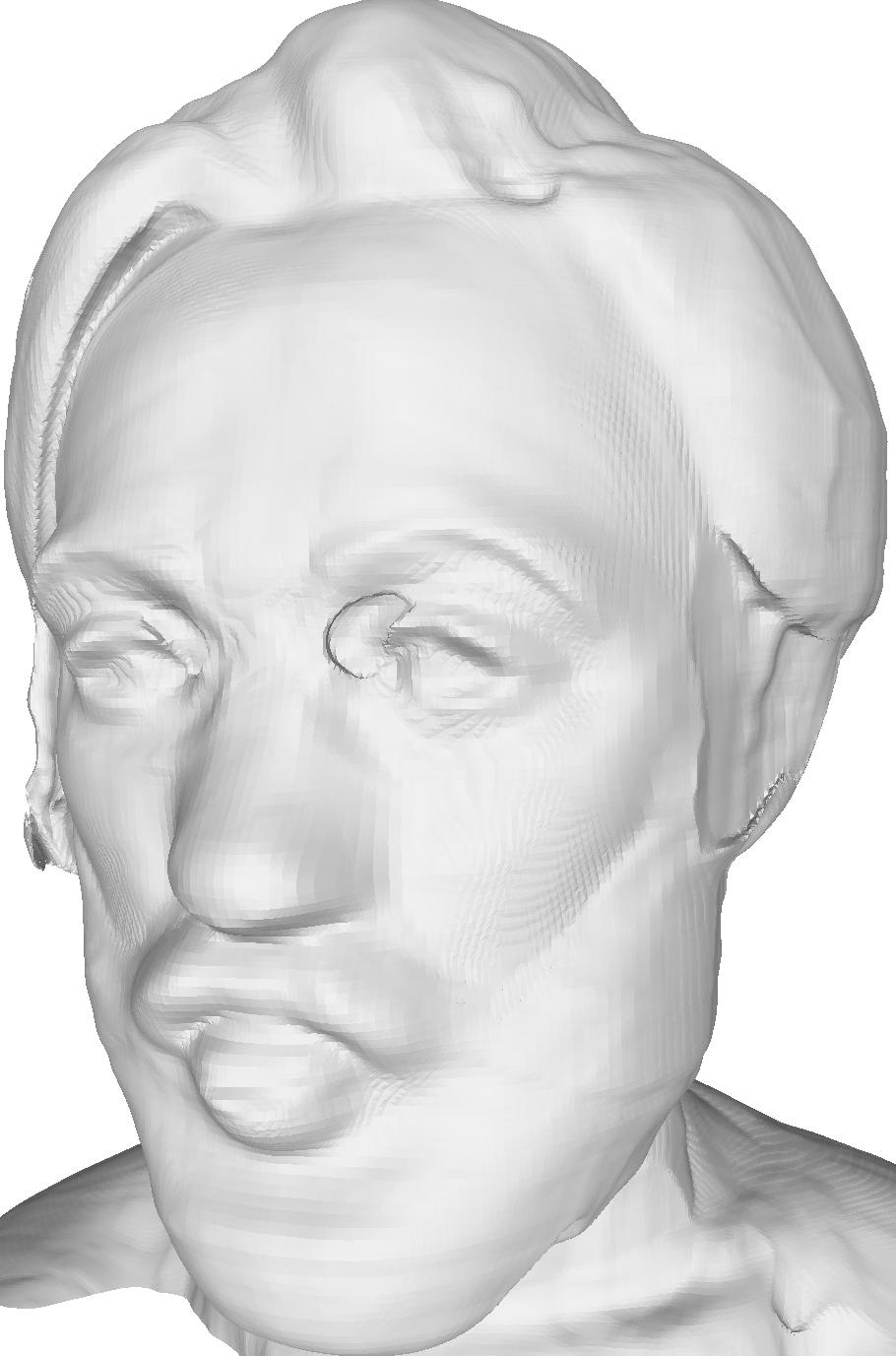}
        &  \includegraphics[width=2.5cm,height=2.5cm,keepaspectratio]{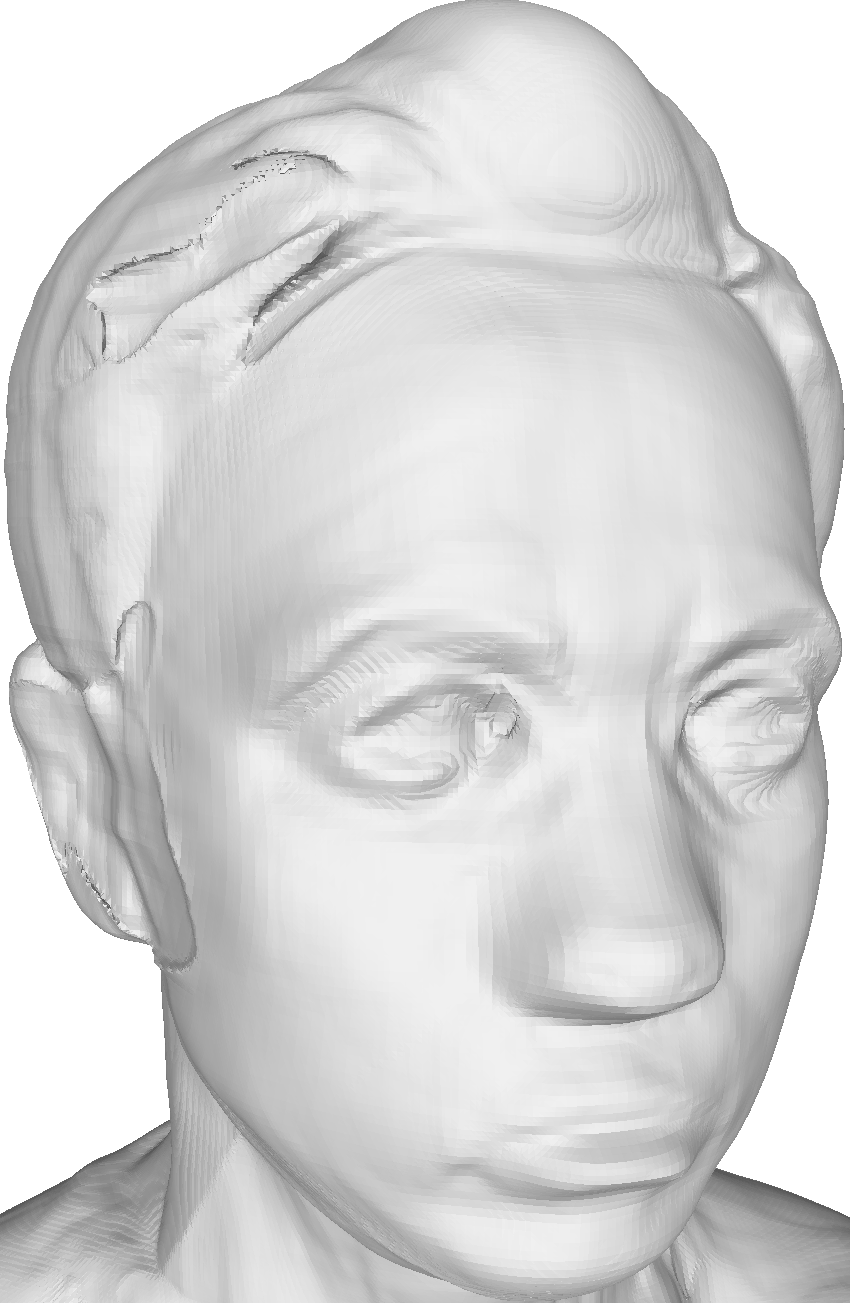} \\
        \rotatebox{90}{\parbox[t]{3cm}{\hspace*{\fill}GeoGen Inversion\hspace*{\fill}}}\hspace*{5pt}
        &  \includegraphics[width=2.5cm,height=2.5cm,keepaspectratio]{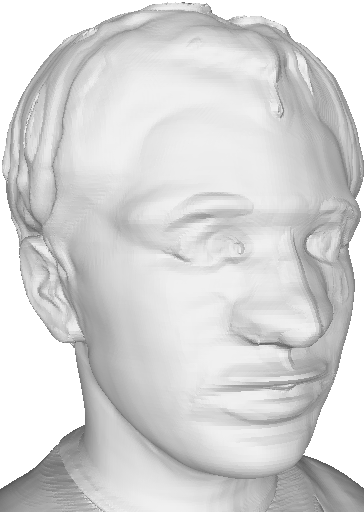}
        &  \includegraphics[width=2.5cm,height=2.5cm,keepaspectratio]{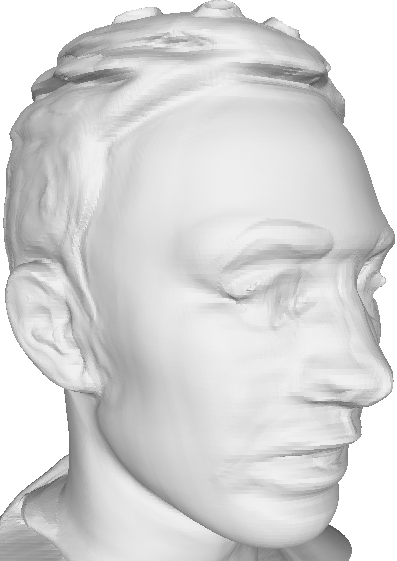}
        &  \includegraphics[width=2.5cm,height=2.5cm,keepaspectratio]{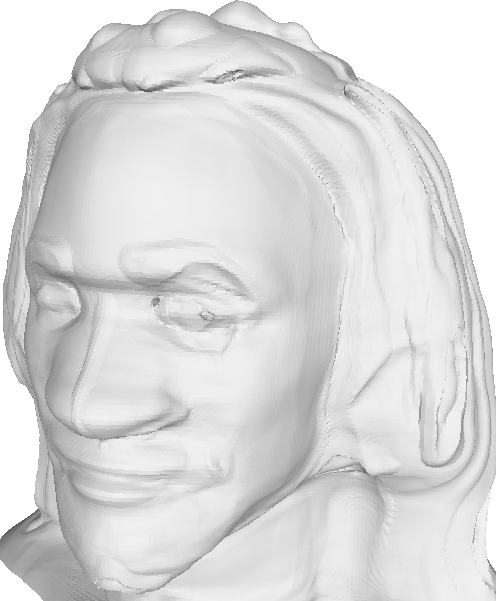}
        &  \includegraphics[width=2.5cm,height=2.5cm,keepaspectratio]{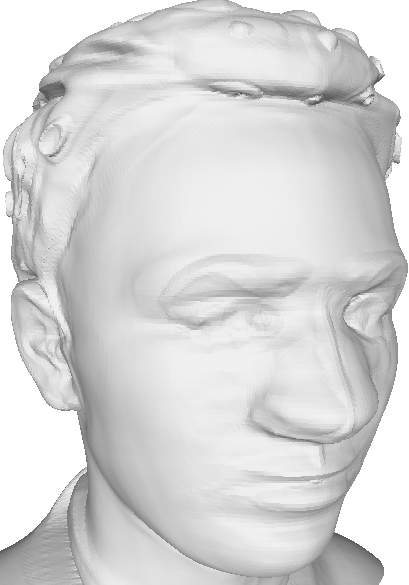} \\
    \end{tabular}
    }
    \vspace{-10pt}
    \caption{Qualitative inversion results on our synthetic face dataset, focusing on the comparison between the EG3D~\cite{chan2023eg3d} and GeoGen inversion methods. The corresponding latent source for the source held-out test input image is estimated for GeoGen using GAN inversion, revealing its ability to capture fine details with reduced noise and artifacts. In contrast, the EG3D~\cite{chan2023eg3d} inversion  meshes are observed to have significant artifacts, particularly around the ears, and display noticeable holes in the top regions of the eyes. Our inversion mesh is meticulously compared against pseudo ground truth, and reconstructed using Poisson surface reconstruction from multi-view images, underscoring the superiority of the GeoGen method in terms of fidelity and accuracy. Moreover, our inversion technique exhibits increased precision, contributing to a more authentic representation of the facial structure.}
    \vspace{-5pt}
    \label{fig:synthetics_inversion_supp}
\end{figure*}

\subsection{3D reconstruction metrics}

The assessment of 3D geometry is a critical aspect of our evaluation, as it reflects the ability of the generative models to faithfully reconstruct and represent the intricate geometric details of the subjects. Table 2 from the paper presents a comprehensive comparison of different 3D reconstruction metrics for generative models on ShapeNet \emph{Cars} and Synthetic Human \emph{Heads}. The selected metrics include Overall Chamfer Distance, Mean Squared Error (MSE), Hausdorff Distance (HD), Earth Mover's Distance, and Mean Surface Distance (MSD).

These metrics were chosen for their ability to capture various aspects of geometric fidelity. Chamfer Distance provides a measure of dissimilarity between two point sets, emphasizing both the precision and recall of the reconstructed surfaces. MSE offers insights into the mean differences between corresponding points, focusing on local accuracy. HD measures the maximum distance from a point in one set to the nearest point in the other set, highlighting global discrepancies. Earth Mover's Distance quantifies the minimum amount of work to transform one point set into the other, capturing overall distribution alignment. Lastly, MSD focuses on the mean distance between surfaces, reflecting surface smoothness and consistency.

In the process of evaluating these metrics, we scaled the generated and ground-truth meshes to fit within a unit sphere to ensure a consistent basis for comparison. We then randomly sampled 20,000 points from the meshes, repeating this process 20 times, in order to compute the mean and standard deviation of the metrics. This methodology allowed us to capture a comprehensive and statistically robust representation of the geometric quality, eliminating potential biases related to specific sampling patterns or scaling discrepancies.

The results, as shown in Table 2 of the main paper indicate that GeoGen demonstrates superior results, reflecting its ability to represent finer geometric details. The table also includes comparisons with GeoGen without SDF and DL constraints, allowing for an understanding of how specific components and constraints influence model performance. The best-performing methods for each dataset are highlighted in bold, striking a balance between quantitative performance and perceptual realism. The rigorous evaluation of these 3D metrics underscores the effectiveness of our approach and contributes to a nuanced understanding of generative modeling for complex geometric structures.

\section{Additional qualitative results}

In Figure~\ref{fig:synthetics_inversion_supp} we present a comparison of synthetic human avatar meshes across EG3D~\cite{chan2023eg3d} and GeoGen. It is qualitatively evident that our model, leveraging the capabilities of the Signed Distance Function (SDF) network with SDF depth consistency loss, surpasses both EG3D and StyleSDF (as shown in the main paper) in reconstructing detailed facial features, including the ears, nose, hair, and eyes.

Additionally, we demonstrate the ability of the GeoGen model in 3D reconstruction on the ShapeNet cars dataset in Figures~\ref{fig:car_zoom1} and Figure~\ref{fig:car_zoom} where it successfully reproduces granular details on the surface of the cars. 
We also show inversion results in Figure~\ref{fig:inversion_cars}. 
This distinction is further highlighted by contrasting the rendering qualities of the generated synthetic samples from the EG3D and GeoGen models, displayed in Figure 5, against some ground truth samples. Unlike the EG3D model~\cite{chan2023eg3d}, which exhibits a lack of granular details, our model's implementation of a more advanced SDF network, combined with robust SDF constraints and feature storage within a triplane, yields more precise and refined reconstructions. Thus, our approach consistently and effectively bridges the gap between visual perception and geometric representation, outperforming other techniques in 3D reconstruction fidelity. That is also visible in Figures~\ref{fig:car_zoom1} and ~\ref{fig:car_zoom} where GeoGen is able to better reconstruct the surface of synthetic faces using a GAN inversion technique~\cite{chan2023eg3d}.

\section{Acknowledgments}

The authors express their sincere appreciation to Microsoft Research for the provision of GPU clusters containing V100s and P100s. SE's work was supported by the UKRI CDT in Biomedical AI, with additional thanks to the UKRI funds and Microsoft for granting access to cloud services. KK's research received funding from Microsoft Research through the EMEA PhD Scholarship Programme, and he extends his gratitude to NVIDIA Corporation for GPU access provided by NVIDIA’s Academic Hardware Grants Program.

%% file: main.bbl
\begin{thebibliography}{43}
\providecommand{\natexlab}[1]{#1}
\providecommand{\url}[1]{\texttt{#1}}
\expandafter\ifx\csname urlstyle\endcsname\relax
  \providecommand{\doi}[1]{doi: #1}\else
  \providecommand{\doi}{doi: \begingroup \urlstyle{rm}\Url}\fi

\bibitem[An et~al.(2023)An, Xu, Shi, Song, Ogras, and Luo]{an2023panohead}
Sizhe An, Hongyi Xu, Yichun Shi, Guoxian Song, Umit~Y Ogras, and Linjie Luo.
\newblock Panohead: Geometry-aware 3d full-head synthesis in 360$^\circ$.
\newblock In \emph{Conference on Computer Vision and Pattern Recognition},
  2023.

\bibitem[Bi{\'n}kowski et~al.(2018)Bi{\'n}kowski, Sutherland, Arbel, and
  Gretton]{binkowski2018demystifying}
Miko{\l}aj Bi{\'n}kowski, Danica~J Sutherland, Michael Arbel, and Arthur
  Gretton.
\newblock Demystifying mmd gans.
\newblock In \emph{International Conference on Learning Representations}, 2018.

\bibitem[Chan(2022)]{chan2023eg3d}
Eric Chan.
\newblock Efficient geometry aware 3d network.
\newblock \emph{Computer Vision and Pattern Recognition}, 2022.

\bibitem[Chan et~al.(2021)Chan, Monteiro, Kellnhofer, Wu, and
  Wetzstein]{chan2021pi}
Eric~R Chan, Marco Monteiro, Petr Kellnhofer, Jiajun Wu, and Gordon Wetzstein.
\newblock pi-gan: Periodic implicit generative adversarial networks for
  3d-aware image synthesis.
\newblock In \emph{Conference on Computer Vision and Pattern Recognition},
  2021.

\bibitem[Chan et~al.(2022)Chan, Lin, Chan, Nagano, Pan, De~Mello, Gallo,
  Guibas, Tremblay, Khamis, et~al.]{chan2022efficient}
Eric~R Chan, Connor~Z Lin, Matthew~A Chan, Koki Nagano, Boxiao Pan, Shalini
  De~Mello, Orazio Gallo, Leonidas~J Guibas, Jonathan Tremblay, Sameh Khamis,
  et~al.
\newblock Efficient geometry-aware 3d generative adversarial networks.
\newblock In \emph{Conference on Computer Vision and Pattern Recognition},
  2022.

\bibitem[Chang et~al.(2015)Chang, Funkhouser, Guibas, Hanrahan, Huang, Li,
  Savarese, Savva, Song, Su, et~al.]{chang2015shapenet}
Angel~X Chang, Thomas Funkhouser, Leonidas Guibas, Pat Hanrahan, Qixing Huang,
  Zimo Li, Silvio Savarese, Manolis Savva, Shuran Song, Hao Su, et~al.
\newblock Shapenet: An information-rich 3d model repository.
\newblock \emph{arXiv:1512.03012}, 2015.

\bibitem[Chen and Zhang(2019)]{chen2019learning}
Zhiqin Chen and Hao Zhang.
\newblock Learning implicit fields for generative shape modeling.
\newblock In \emph{Conference on Computer Vision and Pattern Recognition},
  2019.

\bibitem[Deng et~al.(2019)Deng, Guo, Xue, and Zafeiriou]{deng2019arcface}
Jiankang Deng, Jia Guo, Niannan Xue, and Stefanos Zafeiriou.
\newblock Arcface: Additive angular margin loss for deep face recognition.
\newblock In \emph{Conference on Computer Vision and Pattern Recognition},
  2019.

\bibitem[Deng et~al.(2022)Deng, Yang, Xiang, and Tong]{deng2022gram}
Yu Deng, Jiaolong Yang, Jianfeng Xiang, and Xin Tong.
\newblock Gram: Generative radiance manifolds for 3d-aware image generation.
\newblock In \emph{Conference on Computer Vision and Pattern Recognition},
  2022.

\bibitem[Dhariwal and Nichol(2021)]{dhariwal2021diffusion}
Prafulla Dhariwal and Alexander Nichol.
\newblock Diffusion models beat gans on image synthesis.
\newblock \emph{Advances in Neural Information Processing Systems}, 2021.

\bibitem[Fu et~al.(2022)Fu, Xu, Ong, and Tao]{fu2022geo}
Qiancheng Fu, Qingshan Xu, Yew~Soon Ong, and Wenbing Tao.
\newblock Geo-neus: Geometry-consistent neural implicit surfaces learning for
  multi-view reconstruction.
\newblock \emph{Advances in Neural Information Processing Systems}, 2022.

\bibitem[Gadelha et~al.(2017)Gadelha, Maji, and Wang]{gadelha20173d}
Matheus Gadelha, Subhransu Maji, and Rui Wang.
\newblock 3d shape induction from 2d views of multiple objects.
\newblock In \emph{Conference on 3D Vision}, 2017.

\bibitem[Gao et~al.(2022)Gao, Shen, Wang, Chen, Yin, Li, Litany, Gojcic, and
  Fidler]{gao2022get3d}
Jun Gao, Tianchang Shen, Zian Wang, Wenzheng Chen, Kangxue Yin, Daiqing Li, Or
  Litany, Zan Gojcic, and Sanja Fidler.
\newblock Get3d: A generative model of high quality 3d textured shapes learned
  from images.
\newblock \emph{Advances In Neural Information Processing Systems}, 2022.

\bibitem[Goodfellow et~al.(2014)Goodfellow, Pouget-Abadie, Mirza, Xu,
  Warde-Farley, Ozair, Courville, and Bengio]{goodfellow2014generative}
I. Goodfellow, J. Pouget-Abadie, M. Mirza, B. Xu, D. Warde-Farley, S. Ozair, A.
  Courville, and Y. Bengio.
\newblock Generative adversarial nets.
\newblock In \emph{Advances in Neural Information Processing Systems}, 2014.

\bibitem[He et~al.(2015)He, Zhang, Ren, and Sun]{he2015delving}
Kaiming He, Xiangyu Zhang, Shaoqing Ren, and Jian Sun.
\newblock Delving deep into rectifiers: Surpassing human-level performance on
  imagenet classification.
\newblock \emph{International Conference on Computer Vision}, 2015.

\bibitem[Henderson et~al.(2020)Henderson, Tsiminaki, and
  Lampert]{henderson2020leveraging}
Paul Henderson, Vagia Tsiminaki, and Christoph~H Lampert.
\newblock Leveraging 2d data to learn textured 3d mesh generation.
\newblock In \emph{Conference on Computer Vision and Pattern Recognition},
  2020.

\bibitem[Heusel et~al.(2017)Heusel, Ramsauer, Unterthiner, Nessler, and
  Hochreiter]{heusel2017gans}
Martin Heusel, Hubert Ramsauer, Thomas Unterthiner, Bernhard Nessler, and Sepp
  Hochreiter.
\newblock Gans trained by a two time-scale update rule converge to a local nash
  equilibrium.
\newblock \emph{Advances in Neural Information Processing Systems}, 2017.

\bibitem[Ho et~al.(2020)Ho, Jain, and Abbeel]{ho2020denoising}
Jonathan Ho, Ajay Jain, and Pieter Abbeel.
\newblock Denoising diffusion probabilistic models.
\newblock \emph{Advances in Neural Information Processing Systems}, 2020.

\bibitem[Karras et~al.(2019)Karras, Laine, and Aila]{karras2019style}
Tero Karras, Samuli Laine, and Timo Aila.
\newblock A style-based generator architecture for generative adversarial
  networks.
\newblock In \emph{Conference on Computer Vision and Pattern Recognition},
  2019.

\bibitem[Karras et~al.(2020)Karras, Laine, Aittala, Hellsten, Lehtinen, and
  Aila]{karras2020analyzing}
Tero Karras, Samuli Laine, Miika Aittala, Janne Hellsten, Jaakko Lehtinen, and
  Timo Aila.
\newblock Analyzing and improving the image quality of stylegan.
\newblock In \emph{Conference on Computer Vision and Pattern Recognition},
  2020.

\bibitem[Karras et~al.(2021)Karras, Aittala, Laine, H{\"a}rk{\"o}nen, Hellsten,
  Lehtinen, and Aila]{karras2021alias}
Tero Karras, Miika Aittala, Samuli Laine, Erik H{\"a}rk{\"o}nen, Janne
  Hellsten, Jaakko Lehtinen, and Timo Aila.
\newblock Alias-free generative adversarial networks.
\newblock \emph{Advances in Neural Information Processing Systems}, 2021.

\bibitem[Kazhdan et~al.(2006{\natexlab{a}})Kazhdan, Bolitho, and
  Hoppe]{kazhdan2006poisson}
Michael Kazhdan, Matthew Bolitho, and Hugues Hoppe.
\newblock Poisson surface reconstruction.
\newblock In \emph{Fourth Eurographics symposium on Geometry processing},
  2006{\natexlab{a}}.

\bibitem[Kazhdan et~al.(2006{\natexlab{b}})Kazhdan, Bolitho, and
  Hoppe]{poisson_surface}
Michael Kazhdan, Matthew Bolitho, and Hugues Hoppe.
\newblock Poisson surface reconstruction.
\newblock In \emph{Eurographics symposium on Geometry processing},
  2006{\natexlab{b}}.

\bibitem[Kingma and Welling(2014)]{kingma2013auto}
Diederik~P Kingma and Max Welling.
\newblock Auto-encoding variational bayes.
\newblock In \emph{International Conference on Learning Representations}, 2014.

\bibitem[Mescheder et~al.(2019)Mescheder, Oechsle, Niemeyer, Nowozin, and
  Geiger]{mescheder2019occupancy}
Lars Mescheder, Michael Oechsle, Michael Niemeyer, Sebastian Nowozin, and
  Andreas Geiger.
\newblock Occupancy networks: Learning 3d reconstruction in function space.
\newblock In \emph{Conference on Computer Vision and Pattern Recognition},
  2019.

\bibitem[Mildenhall et~al.(2020)Mildenhall, Tancik, Barron, Ramamoorthi, Ng,
  and Martin-Brualla]{mildenhall2020nerf}
Ben Mildenhall, Matthew Tancik, Jonathan~T Barron, Ravi Ramamoorthi, Ren Ng,
  and Ricardo Martin-Brualla.
\newblock Nerf: Representing scenes as neural radiance fields for view
  synthesis.
\newblock In \emph{European Conference on Computer Vision}, 2020.

\bibitem[Nguyen-Phuoc et~al.(2019)Nguyen-Phuoc, Li, Theis, Richardt, and
  Yang]{nguyen2019hologan}
Thu Nguyen-Phuoc, Chuan Li, Lucas Theis, Christian Richardt, and Yong-Liang
  Yang.
\newblock Hologan: Unsupervised learning of 3d representations from natural
  images.
\newblock In \emph{International Conference on Computer Vision}, 2019.

\bibitem[Nguyen-Phuoc et~al.(2020)Nguyen-Phuoc, Richardt, Mai, Yang, and
  Mitra]{nguyen2020blockgan}
Thu~H Nguyen-Phuoc, Christian Richardt, Long Mai, Yongliang Yang, and Niloy
  Mitra.
\newblock Blockgan: Learning 3d object-aware scene representations from
  unlabelled images.
\newblock \emph{Advances in Neural Information Processing Systems}, 2020.

\bibitem[Niemeyer and Geiger(2021)]{niemeyer2021giraffe}
Michael Niemeyer and Andreas Geiger.
\newblock Giraffe: Representing scenes as compositional generative neural
  feature fields.
\newblock In \emph{Conference on Computer Vision and Pattern Recognition},
  2021.

\bibitem[Oechsle et~al.(2021)Oechsle, Peng, and Geiger]{oechsle2021unisurf}
Michael Oechsle, Songyou Peng, and Andreas Geiger.
\newblock Unisurf: Unifying neural implicit surfaces and radiance fields for
  multi-view reconstruction.
\newblock In \emph{International Conference on Computer Vision}, 2021.

\bibitem[Or-El et~al.(2022)Or-El, Luo, Shan, Shechtman, Park, and
  Kemelmacher-Shlizerman]{or2022stylesdf}
Roy Or-El, Xuan Luo, Mengyi Shan, Eli Shechtman, Jeong~Joon Park, and Ira
  Kemelmacher-Shlizerman.
\newblock Stylesdf: High-resolution 3d-consistent image and geometry
  generation.
\newblock In \emph{Conference on Computer Vision and Pattern Recognition},
  2022.

\bibitem[Park et~al.(2019)Park, Florence, Straub, Newcombe, and
  Lovegrove]{park2019deepsdf}
Jeong~Joon Park, Philip Florence, Julian Straub, Richard Newcombe, and Steven
  Lovegrove.
\newblock Deepsdf: Learning continuous signed distance functions for shape
  representation.
\newblock In \emph{Conference on Computer Vision and Pattern Recognition},
  2019.

\bibitem[Roich et~al.(2022)Roich, Mokady, Bermano, and
  Cohen-Or]{roich2022pivotal}
Daniel Roich, Ron Mokady, Amit~H Bermano, and Daniel Cohen-Or.
\newblock Pivotal tuning for latent-based editing of real images.
\newblock \emph{Transactions on Graphics}, 2022.

\bibitem[Schwarz et~al.(2020)Schwarz, Liao, Niemeyer, and
  Geiger]{schwarz2020graf}
Katja Schwarz, Yiyi Liao, Michael Niemeyer, and Andreas Geiger.
\newblock Graf: Generative radiance fields for 3d-aware image synthesis.
\newblock \emph{Advances in Neural Information Processing Systems}, 2020.

\bibitem[Schwarz et~al.(2022)Schwarz, Sauer, Niemeyer, Liao, and
  Geiger]{schwarz2022voxgraf}
Katja Schwarz, Axel Sauer, Michael Niemeyer, Yiyi Liao, and Andreas Geiger.
\newblock Voxgraf: Fast 3d-aware image synthesis with sparse voxel grids.
\newblock \emph{Advances in Neural Information Processing Systems}, 2022.

\bibitem[Shin et~al.(2023)Shin, Seo, Bae, Choi, Kim, Byun, and
  Uh]{shin2023ballgan}
Minjung Shin, Yunji Seo, Jeongmin Bae, Young~Sun Choi, Hyunsu Kim, Hyeran Byun,
  and Youngjung Uh.
\newblock Ballgan: 3d-aware image synthesis with a spherical background.
\newblock In \emph{International Conference on Computer Vision}, 2023.

\bibitem[Song et~al.(2021)Song, Sohl-Dickstein, Kingma, Kumar, Ermon, and
  Poole]{song2020score}
Yang Song, Jascha Sohl-Dickstein, Diederik~P Kingma, Abhishek Kumar, Stefano
  Ermon, and Ben Poole.
\newblock Score-based generative modeling through stochastic differential
  equations.
\newblock In \emph{ICLR}, 2021.

\bibitem[Tov et~al.(2023)Tov, Doe, and Smith]{tov2023pivotal}
E. Tov, J. Doe, and A. Smith.
\newblock Pivotal tuning inversion.
\newblock \emph{Computer Graphics Forum}, 2023.

\bibitem[Wang et~al.(2021)Wang, Liu, Liu, Theobalt, Komura, and
  Wang]{wang2021neus}
Peng Wang, Lingjie Liu, Yuan Liu, Christian Theobalt, Taku Komura, and Wenping
  Wang.
\newblock Neus: Learning neural implicit surfaces by volume rendering for
  multi-view reconstruction.
\newblock In \emph{Advances in Neural Information Processing Systems}, 2021.

\bibitem[Wang et~al.(2023)Wang, Zhang, Zhang, Gu, Bao, Baltrusaitis, Shen,
  Chen, Wen, Chen, et~al.]{wang2023rodin}
Tengfei Wang, Bo Zhang, Ting Zhang, Shuyang Gu, Jianmin Bao, Tadas
  Baltrusaitis, Jingjing Shen, Dong Chen, Fang Wen, Qifeng Chen, et~al.
\newblock Rodin: A generative model for sculpting 3d digital avatars using
  diffusion.
\newblock In \emph{Conference on Computer Vision and Pattern Recognition},
  2023.

\bibitem[Wood et~al.(2021)Wood, Baltru{\v{s}}aitis, Hewitt, Dziadzio, Cashman,
  and Shotton]{wood2021fake}
Erroll Wood, Tadas Baltru{\v{s}}aitis, Charlie Hewitt, Sebastian Dziadzio,
  Thomas~J Cashman, and Jamie Shotton.
\newblock Fake it till you make it: face analysis in the wild using synthetic
  data alone.
\newblock In \emph{International Conference on Computer Vision}, 2021.

\bibitem[Xu and Tao(2020)]{xu2020planar}
Qingshan Xu and Wenbing Tao.
\newblock Planar prior assisted patchmatch multi-view stereo.
\newblock In \emph{AAAI Conference on Artificial Intelligence}, 2020.

\bibitem[Yariv et~al.(2021)Yariv, Gu, Kasten, and Lipman]{yariv2021volume}
Lior Yariv, Jiatao Gu, Yoni Kasten, and Yaron Lipman.
\newblock Volume rendering of neural implicit surfaces.
\newblock \emph{Advances in Neural Information Processing Systems}, 2021.

\end{thebibliography}
